\documentclass[10pt,twocolumn,letterpaper]{article}

\usepackage[pagenumbers]{cvpr} 
\usepackage{graphicx}
\usepackage{array}
\usepackage{epigraph}
\usepackage{multirow}
\usepackage{booktabs}
\usepackage{pifont}
\usepackage{float}
\usepackage{url}            

\usepackage[pagebackref,breaklinks,colorlinks,citecolor=cvprblue]{hyperref}
\usepackage{makecell}
\RequirePackage[breakable,skins]{tcolorbox}
\RequirePackage{xcolor}
\definecolor{colorcommentfg}{RGB}{0,63,87}
\colorlet{colorchangebg}{black!2}
\colorlet{colorchangeframe}{black!20}
\definecolor{lightpastelpurple}{rgb}{0.69, 0.61, 0.85}
\definecolor{mediumpurple}{rgb}{0.58, 0.44, 0.86}
\definecolor{cvprblue}{rgb}{0.21,0.49,0.74}
\usepackage{colortbl}

\definecolor{colorcommentbg}{HTML}{CCCCFF}
\definecolor{colorcommentframe}{HTML}{76608A}
\newenvironment{intentionprompt}[1][]{\refstepcounter{reviewcomment@counter}
	\begin{tcolorbox}[adjusted title={Intention Generation Prompt for GPT3.5}, fonttitle={\bfseries\footnotesize}, fontupper=\scriptsize, colback={colorcommentbg!40}, colframe={colorcommentframe!90},coltitle={white},#1]
}{\end{tcolorbox}}

\definecolor{colorcommentbg_qualityprompt}{HTML}{A0522D}
\definecolor{colorcommentframe_qualityprompt}{HTML}{6D1F00}
\newenvironment{qualityprompt}[1][]{\refstepcounter{reviewcomment@counter}
	\begin{tcolorbox}[adjusted title={Quality Assurance Prompt for GPT4V(ision)}, fonttitle={\bfseries\footnotesize}, fontupper=\scriptsize, colback={colorcommentbg_qualityprompt!30}, colframe={colorcommentframe_qualityprompt!80},coltitle={white},#1]
}{\end{tcolorbox}}

\definecolor{colorcommentbg_pkprompt}{HTML}{5D8AA8}
\definecolor{colorcommentframe_pkprompt}{HTML}{0093AF}
\newenvironment{pkprompt}[1][]{\refstepcounter{reviewcomment@counter}
	\begin{tcolorbox}[adjusted title={Pair-wise Comparison Prompt for GPT4V(ision)}, fonttitle={\bfseries\footnotesize}, fontupper=\scriptsize, colback={colorcommentbg_pkprompt!30}, colframe={colorcommentframe_pkprompt!80},coltitle={white},#1]
}{\end{tcolorbox}}

\newlength\savewidth\newcommand\shline{\noalign{\global\savewidth\arrayrulewidth
  \global\arrayrulewidth 1pt}\hline\noalign{\global\arrayrulewidth\savewidth}}
\newcommand{\tablestyle}[2]{\setlength{\tabcolsep}{#1}\renewcommand{\arraystretch}{#2}\centering\footnotesize}

\makeatletter
\def\@fnsymbol#1{\ensuremath{\ifcase#1\or \dagger\or \ddagger\or
\mathsection\or \mathparagraph\or \|\or **\or \dagger\dagger
\or \ddagger\ddagger \else\@ctrerr\fi}}
\makeatother

\newcommand\codeurl[1]{{{\color{blue}{\url{#1}}}}}

\newcommand{\dalle}{DALL$\cdot$E3\xspace}

\newcommand{\gptv}{GPT-4V(ision)\xspace}
\newcommand{\ourname}{\textsc{Cole}\xspace}
\newcommand{\ourbenchmark}{\textsc{DesignerIntention}\xspace}
\newcommand{\cmark}{\ding{51}}%
\newcommand{\xmark}{\ding{55}}%

\title{\ourname: A Hierarchical Generation Framework for\\ Multi-Layered and Editable Graphic Design}

\author{
{\normalsize Peidong Jia$^{1}$ \quad\quad  Chenxuan Li$^{1}$ \quad\quad  Yuhui Yuan$^{1,3}$ \quad\quad  Zeyu Liu$^{2}$
 \quad\quad  Yichao Shen$^{2}$  \quad\quad  Bohan Chen$^{2}$}\\[0mm]
{\normalsize Xingru Chen$^{2}$ \hspace{2.1mm} Yinglin Zheng$^{2}$ \hspace{2.1mm} Dong Chen\hspace{2.1mm} Ji Li \hspace{2.1mm}
Xiaodong Xie \hspace{2.1mm} Shanghang Zhang \hspace{2.1mm} Baining Guo}\\[0mm]
 {\small$^1$joint core contribution \qquad $^2$interns at microsoft \qquad $^3$project lead}\\[1mm]
\normalsize{Microsoft Research Asia\quad\quad Peking University}\\
{\footnotesize\codeurl{{https://graphic-design-generation-github-io.vercel.app}}}\vspace{-4mm}}
\begin{document}

\twocolumn[{%
\renewcommand\twocolumn[1][]{#1}%
\maketitle
\begin{center}
\begin{minipage}[t]{1\linewidth}
\centering
\begin{minipage}{1\textwidth}
\centering
{\includegraphics[width=0.97\textwidth]{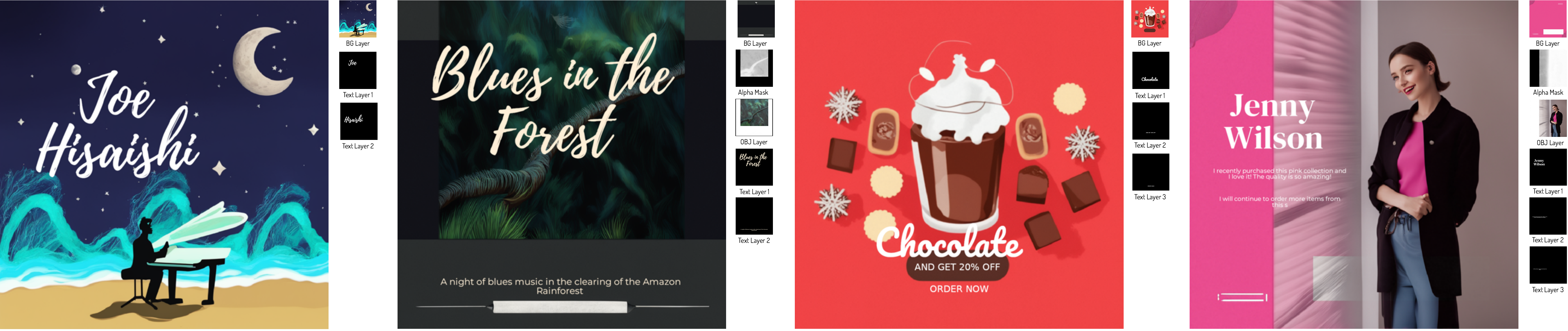}}
\end{minipage}
\begin{minipage}{0.1165\textwidth}
{\includegraphics[width=\textwidth]{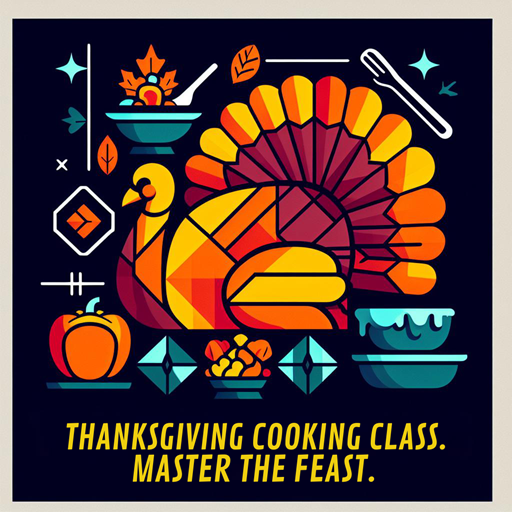}}
\vspace{-0mm}
\end{minipage}
\begin{minipage}{0.1165\textwidth}
{\includegraphics[width=\textwidth]{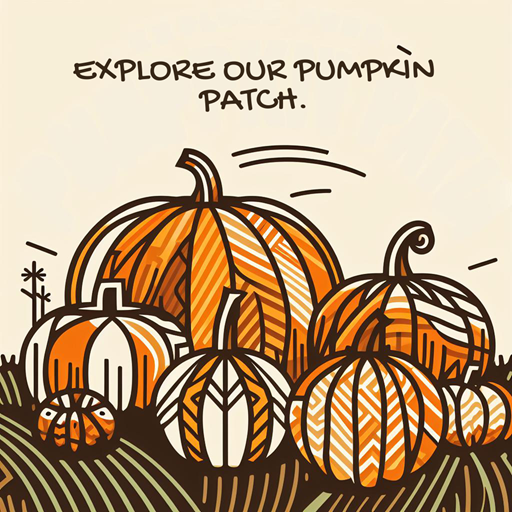}}
\vspace{-0mm}
\end{minipage}
\begin{minipage}{0.1165\textwidth}
{\includegraphics[width=\textwidth]{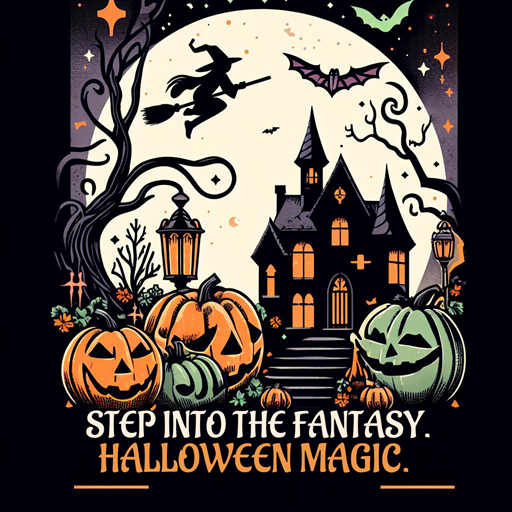}}
\vspace{-0mm}
\end{minipage}
\begin{minipage}{0.1165\textwidth}
{\includegraphics[width=\textwidth]{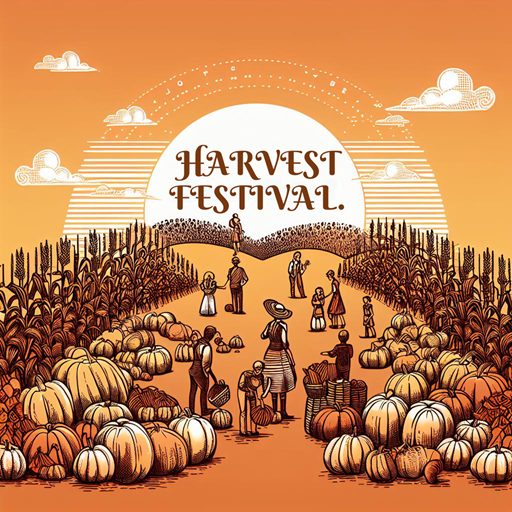}}
\vspace{-0mm}
\end{minipage}
\begin{minipage}{0.1165\textwidth}
{\includegraphics[width=\textwidth]{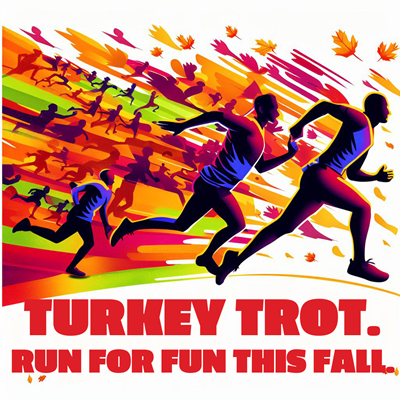}}
\vspace{-0mm}
\end{minipage}
\begin{minipage}{0.1165\textwidth}
{\includegraphics[width=\textwidth]{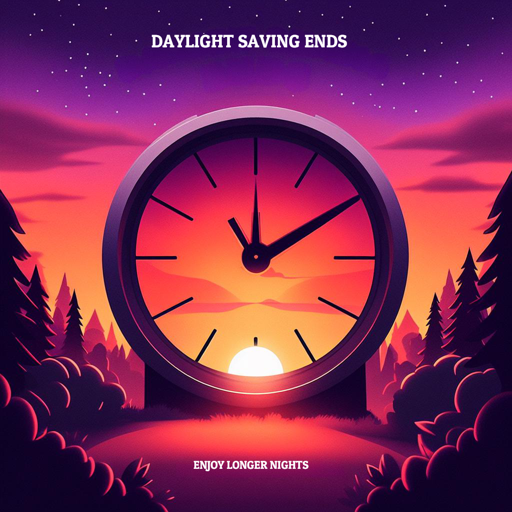}}
\vspace{-0mm}
\end{minipage}
\begin{minipage}{0.1165\textwidth}
{\includegraphics[width=\textwidth]{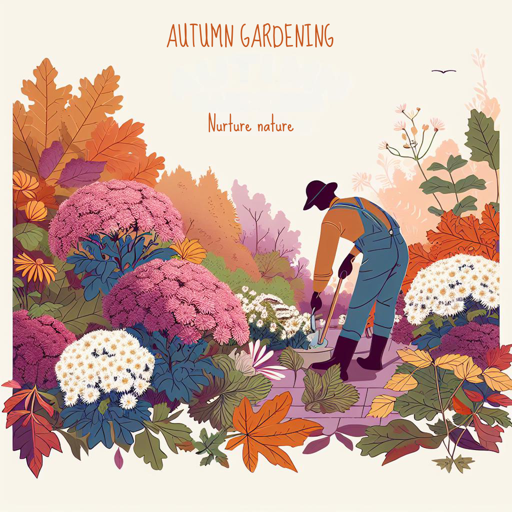}}
\vspace{-0mm}
\end{minipage}
\begin{minipage}{0.1165\textwidth}
{\includegraphics[width=\textwidth]{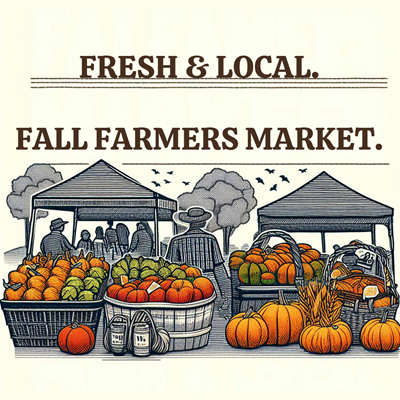}}
\vspace{-0mm}
\end{minipage}
\vspace{-2mm}
\captionof{figure}{\small{Illustrating the multi-layered graphic design images generated by our \ourname system (first row, we display the multi-layer image layers at the top-right corner of each design image) and the combination of \dalle background images and \ourname system (second row). 
See the appendix for detailed intention prompts. As shown in the second row, our \ourname system skillfully plans design layouts and selects harmonious fonts, colors, sizes, and positions through insightful analysis and reasoning, even with out-of-domain \dalle background images after pre-processing. By default, we do not use \dalle background images in all other results.}}
\label{fig:teaser}
\end{minipage}
\end{center}
}]

\begin{abstract}
Graphic design, which has been evolving since the 15th century, plays a crucial role in advertising. The creation of high-quality designs demands \emph{design-oriented planning, reasoning, and layer-wise generation}. This intricate task involves understanding vague intentions and faithfully generating multi-layered visual elements, including the background, decorations, objects, and fonts, among others. It also requires layout planning for all elements and visual reasoning to generate all visual elements that satisfy visual design principles.
Unlike the recent CanvaGPT, which integrates GPT-4 with existing design templates to build a custom GPT, this paper introduces the \ourname\footnote{\scriptsize{In 1843, Henry Cole introduced the world's first commercial Christmas card~\cite{wiki_henry_cole}. We commemorate his contributions with the \ourname system.}} system—a hierarchical generation framework designed to comprehensively address these challenges.
This \ourname system can transform a vague intention prompt into a high-quality multi-layered graphic design, while also supporting flexible editing based on user input. Examples of such input might include directives like ``design a poster for Hisaishi's concert.''
The key insight is to \emph{dissect the complex task of text-to-design generation into a hierarchy of simpler sub-tasks, each addressed by specialized models working collaboratively.} The results from these models are then consolidated to produce a cohesive final output. Our hierarchical task decomposition can streamline the complex process and significantly enhance generation reliability.  
Our \ourname system comprises multiple fine-tuned Large Language Models (LLMs), Large Multimodal Models (LMMs), and Diffusion Models (DMs), each specifically tailored for design-aware layer-wise captioning, layout planning, reasoning, and the task of generating images and text.
Furthermore, we construct the \ourbenchmark benchmark to demonstrate the superiority of our \ourname system over existing methods in generating high-quality graphic designs from user intent.
Last, we present a Canva-like multi-layered image editing tool to support flexible editing of the generated multi-layered graphic design images.
We perceive our \ourname system as an important step towards addressing more complex and multi-layered graphic design generation tasks in the future.
\end{abstract}
\vspace{-3mm}
\section{Introduction}
\label{sec:intro}

The recent advancement in the quality of natural image generation has been remarkable, elevating it to the level of professional photography. This leap forward can be attributed to the development of technologies such as Imagen~\cite{deng2009imagenet}, SDXL~\cite{podell2023sdxl}, and \dalle~\cite{dalle3paper}. The pivotal factors contributing to these advancements include the use of powerful Large Language Model (LLM) as a text encoder, the scaling up of training datasets, increased model complexity, the design of advanced sampling strategies, and the enhancement of data quality, among others. We believe that now is the opportune moment to redirect our efforts towards more professional image generation, particularly the graphic design generation considering its pivotal roles in advertising, branding, and marketing.

Graphic design~\cite{wiki_graphic_design}, a professional discipline, leverages the power of visual communication to convey targeted messages to specific social groups, with clear objectives. It is a realm that necessitates creativity, innovation, and lateral thinking. Typically, graphic design employs either manual or digital tools to amalgamate text and graphics, thus creating compelling visual narratives. Its primary goal is to structure information, give substance to ideas, and infuse expression and emotion into artifacts that chronicle human experiences. Graphic design often utilizes the artistry of typography, text composition, ornamentation, and imagery to convey thoughts, emotions, and attitudes that transcend the capabilities of language alone. In essence, it demands a high degree of creativity, innovation, and lateral thinking to produce top-tier designs.

\begin{figure}[t]
\begin{minipage}[t]{1\linewidth}
\centering
\begin{subfigure}[b]{0.242\textwidth}
\includegraphics[width=\textwidth]{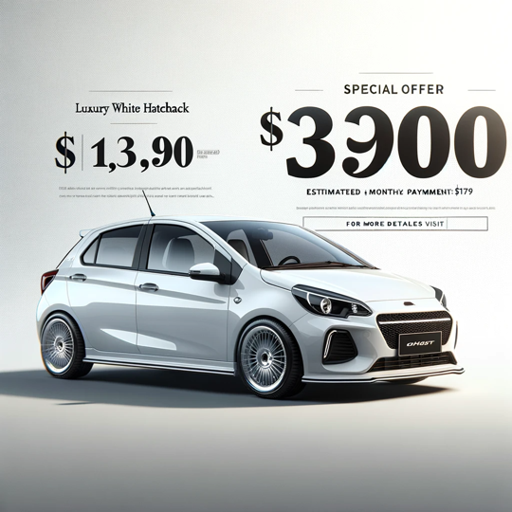}
\vspace{-3mm}
\end{subfigure}
\begin{subfigure}[b]{0.242\textwidth}
{\includegraphics[width=\textwidth]{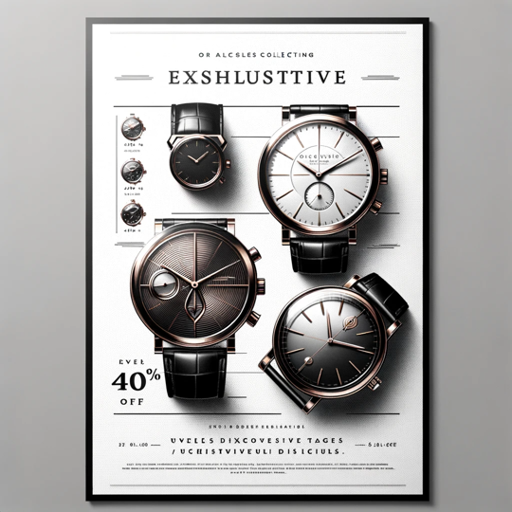}}
\vspace{-3mm}
\end{subfigure}
\begin{subfigure}[b]{0.242\textwidth}
{\includegraphics[width=\textwidth]{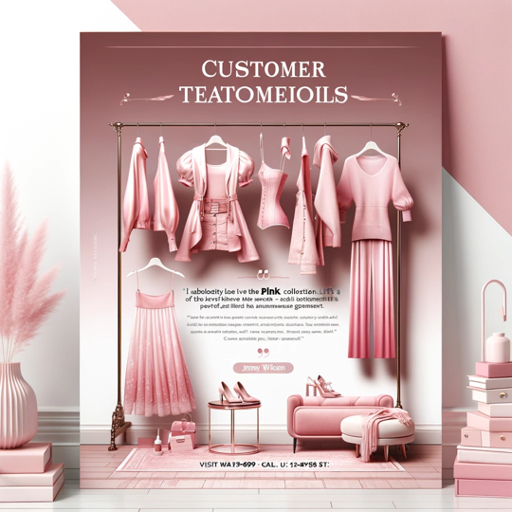}}
\vspace{-3mm}
\end{subfigure}
\begin{subfigure}[b]{0.242\textwidth}
{\includegraphics[width=\textwidth]{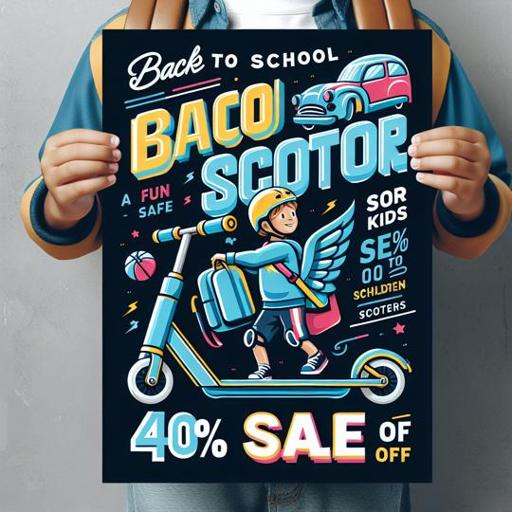}}
\vspace{-3mm}
\end{subfigure}
\begin{subfigure}[b]{0.242\textwidth}
\includegraphics[width=\textwidth]{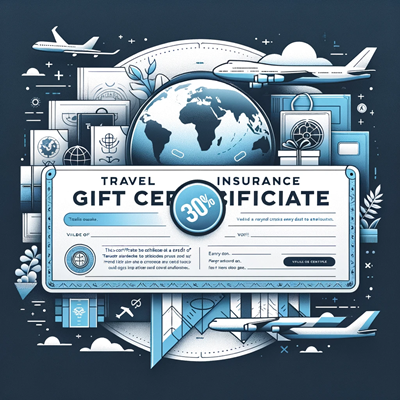}
\vspace{-3mm}
\end{subfigure}
\begin{subfigure}[b]{0.242\textwidth}
{\includegraphics[width=\textwidth]{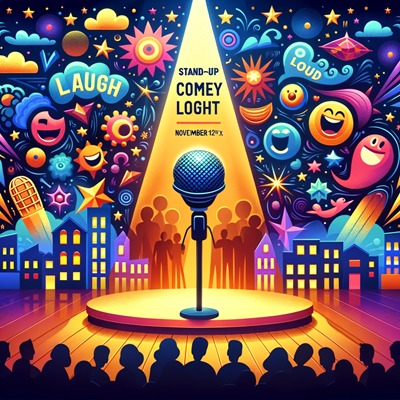}}
\vspace{-3mm}
\end{subfigure}
\begin{subfigure}[b]{0.242\textwidth}
{\includegraphics[width=\textwidth]{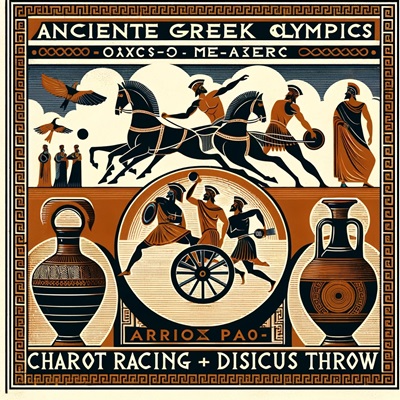}}
\vspace{-3mm}
\end{subfigure}
\begin{subfigure}[b]{0.242\textwidth}
{\includegraphics[width=\textwidth]{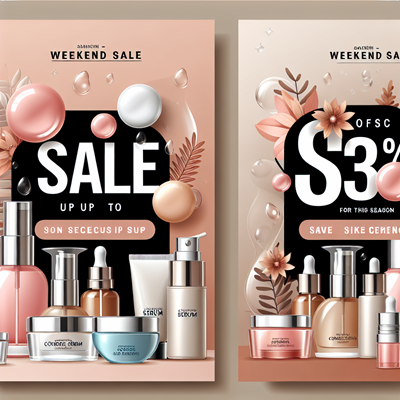}}
\vspace{-3mm}
\end{subfigure}
\caption{\small{Illustrating the design images generated by \dalle (augmented with GPT-$4$), using our \ourbenchmark.}}
\label{fig:DALLE3}
\end{minipage}
\end{figure}

The most recent study~\cite{lin2023designbench} has empirically shown that the groundbreaking \dalle exhibits impressive capabilities in generating high-quality design images, marked by visually striking graphics and layouts, as illustrated in Figure~\ref{fig:DALLE3}. However, these images are not without their flaws. They continue to confront several critical challenges, such as mis-rendered visual text, often resulting in missing or extra characters, a phenomenon also highlighted in ~\cite{dalle3paper}. Furthermore, editing these generated images demands complex operations such as segmentation, erasing, inpainting, and others, owing to their inherently uneditable nature. Another substantial limitation is the necessity for users to provide detailed text prompts - a task in graphic design generation that typically requires a high degree of professional expertise to develop effective prompts.
In contrast to \dalle, our \ourname system, as demonstrated in Figure~\ref{fig:teaser}, is capable of generating multi-layered and editable graphic design images from simple user intentions.

The most recent study~\cite{lin2023designbench} has empirically shown that the groundbreaking \dalle exhibits impressive capabilities in generating high-quality design images, marked by visually striking graphics and layouts, as illustrated in Figure~\ref{fig:DALLE3}. However, these images are not without their flaws. They continue to confront several critical challenges, such as mis-rendered visual text, often resulting in missing or extra characters, a phenomenon also highlighted in ~\cite{dalle3paper}. Furthermore, editing these generated images demands complex operations such as segmentation, erasing, inpainting, and others, owing to their inherently uneditable nature. Another substantial limitation is the necessity for users to provide detailed text prompts - a task in graphic design generation that typically requires a high degree of professional expertise to develop effective prompts.
In contrast to \dalle, our \ourname system, as demonstrated in Figure~\ref{fig:teaser}, is capable of generating multi-layered and editable graphic design images from simple user intentions.

We maintain that these three limitations critically undermine the quality of the generated graphic design images. A scalable, high-quality graphic design generation system should ideally require minimal effort from users, produce accurate and high-quality typography information for a variety of purposes, and offer a flexible editing space. This would allow users to further refine the output, integrating human expertise when necessary.  

\begin{figure}[t]
\centering
\begin{minipage}{0.5\linewidth}
\centering
{\includegraphics[width=1\textwidth]{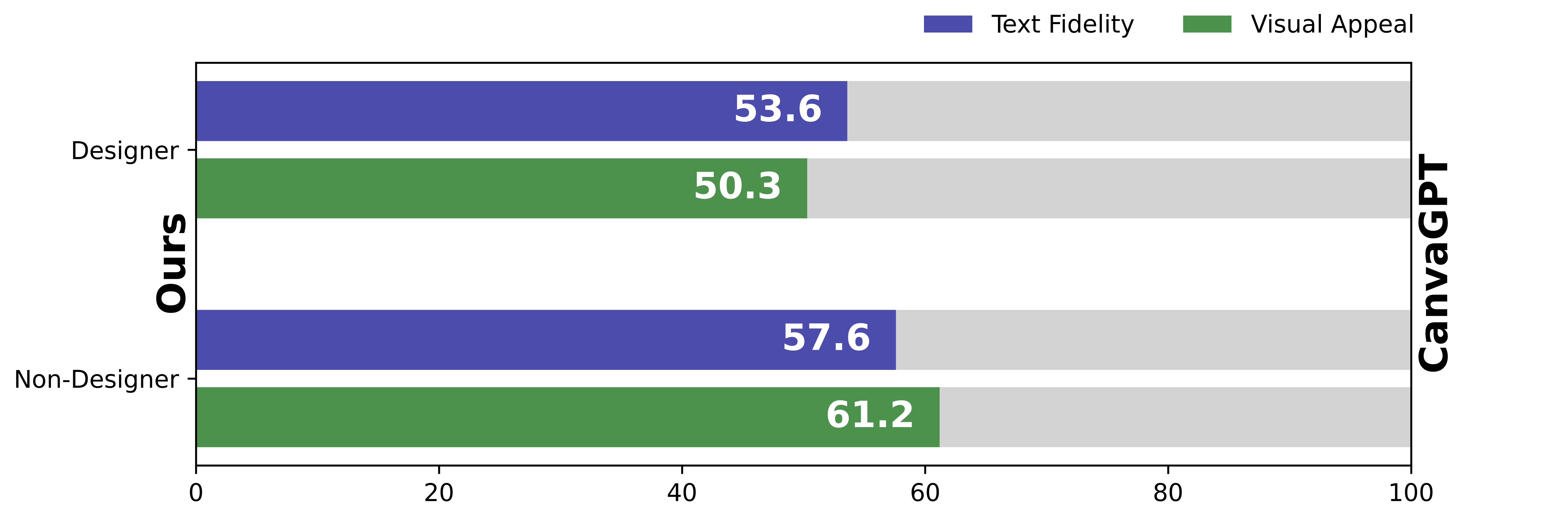}}
\end{minipage}%
\begin{minipage}{0.5\linewidth}
{\includegraphics[width=1\textwidth]{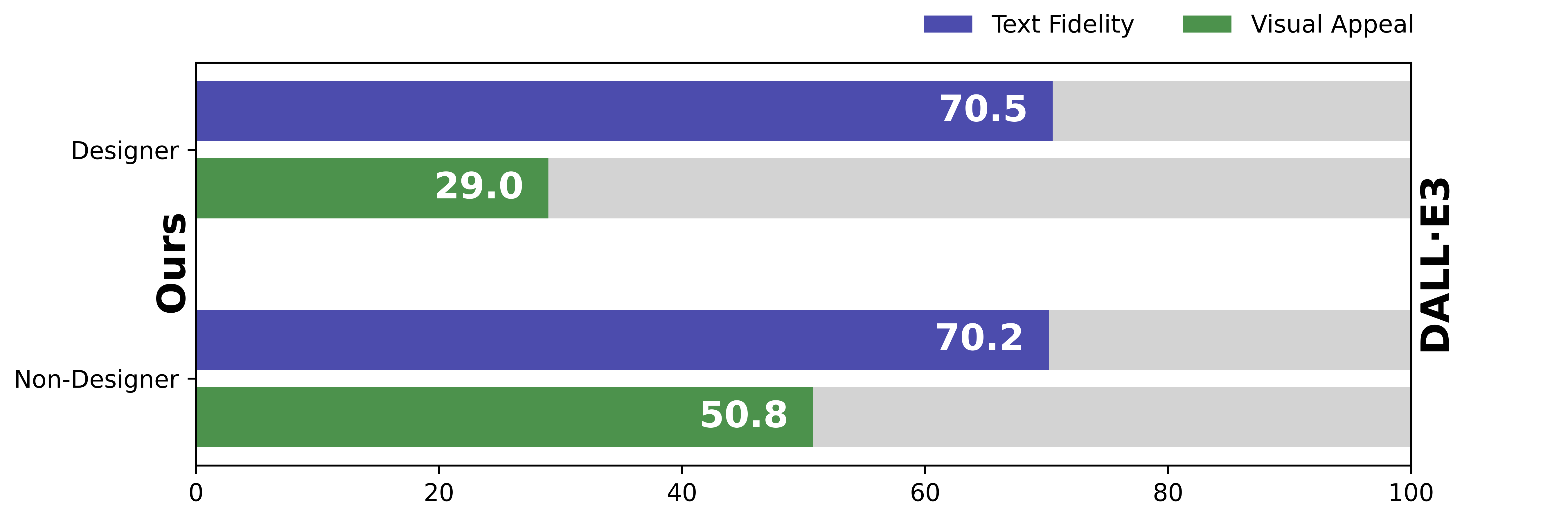}}
\end{minipage}
\caption{\small{Comparison with \dalle and CanvaGPT based on user study.}}
\label{fig:user_study}
\end{figure}

\begin{table*}[!t]
\begin{minipage}[t]{1\linewidth}
\centering
\tablestyle{10pt}{1.1}
\resizebox{1.0\linewidth}{!}
{
\begin{tabular}{l|c|c|c|c|ccccc}
\multirow{2}{*}{Method} & \multirow{2}{*}{Accuracy visual text} & \multirow{2}{*}{Multi-layered output} & \multirow{2}{*}{Flexible editing} & \multirow{2}{*}{Generated design} &  \multicolumn{5}{c}{GPT4-V Score} \\\cline{6-10}
& & & & & Layout & Image & Typography & Content & Innovation \\
\shline
DeepFloyd/IF~\cite{Deepfloyd} & \xmark & \xmark & \xmark & \cmark & 65.31 & 56.59 & 55.32 & 83.69 & 78.23 \\
SDXL~\cite{podell2023sdxl} & \xmark & \xmark & \xmark & \cmark & 60.89 &  60.12& 61.21& 67.33 & 69.92\\
\dalle~\cite{dalle3paper} & \xmark & \xmark & \xmark & \cmark & 80.29& 69.77 & 74.73 & \textbf{89.16} & \textbf{90.61}\\
CanvaGPT~\cite{canvagpt} & \cmark & \cmark & \cmark & \xmark  & 75.66 & 57.01 & 75.35 & 70.10 & 63.27\\
\rowcolor{blue!15} Ours \ourname & \cmark & \cmark & \cmark & \cmark & \textbf{80.82} & \textbf{72.53} & \textbf{75.46} & 87.44 & 88.72\\
\end{tabular}
}
\caption{
\small{Comparing our approach with the existing text-to-image generation models.}}
\label{tab:approach_comparison}
\end{minipage}
\end{table*}

Our objective in this work is to establish an efficient and dependable autonomous text-to-design system capable of generating high-quality graphic design images from user intent prompts.  
We suggest breaking down the complex process of graphic design image generation into a hierarchical generation process. This process involves multiple specialized generation models, each designed to handle different sub-tasks:
First, we train a Design-LLM to \emph{comprehend vague intentions and undertake design-oriented planning tasks}. We formulate this as a text-to-JSON prediction task, which requires transforming naive and sparse user intentions into structured, layer-wise captions or attributes essential for the following task decomposition.
Second, to perform \emph{design-oriented visual reasoning and generation}, we train additional three models including
(i) a text-to-background diffusion model for background layer generation: it is crucial to guarantee that the generated background images reserve adequate space for the incorporation of various elements, which is different from conventional methods such as \dalle, which often produce objects filling the entire image space.
(ii) a text-to-object diffusion model for object layer generation, it's essential to ensure the model to understand both the content and style of the background image. Subsequently, the model should generate an object layer that not only aligns stylistically but is also placed in logical positions.
(iii) a Typography-LMM for typography prediction, it needs to predict $15$ distinct typography attributes for each text box. This component requires a sophisticated level of reasoning about the visual contents of the merged background and object image layers, then predicting the typography attributes accordingly.
We also demonstrate the generalization capability of our Typography-LLM by applying it to processed design images originally generated using \dalle. Refer to Figure~\ref{fig:teaser} for the visual results.
Besides, we also introduce a \emph{feedback and quality assurance} process to enhance the quality.

Last, to access the capability of our system, we construct a \ourbenchmark, which consists of $\sim500$ professional graphic design intention prompts covering diverse categories and $\sim50$ creative ones, to compare our method with existing state-of-the-art image generation system, conduct thorough ablation experiments for each generation model on different sub-tasks, offer a detailed analysis of the graphic designs created by our system, and engage in a discussion about both the limitations and prospective future pathways of graphic design image generation. Figure~\ref{fig:user_study} shows that, among both non-designers and designers, our \ourname system outperforms \dalle in text fidelity and message conveyance, achieving win rates of 69.6\% and 70.5\%, respectively. Regarding visual appeal, non-designers show a slight preference for \ourname, while designers favor \dalle. Our results are also preferred by designers over those from the latest commercial product, CanvaGPT.
Table~\ref{tab:approach_comparison} provides a detailed comparison highlighting the key differences between our approach, earlier text-to-image generation models, and the latest CanvaGPT. Additionally, we present GPT4-V scores from various perspectives, with our \ourname outperforming others in three critical areas: layout, image, and typography quality.
\section{Related Work}
\label{sec:relate_work}

\noindent\textbf{Autonomous Agent based on LLMs}
Recent studies~\cite{yang2023idea2img,minidalle3,wu2023visual,yang2023mmreact,shen2023hugginggpt,hong2023metagpt,qian2023communicative,wu2023autogen,chen2023agentverse,wu2023autogen,li2023camel} have proposed the development of advanced (multi-)agent systems leveraging powerful LLMs to tackle various tasks. Some of them~\cite{yang2023idea2img,minidalle3,wu2023visual,yang2023mmreact,shen2023hugginggpt} concentrate more on image generation tasks.
Our model primarily distinguishes itself from these works in the mechanism of task decomposition and the usage of tools. While most existing studies incorporate tool usage knowledge by designing tailored prompt instructions or in-context examples for GPT-$3$/$4$, or \gptv to iteratively refine the text prompts, our \ourname decomposes the complicated graphic design image generation task into a hierarchical set of generation tasks, and performs dedicated fine-tuning of each generation model on our constructed professional graphic design dataset.

\noindent\textbf{Design Image Generation}
The majority of studies~\cite{hsu2023densitylayout,lin2023autoposter,chai2023two,zhang2023layoutdiffusion,shimoda2023towards,wei2023boosting,inoue2023towards,yamaguchi2021canvasvae} primarily concentrate on individual aspects within the graphic design generation task, assuming that all other information is at hand.
Unlike the these methods, this study aims to develop an autonomous, comprehensive system capable of executing all design generation tasks from clear user intent descriptions. This task formulation presents a significantly greater challenge compared to most existing approaches. To the best of our knowledge, our \ourname is among the pioneering works that seek to tackle the complex task of autonomous graphic design generation based solely on a basic user intent prompt.

\section{Our Approach}
\label{sec:approach}

\subsection{\textbf{\ourname} Framework Overview}

Motivated by the necessity to generate high-quality graphic designs that accurately reflect user intentions, we recognize the importance of a seamless integration of multiple roles. Each of these roles is designed to address specific aspects of the graphic design process, as indicated in the references~\cite{wiki_graphic_design,vise2021interdisciplinary}. Consequently, we train an LLM for multi-layered task planning, breaking down the text-to-design generation into a hierarchical structure of tasks. These include the intention-to-JSON generation, text-to-background image generation, text-to-object image and alpha mask generation, typography attribute generation, and the multi-layered graphic editing and rendering tasks.
The tasks of intention-to-JSON and typography attribute generation both adhere to a hierarchical structure, requiring the prediction of missing values based on a predefined structured request.

\begin{figure*}[t]
\begin{minipage}[t]{1\linewidth}
\begin{subfigure}[b]{1\textwidth}
\centering
\vspace{3mm}
\includegraphics[width=1\textwidth]{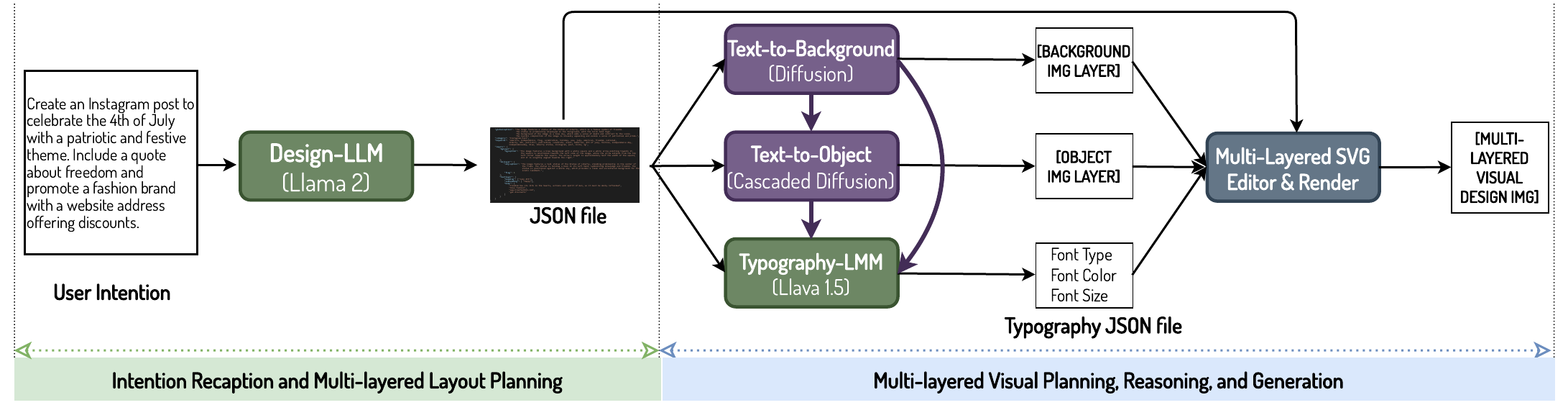}
\vspace{-3mm}
\end{subfigure}
\end{minipage}
\caption{\small{
Illustrating the detailed hierarchical pipelines of the proposed \ourname system.
Upon receiving a user's intention, our initial step involves using a Design-LLM to translate the intention into a detailed JSON file. This process necessitates multi-layered layout planning by predicting a wide range of attributes for the required visual elements.
Next, we engage a pair of cascaded diffusion models for the text-to-background generation and text-to-object (and alpha mask) generation processes. These models play a crucial role in creating visual assets, guided not only by the text instructions specified in the JSON file but also by the need to reason about their visual spatial relationships to ensure a coherent design.
Additionally, we have developed a typography-LMM that predicts the typography JSON file by analyzing and reasoning about the previously predicted text contents, background image, and object image.
Last, we apply a multi-layered SVG editor and rendering system to enable flexible user modifications on individual layers, allowing for the composition and output of the final image.
}}
\label{fig:framework}
\vspace{-5mm}
\end{figure*}

We introduce the comprehensive hierarchical framework of our \ourname system, as depicted in Figure~\ref{fig:framework}. To effectively train the \ourname system, we initially amass approximately 100,000 high-quality raw graphic design images from the internet, covering multiple graphic design sources. Following this, we organize the graphic design data into a hierarchical structure that includes components such as (intention, JSON) pairs, (caption, background-image) pairs, (caption, object-image) pairs, and (text-image, typography JSON) pairs. We will further explore the process of constructing each component in the ensuing discussion.

\subsection{Design LLM: Intention Recaption and Layout Planning}

We first detail the process of generating (intention, JSON) pair data for graphic design and then demonstrate the implementation of the Design LLM, aimed at intention recapitulation and planning multi-layered layouts.

\vspace{2mm}
\noindent\textbf{Creating Intention for Design Image} Given that both intentions and detailed captions are absent in the raw graphic design data, we first employ GPT-$3.5$ to generate the intentions from the raw image information. This includes the image title, format, keywords, and all visual text present in the image. This is accomplished with the aid of the system prompt displayed in the ``Intention Generation Prompt'' (shown in Supplementary). The left of Figure~\ref{fig:intention2json} presents a representative example, showcasing the ground-truth pairs of graphic design images and the generated user intentions. We generate the intention for nearly $100,000$ graphic design images.

\begin{figure}[t]
\centering
\begin{minipage}{.475\textwidth}
\centering
\includegraphics[width=\textwidth]{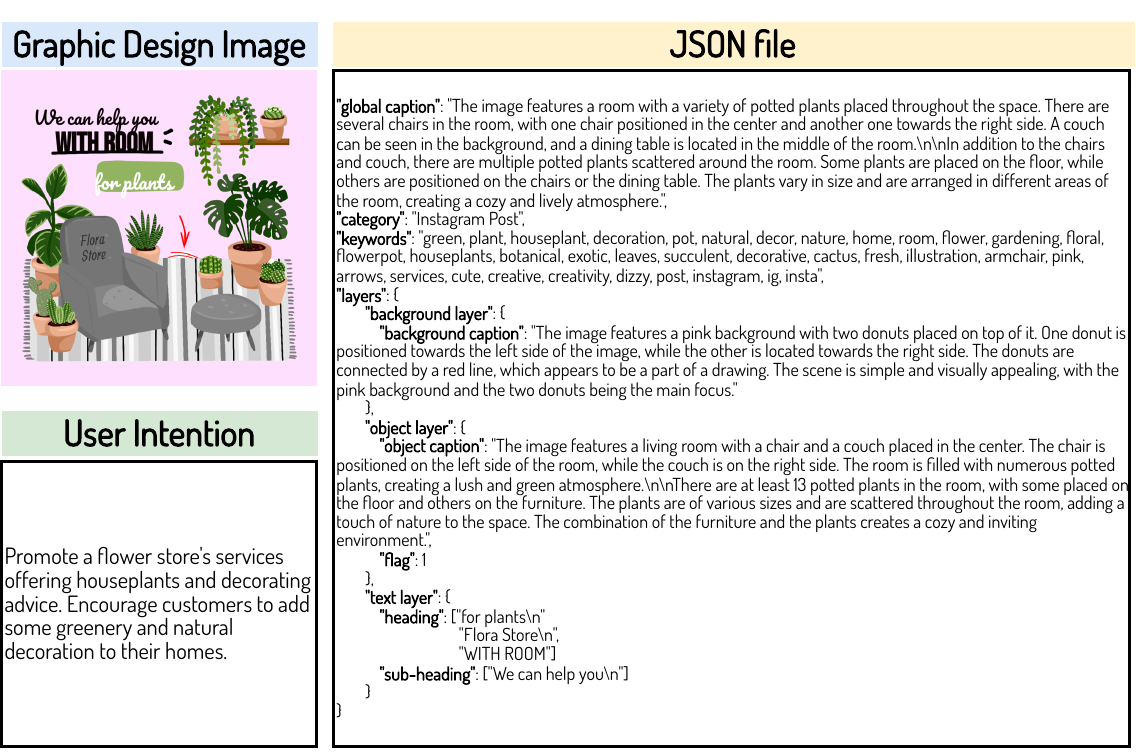}
\end{minipage}%
\caption{\small{Illustrating the example of generated intention-to-JSON pairs data for the given image. We can see that the user intention is vague and the JSON file is much more informative. Readers are kindly suggested to zoom into the figure for a clearer view. The image originates from our training dataset.}}
\label{fig:intention2json}
\end{figure}

\noindent\textbf{Creating JSON for Design Image}
Subsequently, we prepare the structured JSON data to represent the design images, which include multiple key-value pairs such as \emph{`global caption'}, \emph{`category'}, \emph{`keywords'}, and \emph{`layers'}. Within \emph{`layers'}, we further decompose it into three sub-layers: \emph{`background layer'}, \emph{`object layer'}, and \emph{`text layer'}. Considering that not all design images contain the object layer, we also use a \emph{`flag'} to mark whether the object layer exist. For all captions, we opt to utilize the recently developed large multimodal model known as the LLaVA-1.5-13B model~\cite{liu2023improved}, which is effective in generating detailed captions for the entire image, including background and object image captions.
The right side of Figure~\ref{fig:intention2json} presents a representative example, showcasing the generated JSON file for the given image. We create matching JSON files for nearly $100,000$ graphic design images, each paired with the above generated intentions.

\vspace{2mm}
\noindent\textbf{Implementation and Training Settings} After constructing the intention and JSON pairs for the graphic design images, we formulate the intention-to-JSON prediction task as a masked field prediction problem following~\cite{inoue2023document}, which is very different from the conventional text-to-text prediction task. This approach learns to translate intentions into detailed JSON files and conducts multi-layered layout planning by predicting the global captions, background layer captions, object layer captions, contents of the text layers including the heading, sub-heading, and body-text. Furthermore, we adopt a strategy from~\cite{aghajanyan2022cm3} to construct causally masked multi-modal LLMs by leveraging the Llama$2$-$13$B~\cite{touvron2023llama}.
We have fine-tuned the causally masked Llama$2$-$13$B model using our assembled dataset of nearly $100,000$ (intention, JSON) pairs, over an approximate span of $\sim10$ epochs. Our empirical findings suggest that incorporating additional predictions of the \emph{`global caption'}, along with \emph{`keywords'} from the graphic design image, facilitates subsequent tasks of background and object image generation.

\subsection{Text-to-Background Diffusion Model: Visual Planning to Generate Canvas Placeholders}

We now present the methodology for preparing (caption, background-image) pairs and delve into the complexities of training a dependable background image to serve as a canvas placeholder. This involves strategically planning about where to allocate sufficient space for the seamless integration of diverse elements.

\begin{figure}[t]
\centering
\begin{minipage}{.45\textwidth}
\centering
\includegraphics[width=\textwidth]{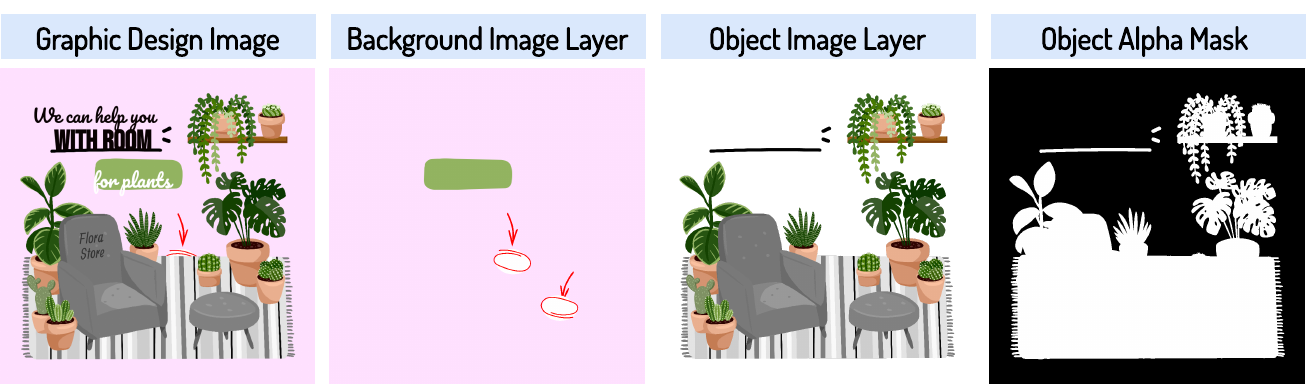}
\end{minipage}%
\caption{\small{Illustrating the decomposition of the graphic design images into the combination of background image layers, object image layers, and object alpha masks.}}
\label{fig:background_object_layer}
\end{figure}

\noindent\textbf{Preparing Text-Background Pairs}
Given the current unavailability of mature text-to-SVG technology and the scarcity of high-quality text-SVG pairs, which leads to the absence of strong text-to-SVG generation models~\cite{jain2023vectorfusion,zhang2023text,wu2023iconshop,xing2023diffsketcher}, our study suggests consolidating all SVG elements and additional embellishments into one unified background image layer.
Then we extract the generated detailed description of the unified background image layer from the JSON file.
The left side of Figure~\ref{fig:background_object_layer} displays an example of the decomposed background image layer from a given graphic design image.
We construct less than $\sim100,000$ (caption, background image) pairs for use as training data.

\noindent\textbf{Implementation and Training Settings} Then we train a text-to-background diffusion model to generate the coherent background images consisting of all the SVG files and decoration elements.
We opt to fine-tune our collected (caption, background image) pairs using the DeepFloyd IF~\cite{Deepfloyd} diffusion models for $\sim10$ epochs. The output resolution is $1,024$$\times$$1,024$. Following this, we employ the generated background images as conditioning input for the text-to-object diffusion model, which predicts coherent high-resolution object image layers based on their corresponding object prompts.

One key challenge in generating the background image lies in ensuring it serves effectively as a canvas placeholder. This requires reserving adequate space for incorporating various elements, including object image layers and text layers. Consequently, the text-to-background generation model must perform visual planning to determine where to allocate space. This approach significantly diverges from conventional methods like \dalle, which often generate objects occupying the entire image space. We provide a qualitative comparison of the background images generated by different methods in the appendix. Another observed challenge is the length of the generated detailed captions, which can extend to tens of words, resulting in hundreds of text tokens - far exceeding the original default maximum length. To overcome this, we have increased the maximum number of text tokens from $77$ to $512$.

\noindent\textbf{Offset-noise is important for background image generation} We observe the graphic design background images tend to have a uniform color distribution and this aspect sets them apart from the pre-trained distributions of photorealistic images. Our empirical findings emphasize the effectiveness of the offset noise technique~\cite{lin2023common,offsetnoise} for high-quality background image generation. This is because, even when $t = T$, the model's input isn't purely noise. The leaked signal carries low-frequency information, such as the overall mean of each channel. The implementation details are as follows:
$\mathsf{noise = torch.randn\_like(latents) + \alpha * torch.randn(}$
$\mathsf{latents.shape[0], latents.shape[1], 1, 1)}$.
We choose $\alpha=0.1$ by default.

\subsection{Text-to-Object Diffusion Model: Visual Reasoning based on the Generated Background Image}

This section outlines the method for creating (caption, object image) pair data and showcases the deployment of the text-to-object diffusion model. This model is designed to produce an object layer that is not only stylistically coherent but also logically positioned. A key technical insight in this endeavor is the model's need to predict an additional composite image, a step that is essential for ensuring the precision of the object placements.

\noindent\textbf{Preparing Text-Object-Background-AlphaMask Quadruples}
To train a text-to-object diffusion model capable of perceiving both the object captions present in the JSON file and the visual pixels predicted by the text-to-background diffusion models, we prepare the (object-caption, background-image, object-image, alpha-mask) quadruples. We extract the object-image and the alpha-mask from the raw data, while the object-caption and the background-image are prepared following the above-mentioned scheme.

\noindent\textbf{Implementation and Training Settings}
Following the instruct-P2P approach~\cite{brooks2023instructpix2pix}, we concatenate the noise latent vectors with the latent vectors of the conditioning background image with the alpha mask for the object layer, and we add additional input channels to the first convolutional layer. All available weights of the diffusion model are initialized from pre-trained checkpoints, while weights operating on the newly added input channels are initialized to zero.
As a result, we can predict both the object image and its alpha mask layer, which enables us to control the transparency of the object pixels. Consequently, we can seamlessly blend the object image layer with the background image layer in accordance with the alpha mask.
To ensure a more reasonable placement of object pixels and coherence in style with the background image, we propose predicting an additional composed image. This is achieved by adding three extra channels to both the input noise vectors and the output prediction, a step that proves critical for accurate object layer prediction. Therefore, our text-to-object diffusion model needs to predict $\sim7$ channels, where the object image consists of $\sim3$ channels, the object mask consists of $1$ channel, and the composed image consists of $\sim3$ channels. Besides, we also increased the maximum number of text tokens from $77$ to $512$ following the text-to-background diffusion model.

Similar to the text-to-background generation model, we optimize our assembled dataset, comprising approximately $\sim55,000$ (object-caption, background-image, object-image, alpha-mask) quadruples. The relatively smaller number of quadruples is due to the frequent absence of the object image layer in many graphic design images. We employ the instruct-P2P scheme for optimization, which is based on the DeepFloyd~\cite{Deepfloyd} IF diffusion models, and this process is carried out over an estimated span of $\sim20$ epochs.

\noindent\textbf{Qualitative Examples}
Figure~\ref{fig:background_object_layer} provides a selection of representative examples that demonstrate our method of decomposing complex graphic design images into distinct layers of background and object images.
It's important to note that predicting the object alpha mask is essential for facilitating flexible editing operations on the object pixels.

\subsection{Typography LMM: Layout Planning and Attribute Reasoning for Visual Text}

Typography~\cite{wiki_typography} is the art and technique of arranging type to make written language legible, clear, and visually engaging for the reader. It therefore requires complex reasoning and planning capabilities. It encompasses the design of font style, appearance, and structure, with the goal of eliciting specific emotions and conveying certain messages. In essence, typography brings text to life.
Contrary to recent efforts, such as~\cite{tanveer2023ds}, \cite{iluz2023word}, and \cite{he2023wordart}, which primarily employ diffusion models to generate novel letter-level or word-level semantic or artistic typographies, our focus is more directed towards the utilization of an LMM (Large Multimodal Model). This approach allows us to address the highly challenging task of image-level typography. This task encompasses selecting font types, establishing hierarchical arrangements, and adjusting leading (line spacing), tracking (character spacing), and kerning (space between specific characters) based on the composed images. Consequently, it demands a robust capability for layout planning and attribute reasoning from the Typography LMM.

\begin{figure}[t]
\centering
\begin{minipage}{1\linewidth}
\centering
\includegraphics[width=\textwidth]{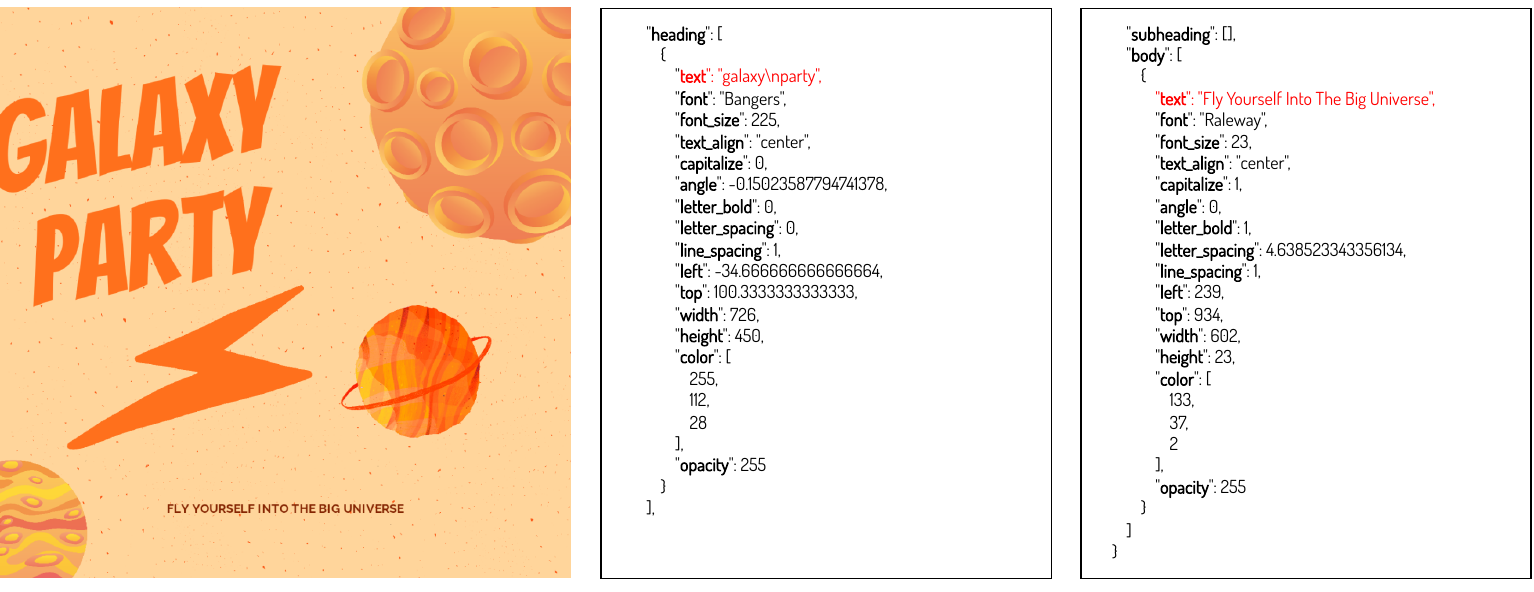}
\end{minipage}
\caption{\small{Illustrating the Typography JSON file. The text highlighted in red is predicted by the Design LLM. The image on the left depicts the effects of rendering visual text accordingly.}}
\label{fig:typography_example}
\end{figure}

\noindent\textbf{Preparing Composed Images and Typography Data} 
We construct nearly $100,000$ triplets of data, each consists of multiple texts, composed images of backgrounds and objects, and typography information.
Figure~\ref{fig:typography_example} displays a representative example of a typography JSON file, illustrating that it comprises $16$ attributes, with the value of `\emph{text}' already predicted by the Design-LLM. We extract all these typography attributes directly from the raw data of each graphic design image.

\noindent\textbf{Implementation and Training Settings}
In our experiments, we elected to fine-tune the LLaVA-$1.5$-$13$B model~\cite{liu2023improved} using nearly $100,000$ triplets for $\sim10$ epochs. It is evident that predicting visually appealing typography information for each text box, based on the input image, is a complex task. This is due to the requirement of predicting approximately $\sim15$ attributes for each text box. To our knowledge, even the most recent effort~\cite{shimoda2023towards} only manages to predict $\sim8$ typography attributes, assuming the other attributes are fixed.
To simplify the prediction process, we further categorize the continuous floating numbers into discrete bins as per the method outlined in~\cite{chen2021pix2seq} and we also utilize a unique token to represent the values that fall within the same bin.
We explain more details of the bin partition scheme in the supplementary material.

\subsection{Multi-Layered SVG Editor and Renderer: Support Layer-wise Flexible User Editing}
\begin{figure}[t]
\centering
\begin{minipage}{1\linewidth}
\centering
\includegraphics[width=\textwidth]{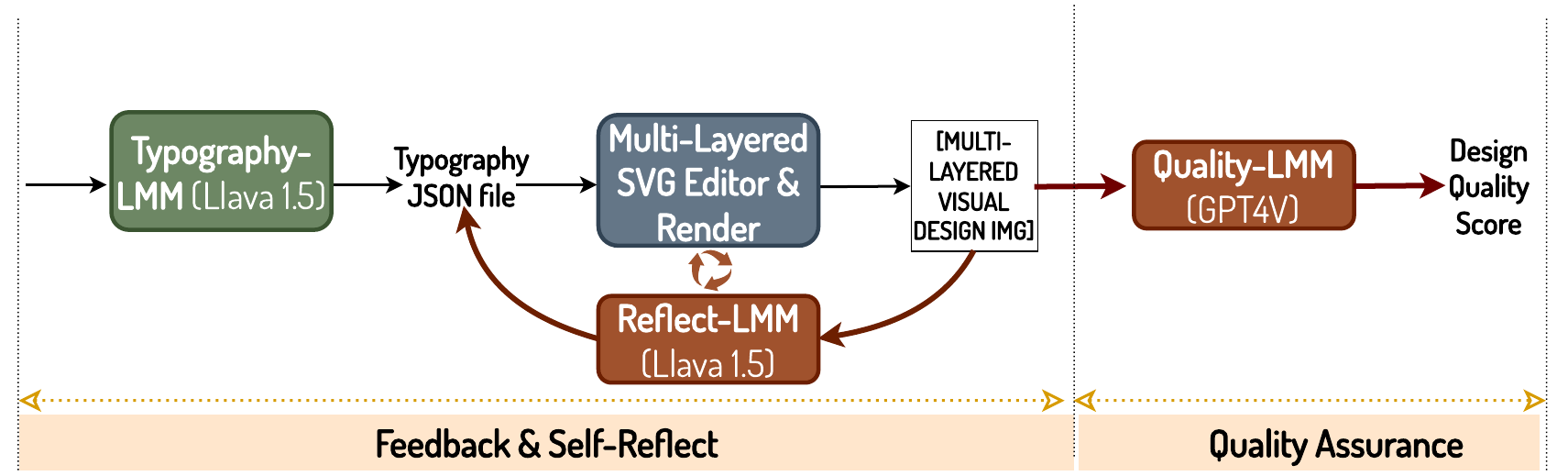}
\end{minipage}%
\caption{\small{Illustrating the stage of feedback and quality assurance in our \ourname system.}}
\label{fig:reflect_framework}
\begin{minipage}[t]{1\linewidth}
\vspace{2mm}
\centering
\begin{subfigure}[b]{0.24\linewidth}
{\includegraphics[width=\textwidth]{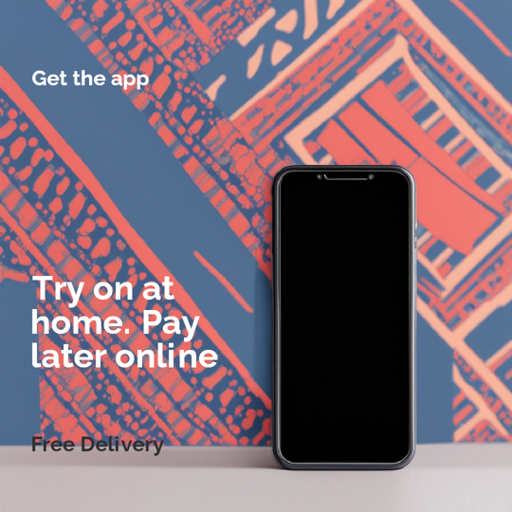}}
\vspace{-4mm}
\end{subfigure}
\begin{subfigure}[b]{0.24\linewidth}
{\includegraphics[width=\textwidth]{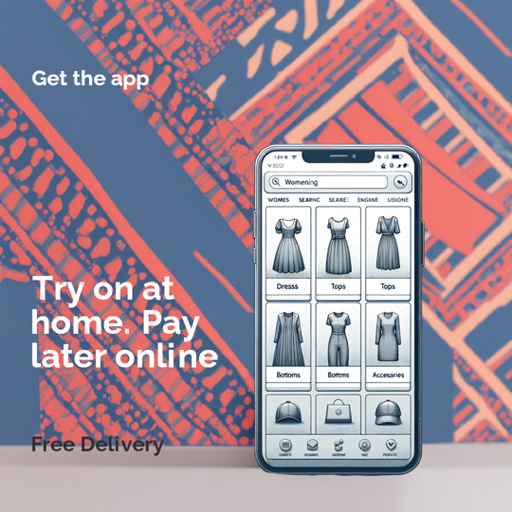}}
\vspace{-4mm}
\end{subfigure}
\begin{subfigure}[b]{0.24\linewidth}
{\includegraphics[width=\textwidth]{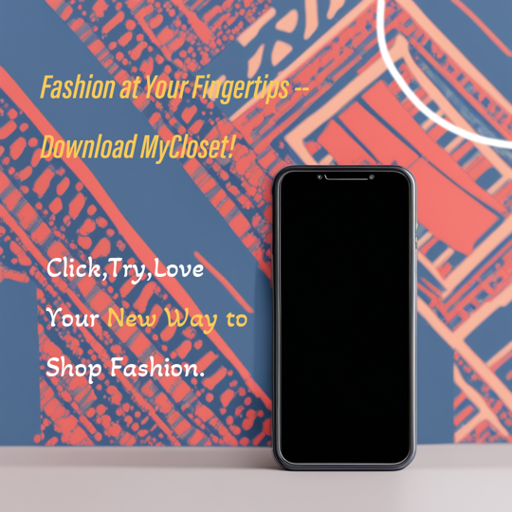}}
\vspace{-4mm}
\end{subfigure}
\begin{subfigure}[b]{0.24\linewidth}
{\includegraphics[width=\textwidth]{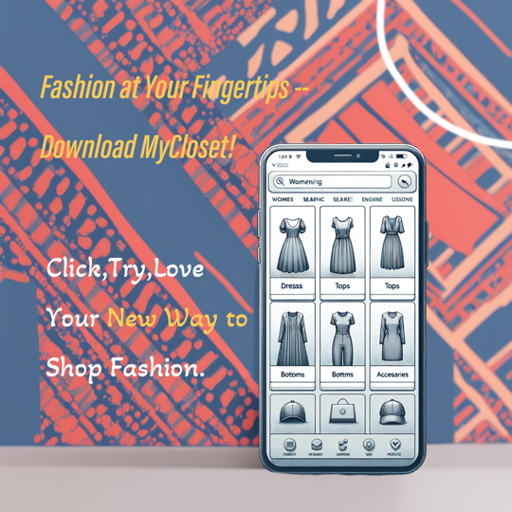}}
\vspace{-4mm}
\end{subfigure}
\end{minipage}
\begin{minipage}{1\linewidth}
\centering
\begin{subfigure}[b]{0.24\linewidth}
{\includegraphics[width=\textwidth]{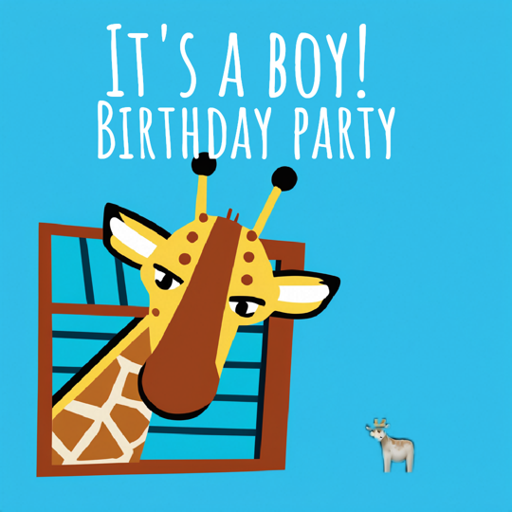}}
\vspace{-4mm}
\end{subfigure}
\begin{subfigure}[b]{0.24\linewidth}
{\includegraphics[width=\textwidth]{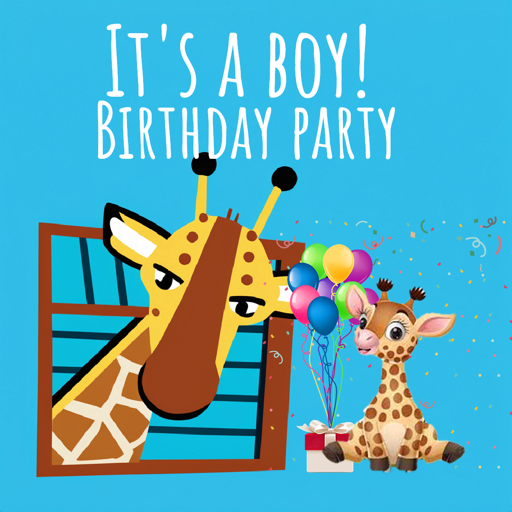}}
\vspace{-4mm}
\end{subfigure}
\begin{subfigure}[b]{0.24\linewidth}
{\includegraphics[width=\textwidth]{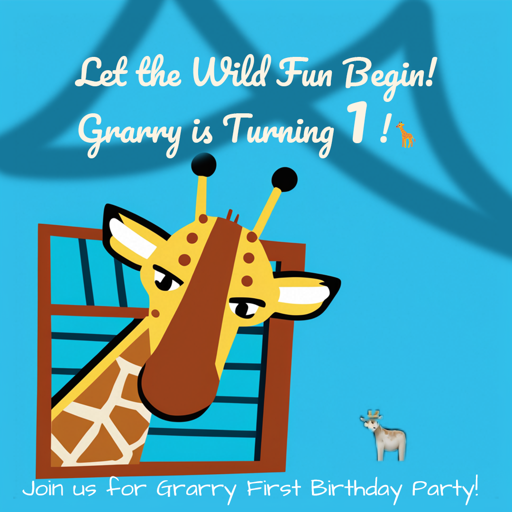}}
\vspace{-4mm}
\end{subfigure}
\begin{subfigure}[b]{0.24\linewidth}
{\includegraphics[width=\textwidth]{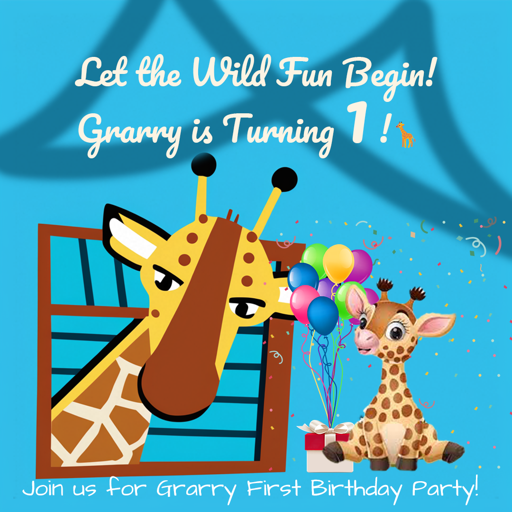}}
\vspace{-4mm}
\end{subfigure}
\caption{\small{Examples of flexible user editing. Left to right: original image (1-st, 5-th), object editing (2-ed, 6-th), text editing (3-rd, 7-th), and combined editing (4-th, 8-th).}}
\label{fig:edit_example}
\end{minipage}
\end{figure}

After generating all the design image layers, we propose compiling them into a single SVG file using the open-source tool \texttt{SVG-Edit}\cite{svgedit}. Following this, we opt to utilize  \texttt{Justinmind}\footnote{https://www.justinmind.com/}, a tool that facilitates a wide range of flexible, layer-wise user editing operations. These operations include repositioning and resizing text boxes or object layers. Additionally, Justinmind allows for modifications to be made to font size, color, type, and other attributes within each text box, enhancing the user's ability to customize their designs further.
We present the individual image layers and the fully composed image in Figure~\ref{fig:teaser}. Additionally, we showcase a series of representative examples that demonstrate the outcomes following various layer-wise editing operations in Figure~\ref{fig:edit_example}.

\subsection{Reflect LMM \& Quality LMM}
Considering that initial multi-layered graphic design outputs may not always align with user expectations, we further propose the integration of an added Reflect-LMM. This component is designed to critically analyze the generated design images and predict necessary adjustments to the JSON file, thereby further improving the resulting graphic designs.

\begin{table*}[htbp] 
\begin{minipage}[t]{1\linewidth}  
\centering  
\tablestyle{10pt}{0.5}  
\resizebox{1.0\linewidth}{!}  
{
\begin{tabular}{l|>{\centering\arraybackslash}m{20cm}}  
Category & Intention Examples \\  
\shline  
advertising & \scriptsize{Create a customer testimonial advertisement showcasing a pink clothing collection. The advertisement includes a positive review from Jenny Wilson. } \\ \hline
events & \scriptsize{Create a birthday card with a festive and colorful design featuring a cake with fruits. The card is an invitation to a birthday party happening on June 22nd from 8pm to 11pm at 123 st. Any city 4567.} \\  \hline
marketing & \scriptsize{Design a gift certificate for a travel insurance company that offers a 30\% discount for all kinds of travel.} \\  \hline
posts & \scriptsize{Create an Instagram story to promote a special offer of 30\% discount on the autumn collection. Encourage users to visit the website by including a link in bio.} \\  \hline
covers\& headers & \scriptsize{Create a thumbnail for a vlogger episode that features a collection of rare antique telephones. The title is 'Top 5 Antique Phones'} \\ \hline
creative & \scriptsize{Design an alluring poster for a blues concert in a clearing of the Amazon Rainforest. The artwork should evoke the soulful tunes of blues music echoing through the wilderness. Use a moody, midnight blue palette and incorporate silhouettes of blues instruments and nocturnal wildlife.
} \\ 
\end{tabular}  
}  
\caption{  
\small{Intention samples in \ourbenchmark benchmark.}}
\label{tab:designer_prompt}  
\end{minipage}
\end{table*}

\noindent\textbf{Reflect LMM} The entire self-reflection framework is illustrated on the left side of Figure~\ref{fig:reflect_framework}. As an integral part of our implementation plan, we fine-tune a LLaVA-$1.5$-$13$B model on a dataset that comprises more than $300,000$ artificially generated noisy JSON files. Each of these files is paired with a corresponding ground-truth JSON file. Our primary modification involves the addition of random shift noise to the position of each ground-truth typography block.
The key insight lies in modeling the conditional transformation of typography attributes from suboptimal to ideal, leveraging the understanding of the rendered composite images.

\noindent\textbf{Quality LMM}
We utilize GPT-4V(ision)~\cite{gpt4vcontribution} to evaluate the quality of the generated graphic design images, as displayed in the ``Quality Assurance Prompt for \gptv'' in the appendix.
Through empirical analysis, we have found that GPT-4V(ision) excels at evaluating the quality of $4$ main aspects such as \emph{design and layout}, \emph{content relevance}, \emph{graphics and images}, and \emph{innovation}. However, it falls short in assessing the quality of \emph{typography and color}.
In addition, we also incorporate the evaluation prompt design as introduced in a recent concurrent effort~\cite{lin2023designbench}. This evaluates the graphic design images from approximately $\sim8$ aspects, including text rendering quality, composition and layout, color harmony, cinematography, style, image-text alignment, aesthetics, and overall design.
\section{Experiment}

\label{sec:experiment}

{
\definecolor{myred}{rgb}{0.94,0.32,0.0.47}
\definecolor{myblue}{rgb}{0.17,0.33,0.79}
\definecolor{myyellow}{rgb}{0.99,0.81,0.43}
\definecolor{mygreen}{rgb}{0.27,0.74,0.61}
\begin{figure*}[t]
\footnotesize
\begin{minipage}[t]{1\linewidth}
\centering
\begin{subfigure}[b]{0.1925\textwidth}
{\includegraphics[width=\textwidth]{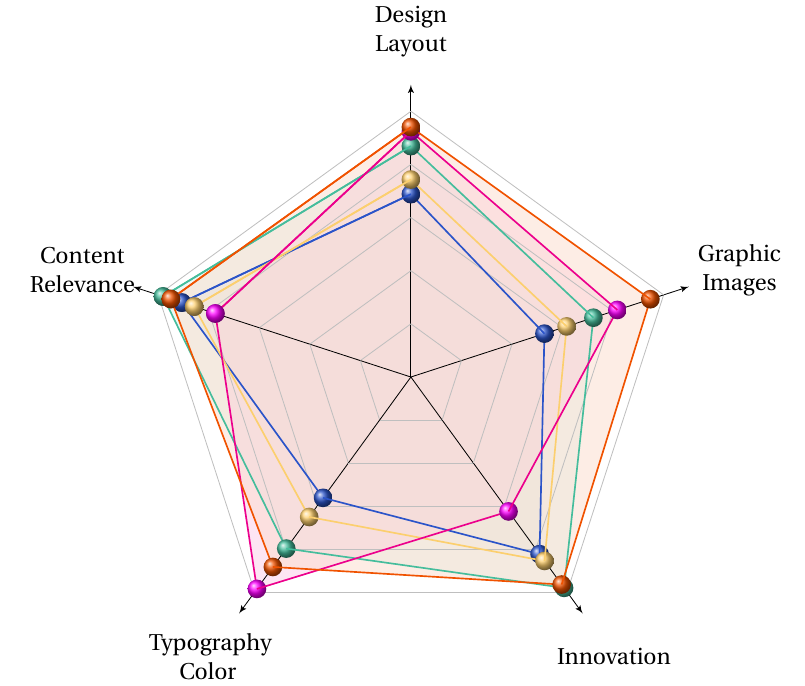}}
\caption{covers \& head.}
\end{subfigure}
\hfill
\begin{subfigure}[b]{0.1925\textwidth}
{\includegraphics[width=\textwidth]{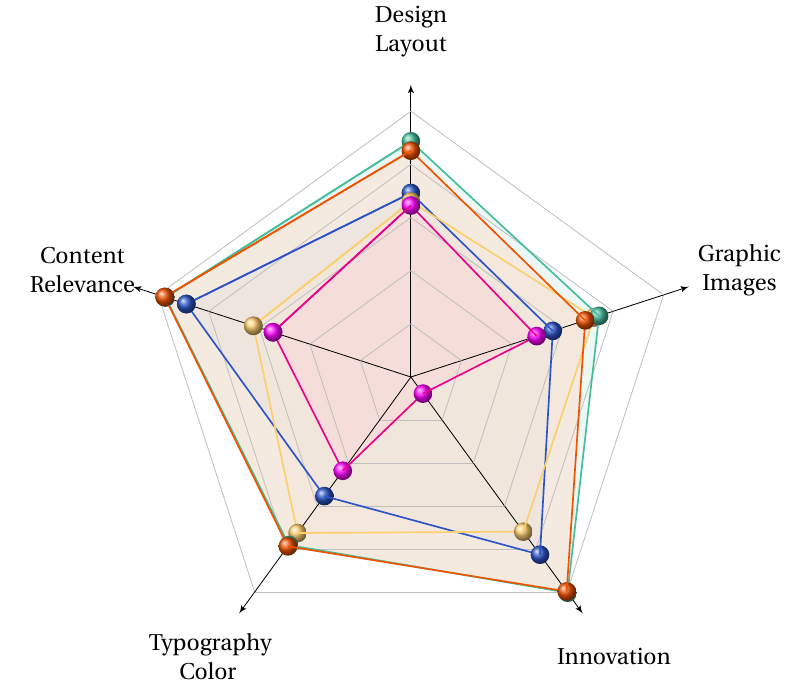}}
\caption{marketing}
\end{subfigure}
\hfill
\begin{subfigure}[b]{0.1925\textwidth}
{\includegraphics[width=\textwidth]{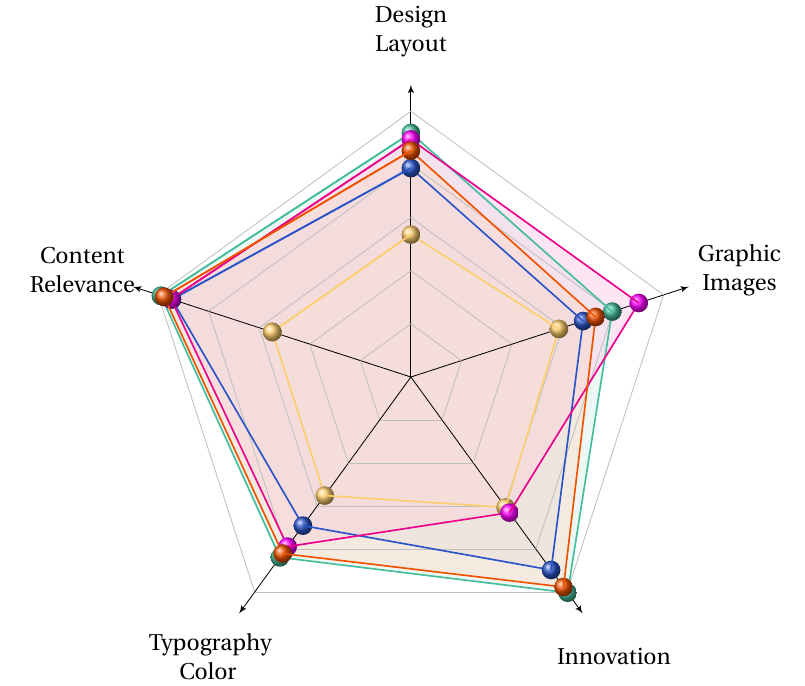}}
\caption{posts}
\end{subfigure}
\hfill
\begin{subfigure}[b]{0.1925\textwidth}
{\includegraphics[width=\textwidth]{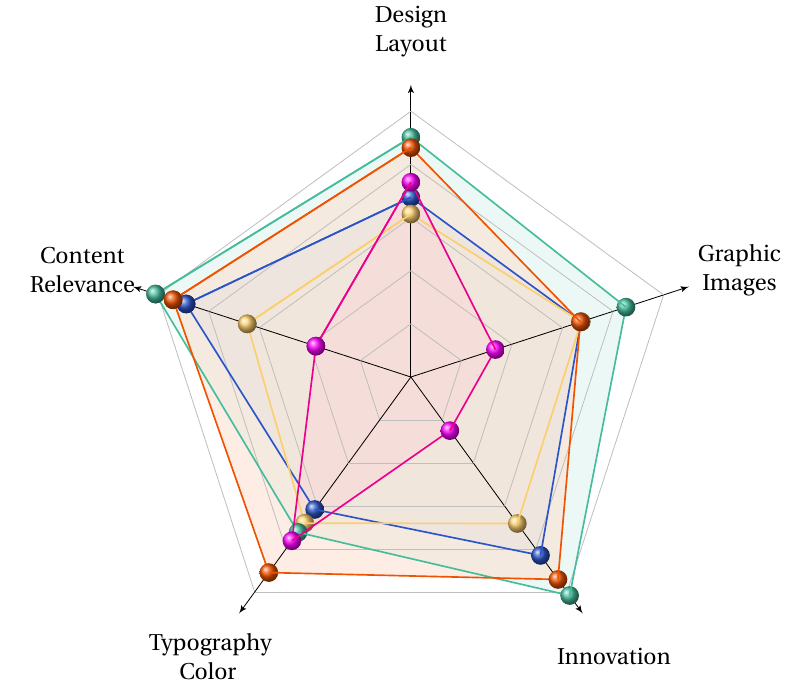}}
\caption{events}
\end{subfigure}
\hfill
\begin{subfigure}[b]{0.1925\textwidth}
{\includegraphics[width=\textwidth]{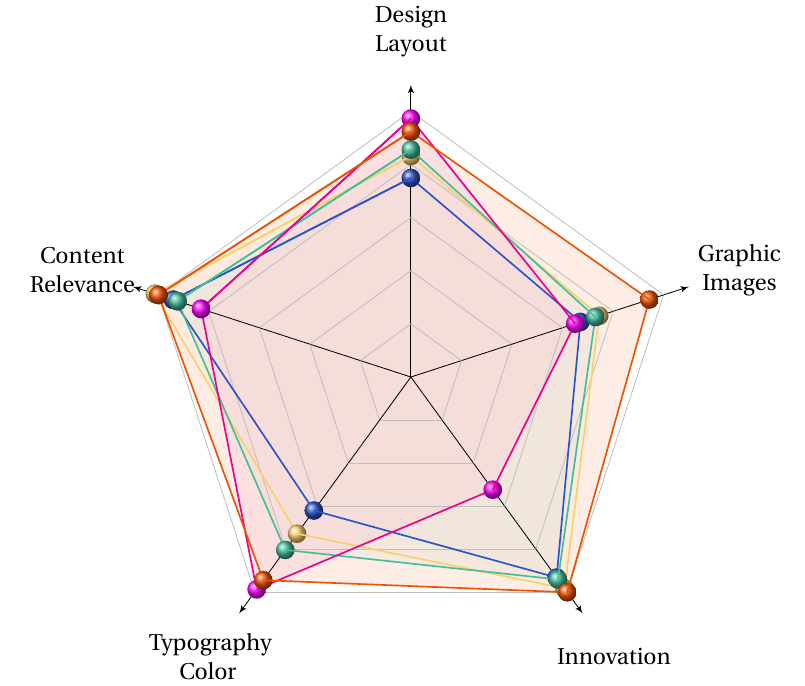}}
\caption{advertising}
\end{subfigure}
\caption{{Illustrating the \gptv scores. We mark the results of our \ourname ({\color{myred}{\ding{108}}}), DeepFloyd/IF+GPT-4 ({\color{myblue}{\ding{108}}}), SDXL+GPT-4 ({\color{myyellow}{\ding{108}}}), \dalle+GPT-4 ({\color{mygreen}{\ding{108}}}) and CanvaGPT ({\color{magenta}{\ding{108}}}), using colored markers. (See the appendix for the zoomed-out version)}}
\label{fig:gpt4v_radar_score}
\vspace{1mm}
\end{minipage}
\begin{minipage}[htbp]{1\linewidth}
\centering
\begin{subfigure}[b]{0.16\textwidth}
{\includegraphics[width=\textwidth]{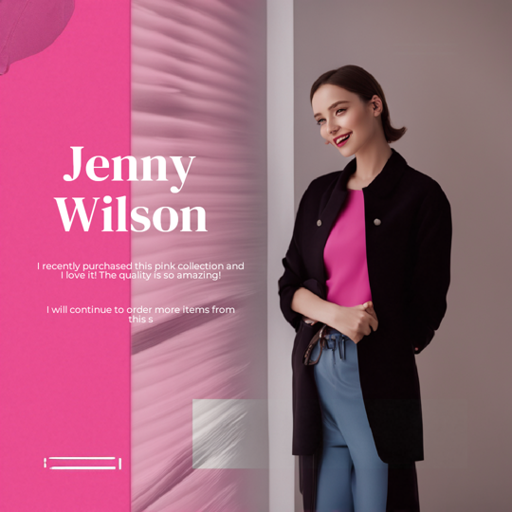}
}
\vspace{-4mm}
\end{subfigure}
\begin{subfigure}[b]{0.16\textwidth}
{\includegraphics[width=\textwidth]{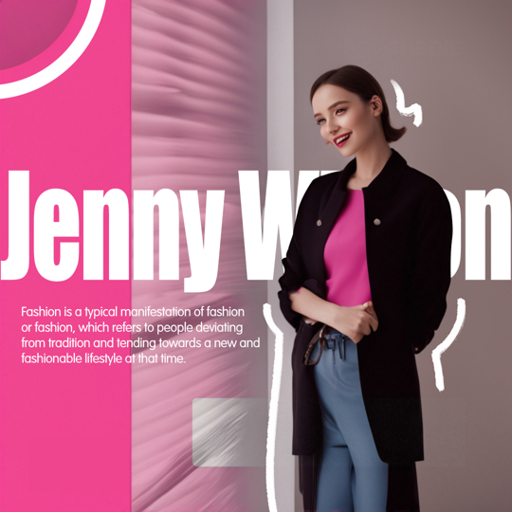}
}
\vspace{-4mm}
\end{subfigure}
\begin{subfigure}[b]{0.16\textwidth}
{\includegraphics[width=\textwidth]{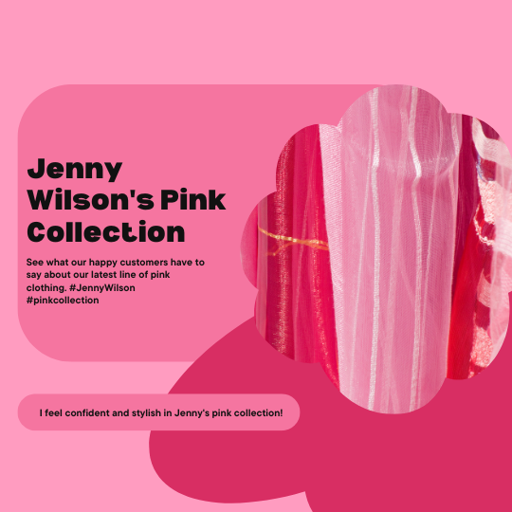}
}
\vspace{-4mm}
\end{subfigure}
\begin{subfigure}[b]{0.16\textwidth}
{\includegraphics[width=\textwidth]{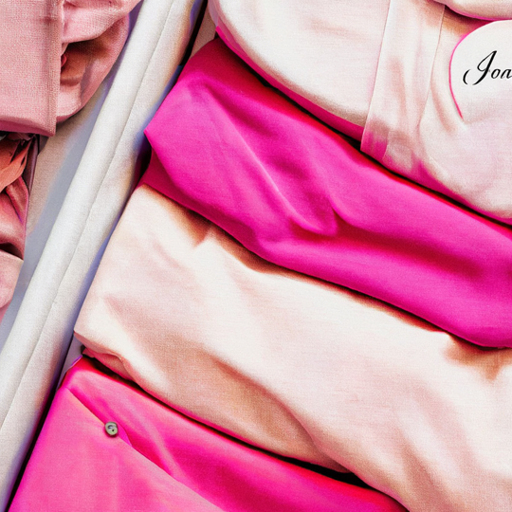}}
\vspace{-4mm}
\end{subfigure}
\begin{subfigure}[b]{0.16\textwidth}
{\includegraphics[width=\textwidth]{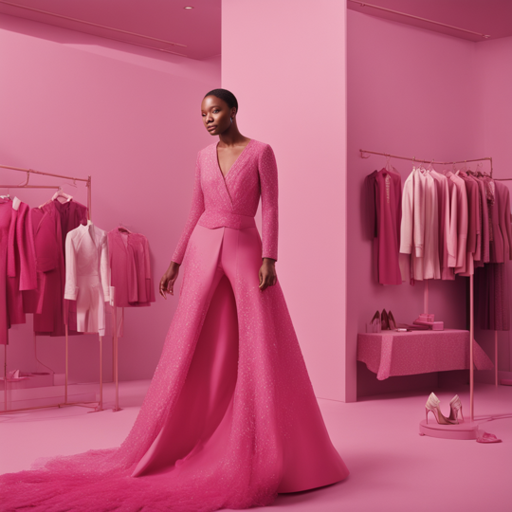}}
\vspace{-4mm}
\end{subfigure}
\begin{subfigure}[b]{0.16\textwidth}
{\includegraphics[width=\textwidth]{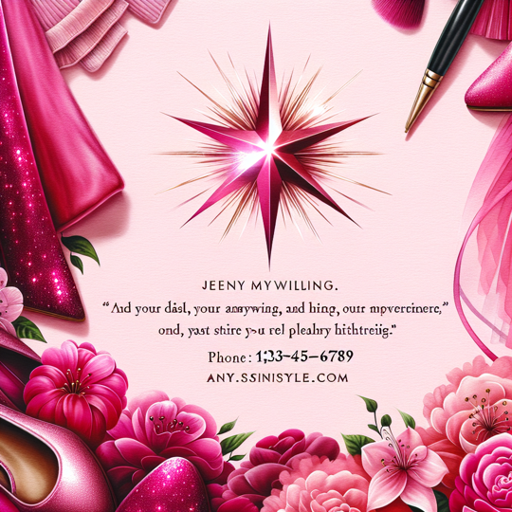}}
\vspace{-4mm}
\end{subfigure}\\
\begin{subfigure}[b]{0.16\textwidth}
{\includegraphics[width=\textwidth]{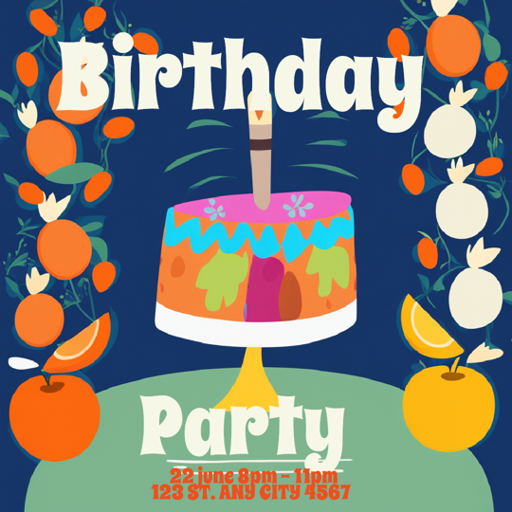}
}
\vspace{-4mm}
\end{subfigure}
\begin{subfigure}[b]{0.16\textwidth}
{\includegraphics[width=\textwidth]{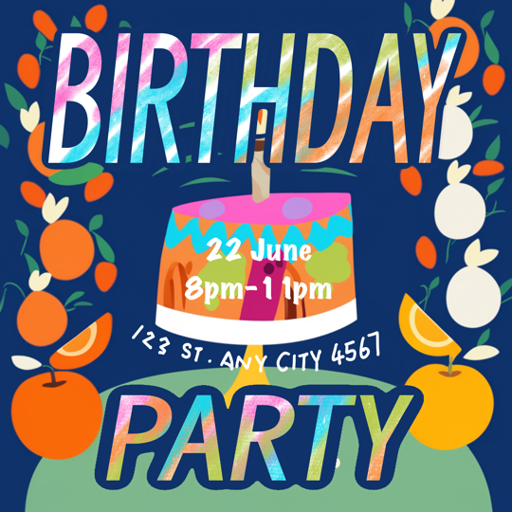}
}
\vspace{-4mm}
\end{subfigure}
\begin{subfigure}[b]{0.16\textwidth}
{\includegraphics[width=\textwidth]{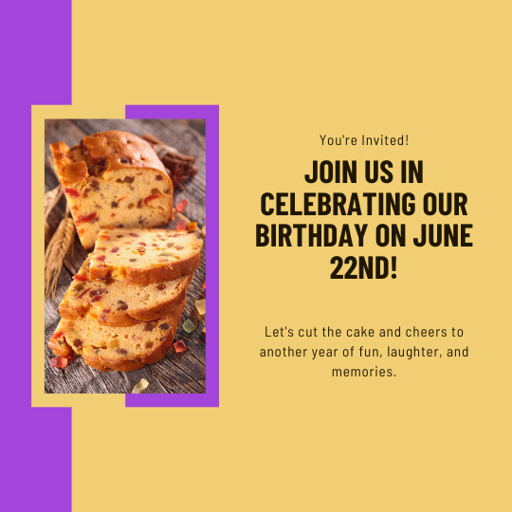}}
\vspace{-4mm}
\end{subfigure}
\begin{subfigure}[b]{0.16\textwidth}
{\includegraphics[width=\textwidth]{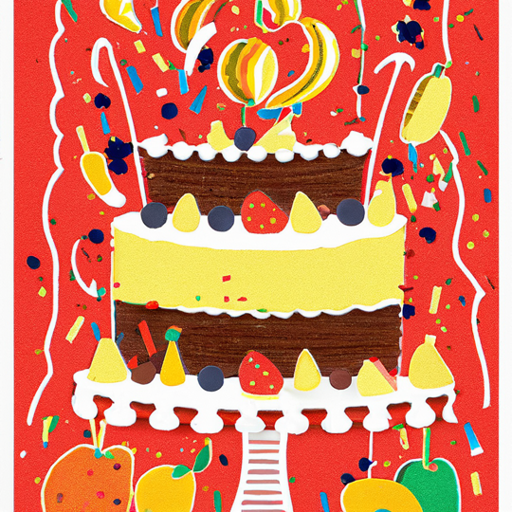}}
\vspace{-4mm}
\end{subfigure}
\begin{subfigure}[b]{0.16\textwidth}
{\includegraphics[width=\textwidth]{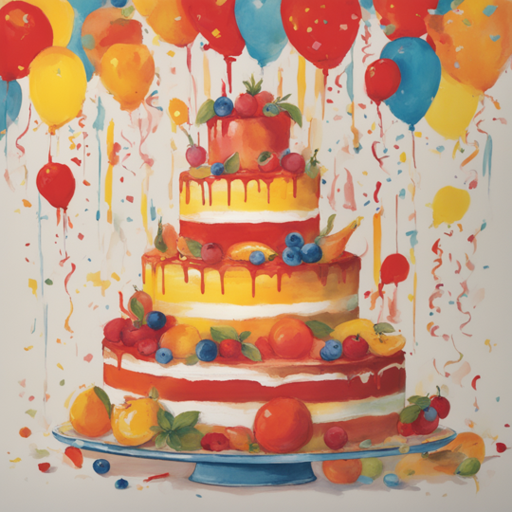}}
\vspace{-4mm}
\end{subfigure}
\begin{subfigure}[b]{0.16\textwidth}
{\includegraphics[width=\textwidth]{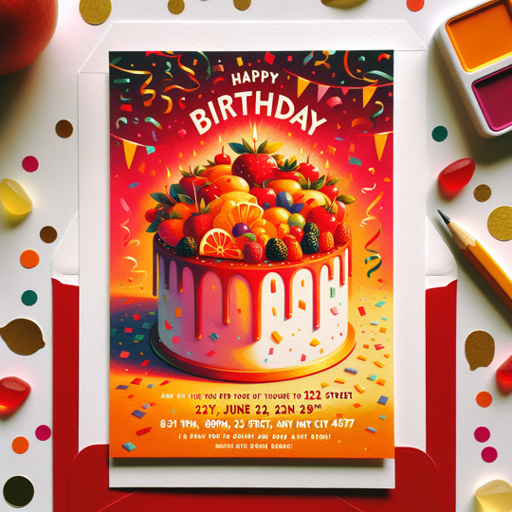}}
\vspace{-4mm}
\end{subfigure}\\
\hspace{0.14mm} 
\begin{subfigure}[b]{0.16\textwidth}
{\includegraphics[width=\textwidth]{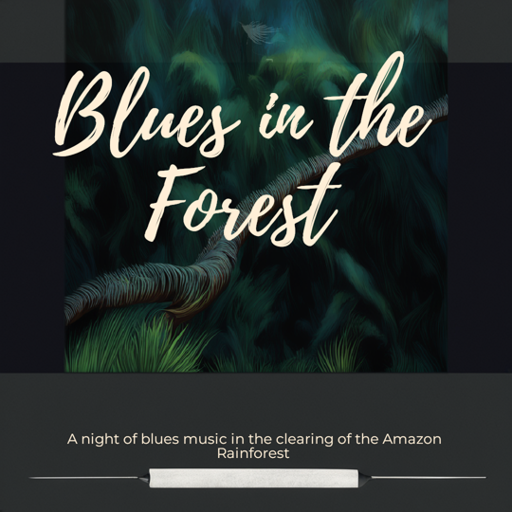}
}
\vspace{-4mm}
\caption*{\scriptsize{Ours}}
\end{subfigure}
\begin{subfigure}[b]{0.16\textwidth}
{\includegraphics[width=\textwidth]{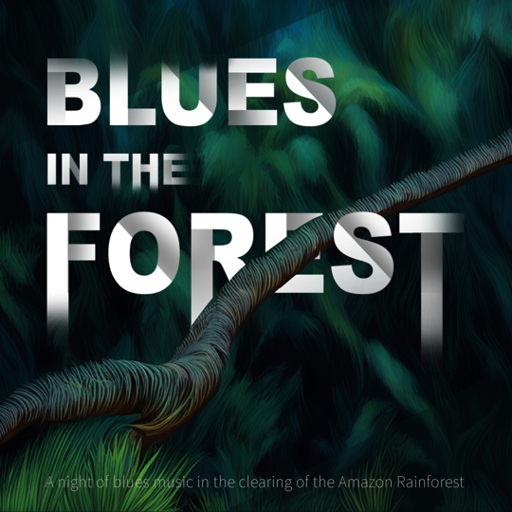}
}
\vspace{-4mm}
\caption*{\scriptsize{Ours+Edit}}
\end{subfigure}
\begin{subfigure}[b]{0.16\textwidth}
{\includegraphics[width=\textwidth]{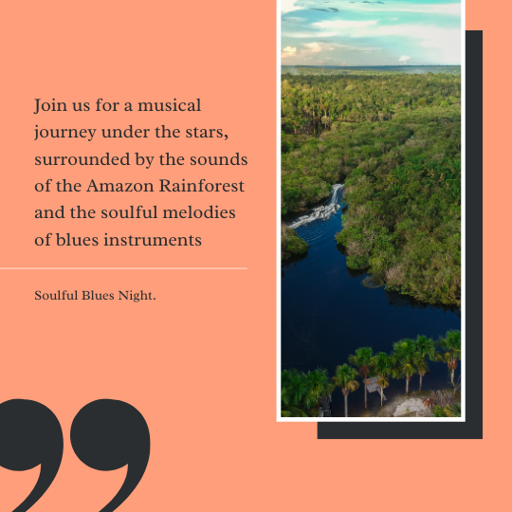}}
\vspace{-4mm}
\caption*{\scriptsize{CanvaGPT$^\dagger$}}
\end{subfigure}
\begin{subfigure}[b]{0.16\textwidth}
{\includegraphics[width=\textwidth]{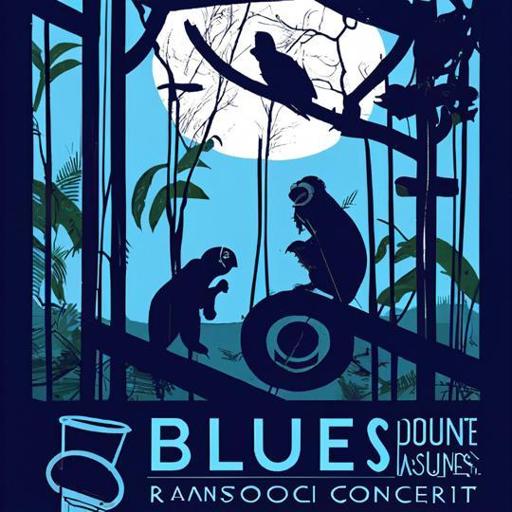}}
\vspace{-4mm}
\caption*{\scriptsize{DeepFloyd/IF$^\dagger$}}
\end{subfigure}
\begin{subfigure}[b]{0.16\textwidth}
{\includegraphics[width=\textwidth]{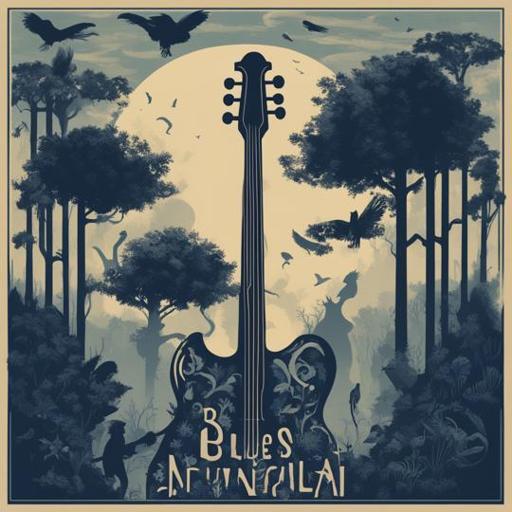}
}
\vspace{-4mm}
\caption*{\scriptsize{SDXL$^\dagger$}}
\end{subfigure}
\begin{subfigure}[b]{0.16\textwidth}
{\includegraphics[width=\textwidth]{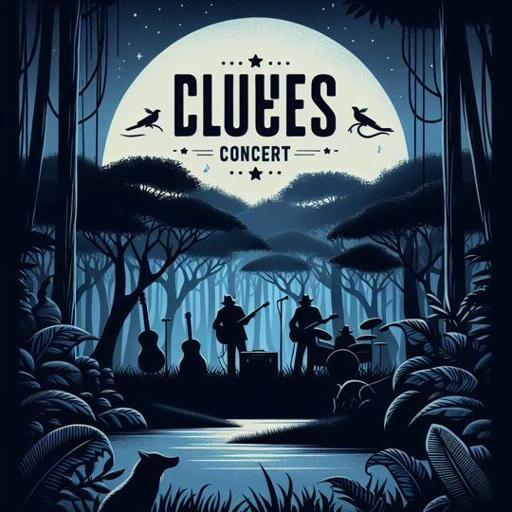}}
\vspace{-4mm}
\caption*{\scriptsize{\dalle$^\dagger$}}
\end{subfigure}
\vspace{-3mm}
\caption{\footnotesize{Comparison to the state-of-the-art systems on \ourbenchmark benchmark. We choose the DeepFloyd/IF-XL and SDXL-1.0 + SDXL-Refiner-1.0 models due to their superior performance. The superscript $\dagger$ symbol indicates that GPT-$4$ is used to convert user intentions into detailed captions.}}
\label{fig:compare_to_sota}
\end{minipage}
\end{figure*}
}

\begin{figure*}[t]
\begin{minipage}[htbp]{1\linewidth}
\centering
\begin{subfigure}[b]{0.12\textwidth}
{\includegraphics[width=\textwidth]{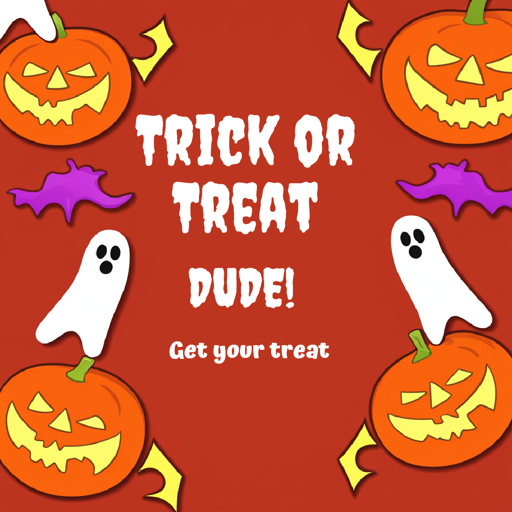}}
\vspace{-4mm}
\end{subfigure}
\begin{subfigure}[b]{0.12\textwidth}
{\includegraphics[width=\textwidth]{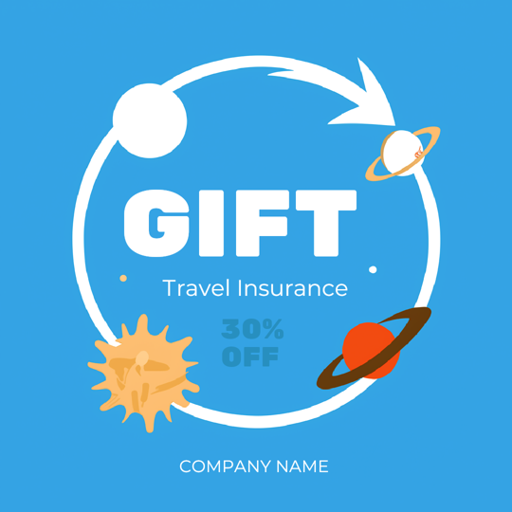}}
\vspace{-4mm}
\end{subfigure}
\begin{subfigure}[b]{0.12\textwidth}
{\includegraphics[width=\textwidth]{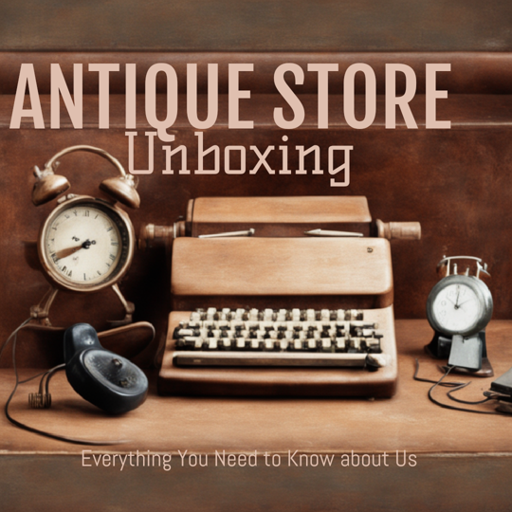}}
\vspace{-4mm}
\end{subfigure}
\begin{subfigure}[b]{0.12\textwidth}
{\includegraphics[width=\textwidth]{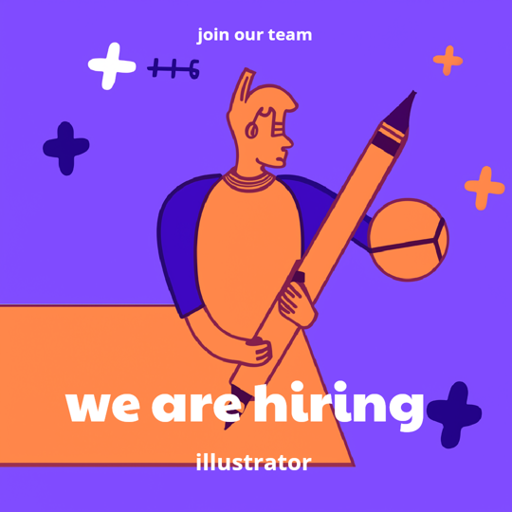}}
\vspace{-4mm}
\end{subfigure}
\begin{subfigure}[b]{0.12\textwidth}
{\includegraphics[width=\textwidth]{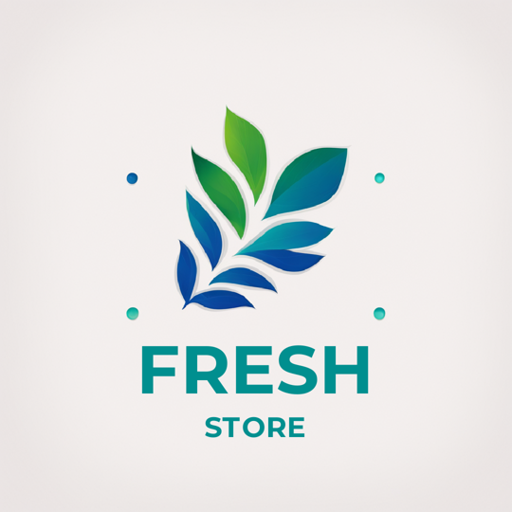}}
\vspace{-4mm}
\end{subfigure}
\begin{subfigure}[b]{0.12\textwidth}
{\includegraphics[width=\textwidth]{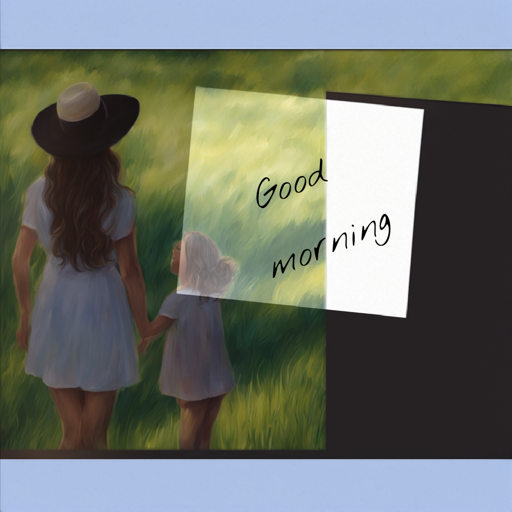}}
\vspace{-4mm}
\end{subfigure}
\begin{subfigure}[b]{0.12\textwidth}
{\includegraphics[width=\textwidth]{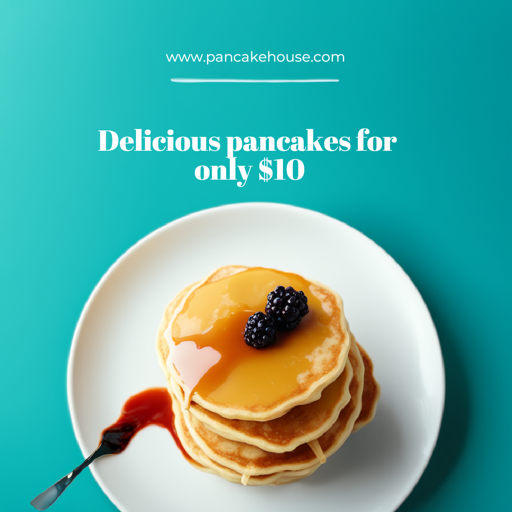}}
\vspace{-4mm}
\end{subfigure}
\begin{subfigure}[b]{0.12\textwidth}
{\includegraphics[width=\textwidth]{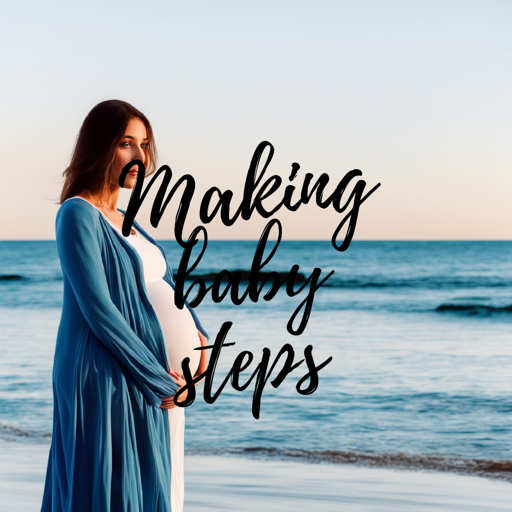}}
\vspace{-4mm}
\end{subfigure}\\
\begin{subfigure}[b]{0.12\textwidth}
{\includegraphics[width=\textwidth]{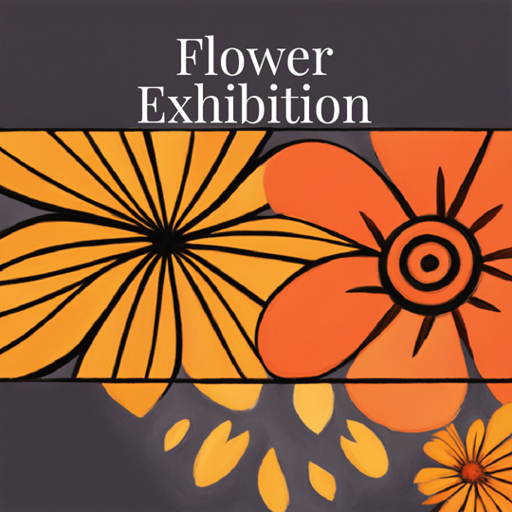}}
\vspace{-4mm}
\end{subfigure}
\begin{subfigure}[b]{0.12\textwidth}
{\includegraphics[width=\textwidth]{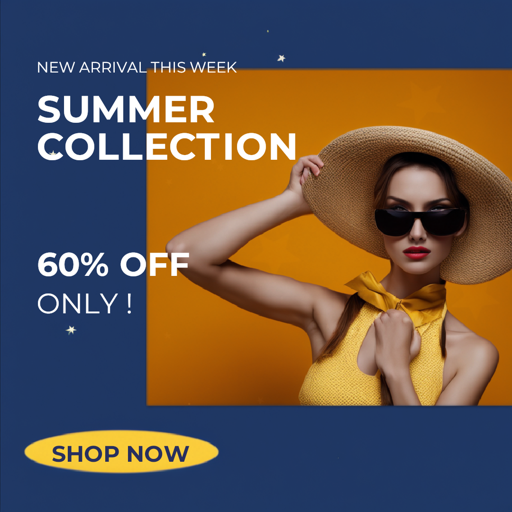}}
\vspace{-4mm}
\end{subfigure}
\begin{subfigure}[b]{0.12\textwidth}
{\includegraphics[width=\textwidth]{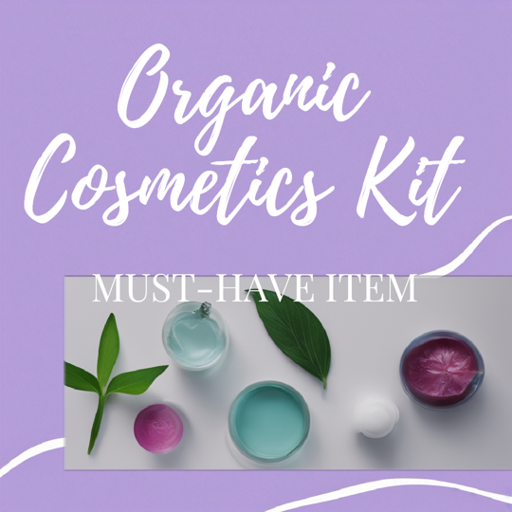}}
\vspace{-4mm}
\end{subfigure}
\begin{subfigure}[b]{0.12\textwidth}
{\includegraphics[width=\textwidth]{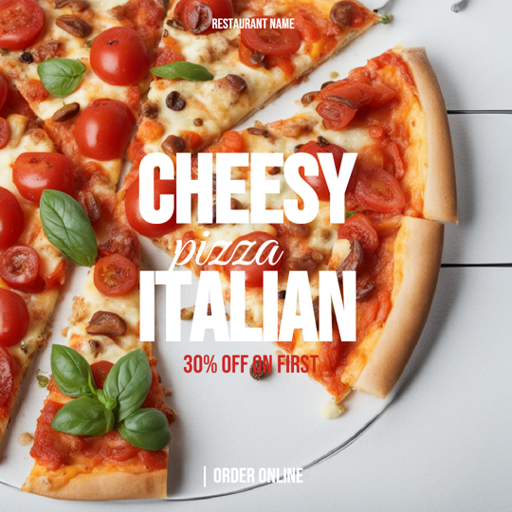}}
\vspace{-4mm}
\end{subfigure}
\begin{subfigure}[b]{0.12\textwidth}
{\includegraphics[width=\textwidth]{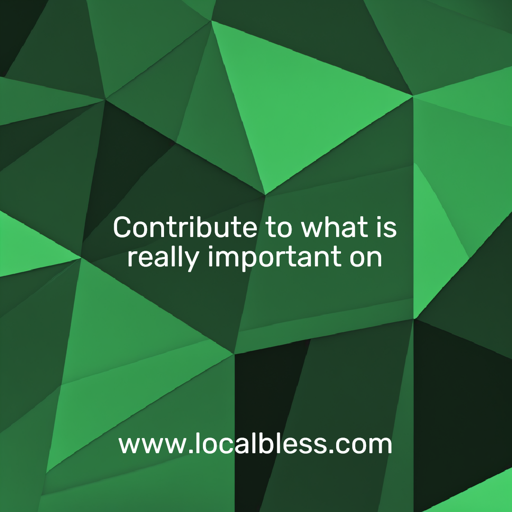}}
\vspace{-4mm}
\end{subfigure}
\begin{subfigure}[b]{0.12\textwidth}
{\includegraphics[width=\textwidth]{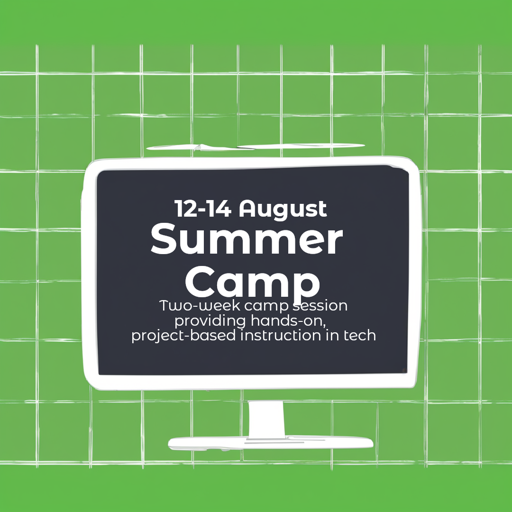}}
\vspace{-4mm}
\end{subfigure}
\begin{subfigure}[b]{0.12\textwidth}
{\includegraphics[width=\textwidth]{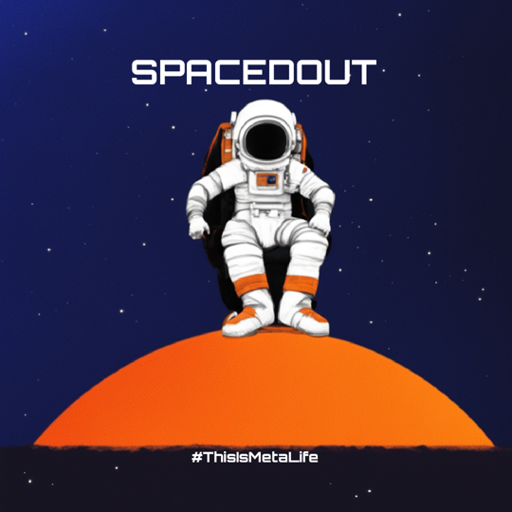}}
\vspace{-4mm}
\end{subfigure}
\begin{subfigure}[b]{0.12\textwidth}
{\includegraphics[width=\textwidth]{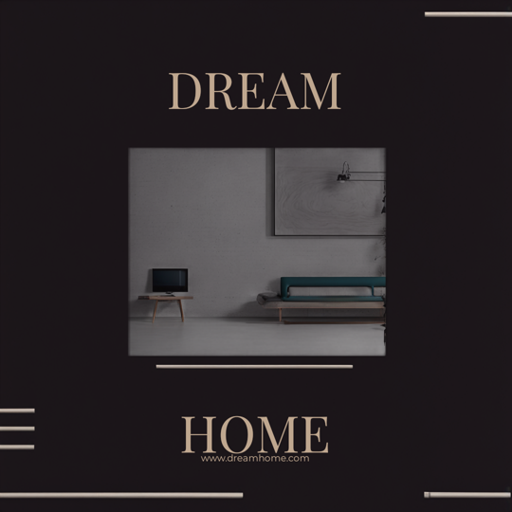}}
\vspace{-4mm}
\end{subfigure}\\
\begin{subfigure}[b]{0.12\textwidth}
{\includegraphics[width=\textwidth]{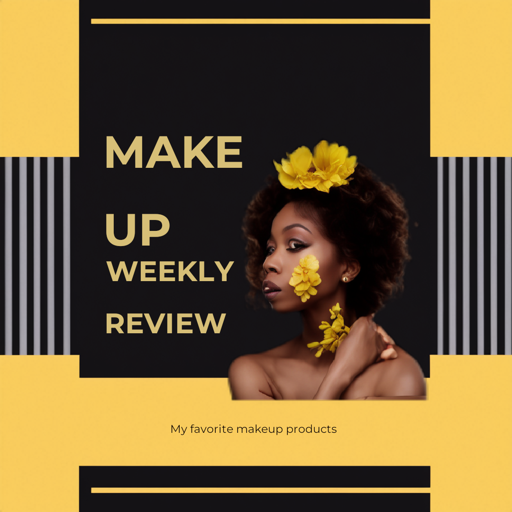}}
\vspace{-4mm}
\end{subfigure}
\begin{subfigure}[b]{0.12\textwidth}
{\includegraphics[width=\textwidth]{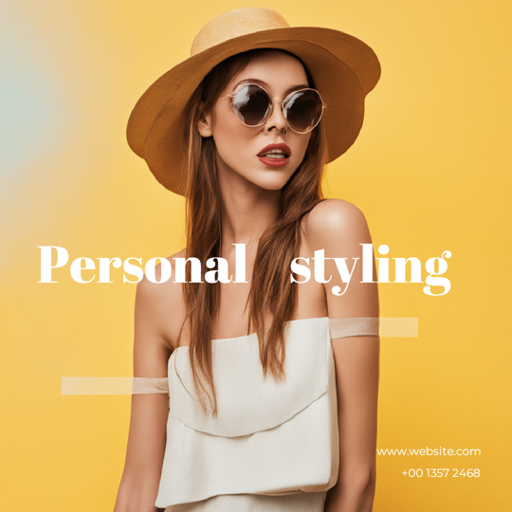}}
\vspace{-4mm}
\end{subfigure}
\begin{subfigure}[b]{0.12\textwidth}
{\includegraphics[width=\textwidth]{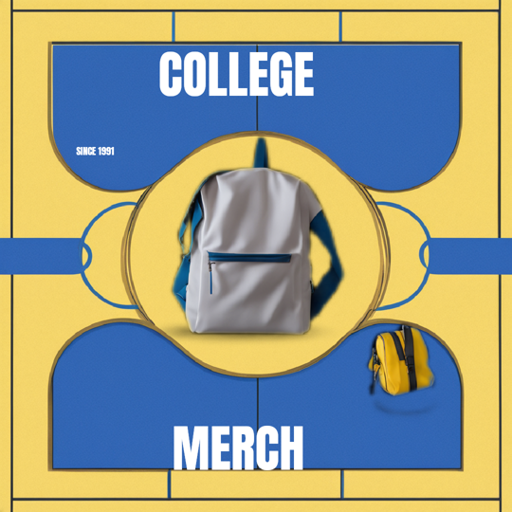}}
\vspace{-4mm}
\end{subfigure}
\begin{subfigure}[b]{0.12\textwidth}
{\includegraphics[width=\textwidth]{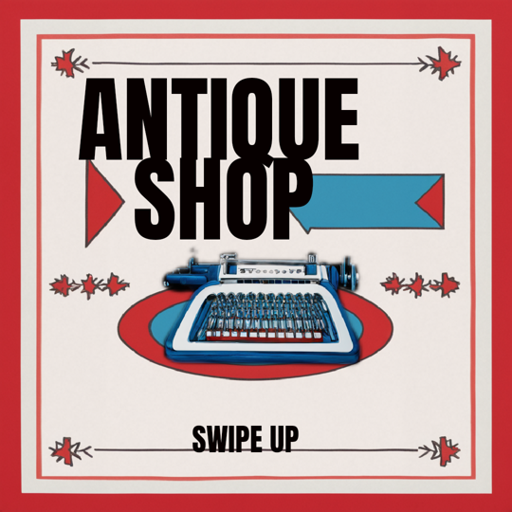}}
\vspace{-4mm}
\end{subfigure}
\begin{subfigure}[b]{0.12\textwidth}
{\includegraphics[width=\textwidth]{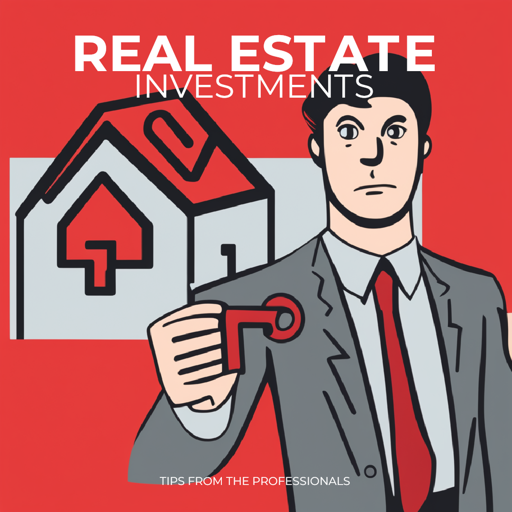}}
\vspace{-4mm}
\end{subfigure}
\begin{subfigure}[b]{0.12\textwidth}
{\includegraphics[width=\textwidth]{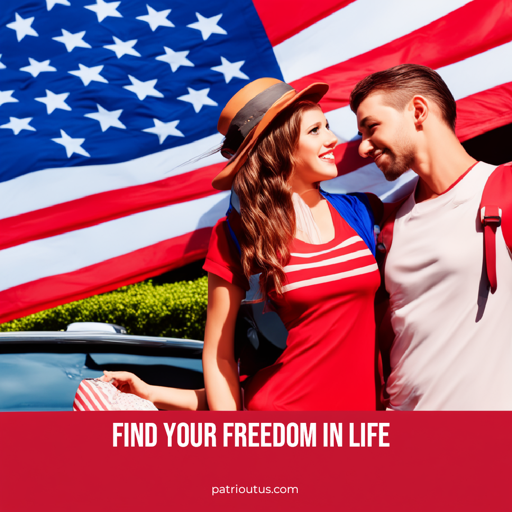}}
\vspace{-4mm}
\end{subfigure}
\begin{subfigure}[b]{0.12\textwidth}
{\includegraphics[width=\textwidth]{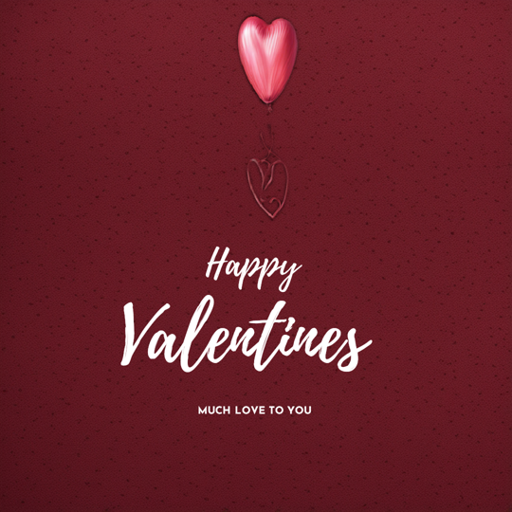}}
\vspace{-4mm}
\end{subfigure}
\begin{subfigure}[b]{0.12\textwidth}
{\includegraphics[width=\textwidth]{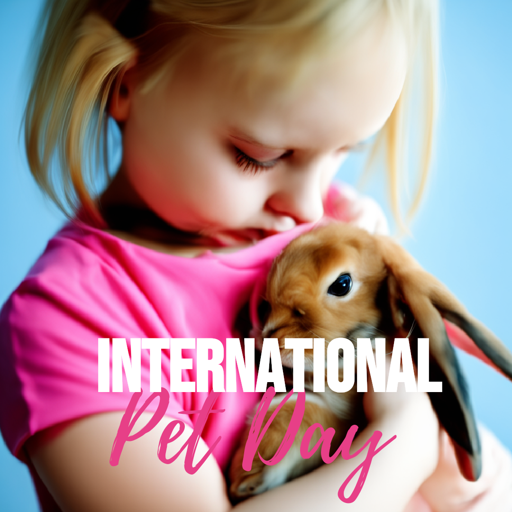}}
\vspace{-4mm}
\end{subfigure}\\
\vspace{-3mm}
\caption{\footnotesize{More qualitative results generated by our \ourname system.}}
\label{fig:more_results}
\end{minipage}
\vspace{-3mm}
\end{figure*}

\begin{figure}[t]
\begin{minipage}[t]{1\linewidth}
\centering
\begin{subfigure}[b]{0.235\textwidth}
{\includegraphics[width=\textwidth]{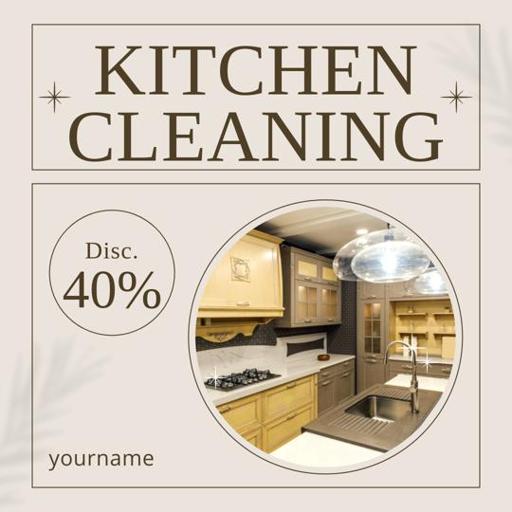}}
\vspace{-4mm}
\end{subfigure}
\begin{subfigure}[b]{0.235\textwidth}
{\includegraphics[width=\textwidth]{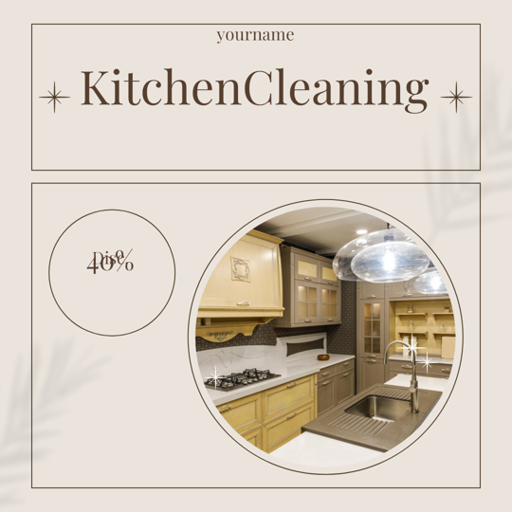}}
\vspace{-4mm}
\end{subfigure}
\begin{subfigure}[b]{0.235\textwidth}
{\includegraphics[width=\textwidth]{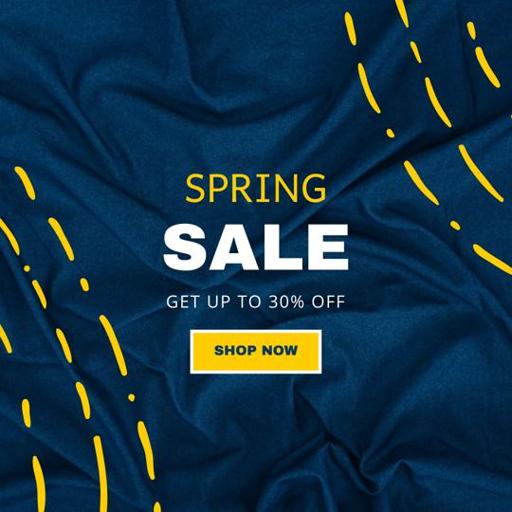}}
\vspace{-4mm}
\end{subfigure}
\begin{subfigure}[b]{0.235\textwidth}
{\includegraphics[width=\textwidth]{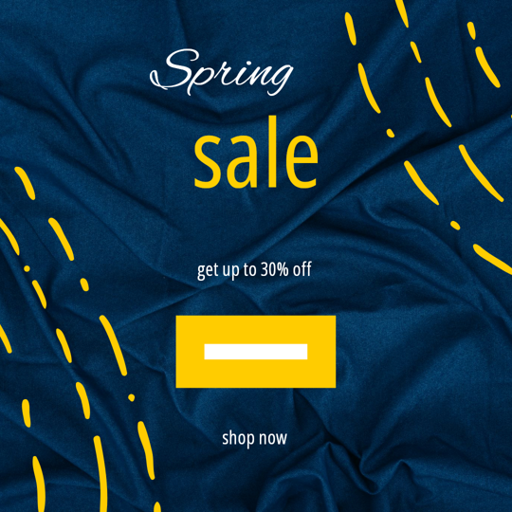}}
\vspace{-4mm}
\end{subfigure}
\caption{\small{Typography-LMM predictions ($2$-ed and $4$-th columns) vs. ground-truth ($1$-st and $3$-th columns).}}
\label{fig:typography_example_results}
\vspace{2mm}
\end{minipage}
\begin{minipage}[t]{1\linewidth}
\centering
\begin{subfigure}[b]{0.235\textwidth}
{\includegraphics[width=\textwidth]{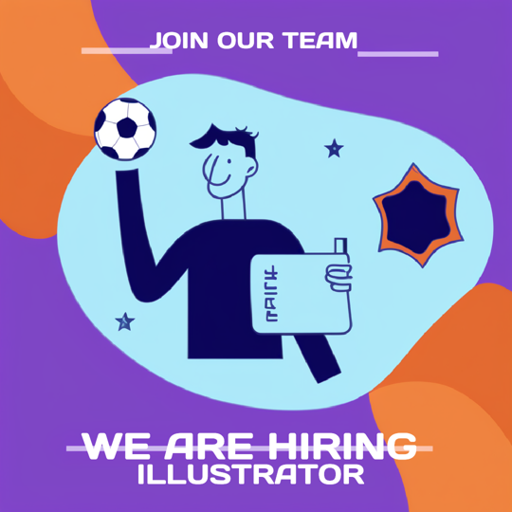}}
\vspace{-4mm}
\caption*{\small{Seed1}}
\vspace{-2mm}
\end{subfigure}
\begin{subfigure}[b]{0.235\textwidth}
{\includegraphics[width=\textwidth]{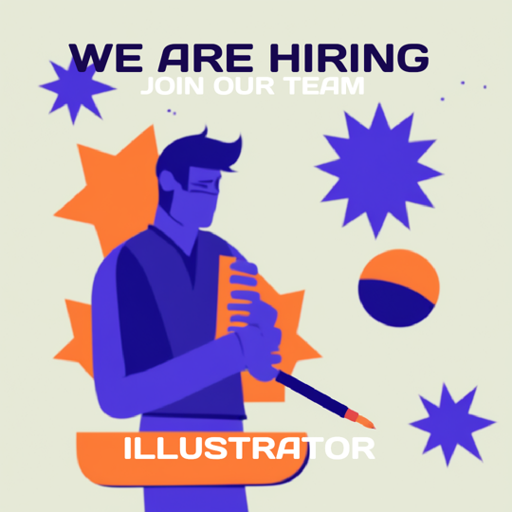}}
\vspace{-4mm}
\caption*{\small{Seed2}}
\vspace{-2mm}
\end{subfigure}
\begin{subfigure}[b]{0.235\textwidth}
{\includegraphics[width=\textwidth]{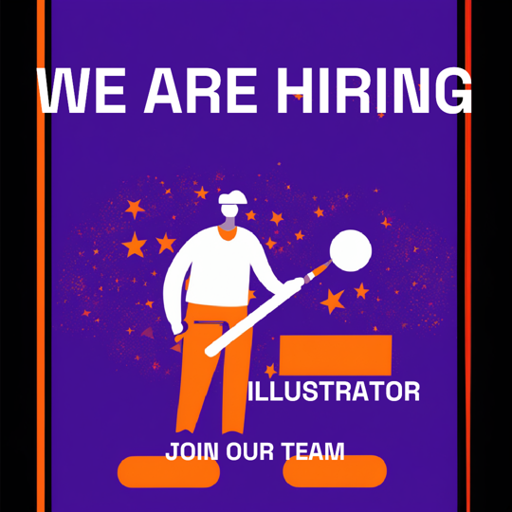}}
\vspace{-4mm}
\caption*{\small{Seed3}}
\vspace{-2mm}
\end{subfigure}
\begin{subfigure}[b]{0.235\textwidth}
{\includegraphics[width=\textwidth]{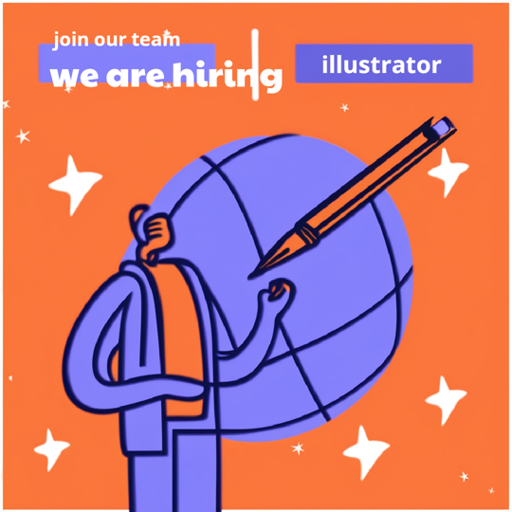}}
\vspace{-4mm}
\caption*{\small{Seed4}}
\vspace{-2mm}
\end{subfigure}
\caption{\footnotesize{Illustrating the variety within the graphic design images generated by our system when simply using different seeds for the diffusion models.}}
\label{fig:diversity}
\vspace{-5mm}
\end{minipage}
\end{figure}

To assess our \ourname's capability for design-oriented planning and reasoning in generating graphic design images, we conducted extensive experiments on our proposed \ourbenchmark, which consists of approximately  $\sim500$ graphic design intention prompts. Leveraging \gptv, we conducted a multi-faceted comparison ~\cite{dalle3paper,lin2023designbench} between our approach and the state-of-the-art image generators DeepFloyd/IF, SDXL, \dalle, and CanvaGPT, which utilize GPT-$4$ as an intention interpreter. Subsequently, we assembled approximately 20 general users and 6 professional designers to conduct a subjective assessment using COLE, comparing it separately with both \dalle and CanvaGPT. Then we show that the zero-shot performance with our Typography-LMM even outperforms the latest layout planning methods in Crello text box placement task. Last, we carry out ablation experiments to scrutinize the performance of each component and discuss the limitations of our system.

\vspace{-2mm}
\subsection{\textbf{\ourbenchmark} Benchmark}

We have constructed a benchmark for designer intentions, amassing approximately $\sim100$ intention prompts for each of the $5$ most pertinent categories, namely \emph{marketing}, \emph{covers and headers}, \emph{posts}, \emph{events}, and \emph{advertising}.
In a complementary effort, we have compiled an extra set of around $\sim50$ unique, creative user intentions.
Refer to the supplementary material for an all-inclusive review of these approximately $500$ intention prompts, along with the approximately $50$ creative ones, and comprehensive statistics.
We conduct all the experiments on \ourbenchmark by default.

\subsection{Main Results}

\noindent\textbf{Comparison with State-of-the-art}
Figure~\ref{fig:compare_to_sota} presents a comparison with various state-of-the-art text-to-image generation models that are augmented with GPT-$4$. Notably, we also include the results of the widely recognized custom CanvaGPT~\cite{canvagpt}, which integrates GPT-4 with Canva's templates. Each Canva template fully produced by professional designers incurs significant costs.
The usage of GPT-$4$ stems from the fact that both traditional text-to-image models and CanvaGPT generally necessitate detailed descriptions, which are often missing in the intention prompt. Consequently, we utilize GPT-$4$ across all models to translate the user's simple intentions into intricate captions. The generation results are then evaluated accordingly across different systems. We provide the original intention prompts in Table~\ref{tab:designer_prompt}.
We provide a detailed enhanced prompts, augmented using GPT-$4$, within the supplementary material.

In addition, we showcase their predicted numerical quality scores from \gptv based on \emph{quality assurance prompt} in Figure~\ref{fig:gpt4v_radar_score}. 
A careful evaluation reveals that our \ourname achieves very competitive quality accessing results even compared to the latest \dalle and CanvaGPT.
Our \ourname secures leading performance across three critical dimensions: design layout, typography color, and innovation, consistently achieving second-best in quality.
An interesting observation is that the background image generated by \dalle is of exceptional quality but the text regions are non-editable. Figure~\ref{fig:teaser} shows some impressive results of our \ourname when replacing our text-to-background generation model with \dalle by leveraging the text segmentation models, inpainting models, and others.

\vspace{1mm}
\noindent\textbf{Qualitative results} Figure~\ref{fig:more_results} showcases a variety of graphic design images entirely created by our \ourname system. 
It's noteworthy that our system performs exceptionally well, considering that each model within it is trained independently in a hierarchically organized manner. However, we recognize several limitations within our \ourname system, including: (i) the arrangement of the typography block, (ii) the limited number of editable visual elements in the image, and (iii) the restricted diversity in typography color selection. Addressing these issues is a direction we'd like to pursue in our future work.

\begin{table*}[htbp]
\centering
\tablestyle{15pt}{1.1}
\resizebox{1.0\linewidth}{!}  
{
\begin{tabular}{c|c|c|c|c|c|c}
& \multicolumn{3}{c|}{SingleText} & \multicolumn{3}{c}{MultiText} \\ 
\cline{2-7}
 & SmartText$+$\cite{li2022smarttext} & FlexDM\cite{inoue2023towards} & Typography$\-$LMM & SmartText$+$\cite{li2022smarttext} & FlexDM\cite{inoue2023towards} & Typography$-$LMM \\ 
\shline
IoU$\uparrow$ & 4.7 & 35.7 & \textbf{40.2} & 2.3 & 11.0 & \textbf{17.2} \\ 
\end{tabular}
}
\caption{\small{Zero-shot performance of Typography$-$LMM on the Crello text box placement task. We have conducted data cleaning to avoid duplicates in our training set.}}
\label{tab:layout_iou}
\end{table*}

\subsection{Ablation Experiments}

\begin{figure}[t]
\centering
\begin{minipage}[t]{1\linewidth}
\vspace{2mm}
\centering
\begin{subfigure}[b]{0.31\linewidth}
{\includegraphics[width=\textwidth]{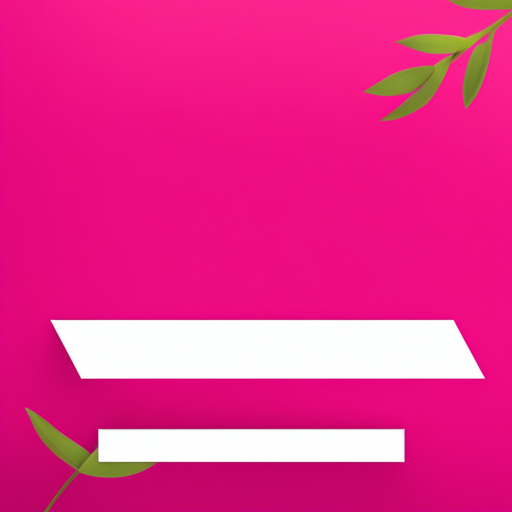}}
\end{subfigure}
\begin{subfigure}[b]{0.31\linewidth}
{\includegraphics[width=\textwidth]{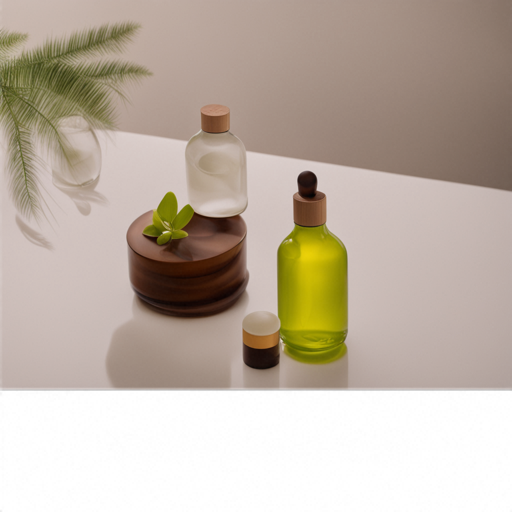}}
\end{subfigure}
\begin{subfigure}[b]{0.31\linewidth}
{\includegraphics[width=\textwidth]{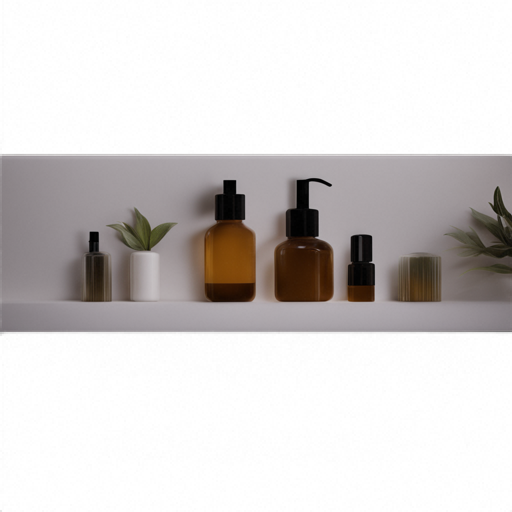}}
\end{subfigure}
\end{minipage}
\begin{minipage}[t]{1\linewidth}
\centering
\begin{subfigure}[b]{0.31\linewidth}
{\includegraphics[width=\textwidth]{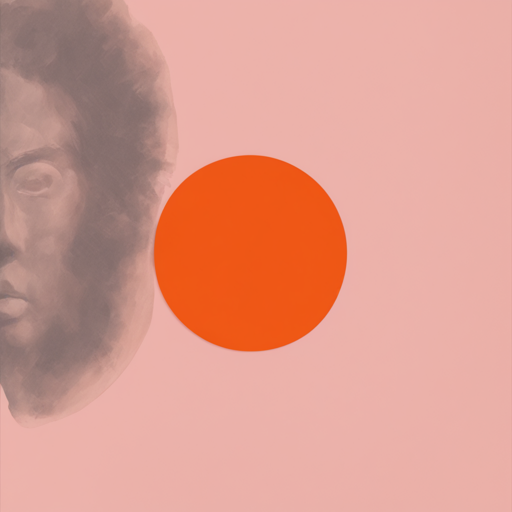}}
\vspace{-4mm}
\end{subfigure}
\begin{subfigure}[b]{0.31\linewidth}
{\includegraphics[width=\textwidth]{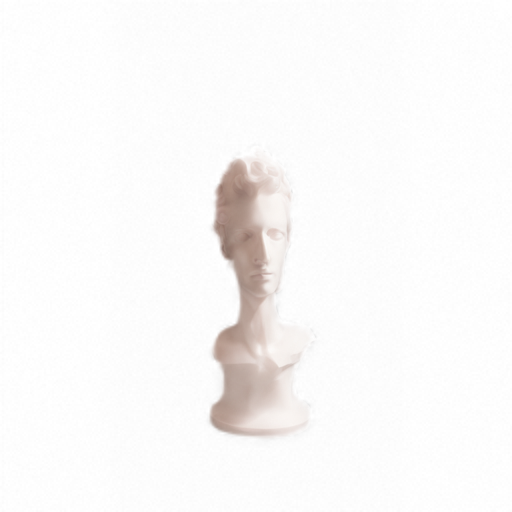}}
\vspace{-4mm}
\end{subfigure}
\begin{subfigure}[b]{0.31\linewidth}
{\includegraphics[width=\textwidth]{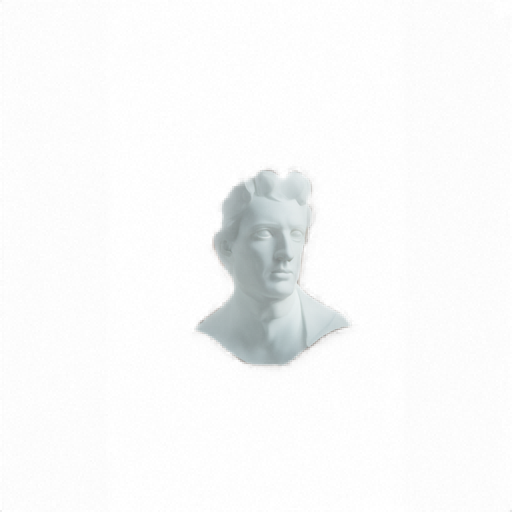}}
\vspace{-4mm}
\end{subfigure}
\end{minipage}
\begin{minipage}[t]{1\linewidth}
\centering
\begin{subfigure}[b]{0.31\linewidth}
{\includegraphics[width=\textwidth]{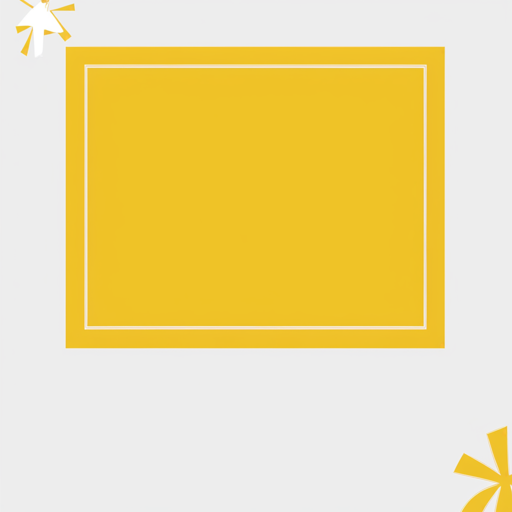}}
\vspace{-4mm}
\end{subfigure}
\begin{subfigure}[b]{0.31\linewidth}
{\includegraphics[width=\textwidth]{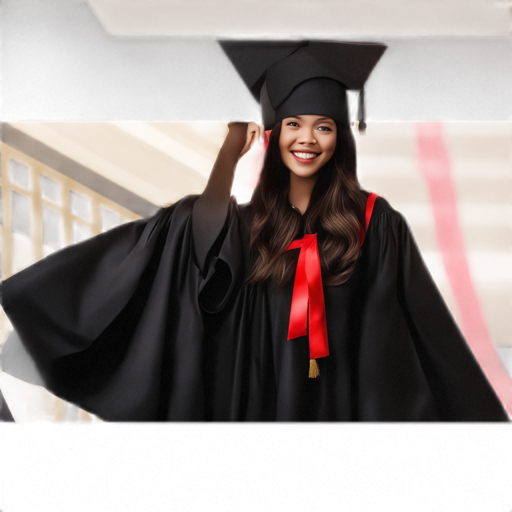}}
\vspace{-4mm}
\end{subfigure}
\begin{subfigure}[b]{0.31\linewidth}
{\includegraphics[width=\textwidth]{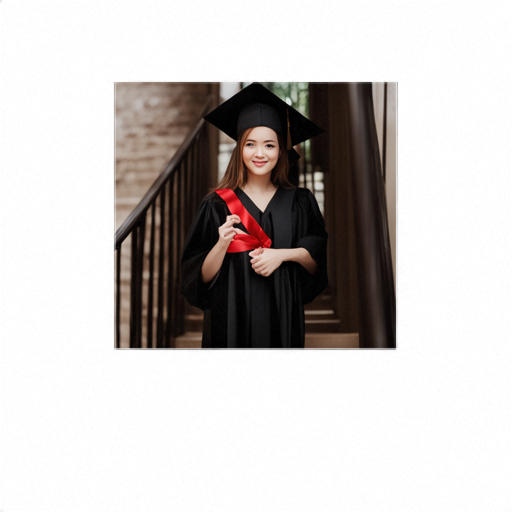}}
\vspace{-4mm}
\end{subfigure}
\caption{\small{Effect of composed image prediction in text-to-object generation. Left to right: background image (1-st, 4-th, 7-th), object image w/o composed image prediction (2-ed, 5-th, 8-th), object image w/ composed image prediction (3-rd, 6-th, 9-th).}}
\label{fig:effect_compose}
\end{minipage}
\vspace{-6mm}
\end{figure}

\noindent\textbf{Text-to-Background \& Text-to-Object}
Figure~\ref{fig:effect_compose} illustrates the critical role of incorporating composed image prediction into the text-to-object diffusion model. This integration markedly enhances the model's ability to plan layouts. Furthermore, we have conducted an exhaustive comparison between the background images produced by our model and those generated by other methods, such as those described in~\cite{podell2023sdxl, Deepfloyd, dalle3paper}. We found that most alternative approaches often result in objects that fill the entire rectangular canvas. For additional information, please consult the supplementary material.

\vspace{1mm}
\noindent\textbf{Typography LMM \& Reflect LMM}
We visualize the predicted results with the ground-truth images as the condition in Figure~\ref{fig:typography_example_results}.
We assess these results from two perspectives: typography scoring with \gptv and the localization accuracy including $\mathsf{mIoU}$, $\mathsf{AP^{25}}$, and $\mathsf{AP^{50}}$.
To assess the generalization capability of our approach in layout planning and attribute reasoning for visual text, we evaluate the zero-shot performance on the Crello benchmark and summarize the comparison results in Table~\ref{tab:layout_iou}. Our Typography-LMM outperforms the previous model, FlexDM, by +4.5 and +6.2 in terms of IoU score for single and multiple text box placement tasks, respectively
Refer to the supplementary material for the ablation of reflect-LMM.

\vspace{1mm}
\noindent\textbf{Design Diversity}
Designers frequently confront the challenge of producing varied yet high-quality designs. Our \ourname system demonstrates exceptional performance in this aspect, as illustrated in Figure~\ref{fig:diversity}.
We provide additional ablation experiments on the Design LLM and others in the supplementary material.

\section{Conclusion}
\label{sec:conclusion}
This paper introduces an exceptionally efficient hierarchical generation framework, \ourname, that simplifies the intricate process of graphic design generation by breaking it down into manageable, coordinated sub-tasks, each yielding distinct outputs. By leveraging design-oriented visual planning, reasoning, and layer-wise generation capabilities during this hierarchical task decomposition, our system can generate multi-layered and editable graphic design images. We demonstrate the advantages of our approach by comparing it to the most powerful systems like \dalle and CanvaGPT. Additionally, we have pinpointed several key limitations of our \ourname system in the context of generating graphic design images, and we plan to address these in future research.

\vspace{3mm}
\noindent\textbf{Acknowledgement}
We extend our gratitude to Xinyang Li for his invaluable assistance with data preparation. Additionally, we are grateful to Yang Ou, along with both designers and non-designers, for their insightful feedback from a professional design and normal user perspective. We appreciate the enlightening discussions offered by Chunyu Wang, Jianmin Bao, Mingxi Cheng, and Chong Luo. Additionally, we want to give a special mention to Weicong Liang for his indispensable help in addressing the LaTeX compilation errors.

{
\small
\bibliographystyle{ieeenat_fullname}
\bibliography{main}
}

\newcounter{reviewcomment@counter}  

\maketitlesupplementary

\section*{A Graphic Design Pipeline of Human Designer}

\begin{figure}[htbp]
\centering
\includegraphics[width=1\linewidth]{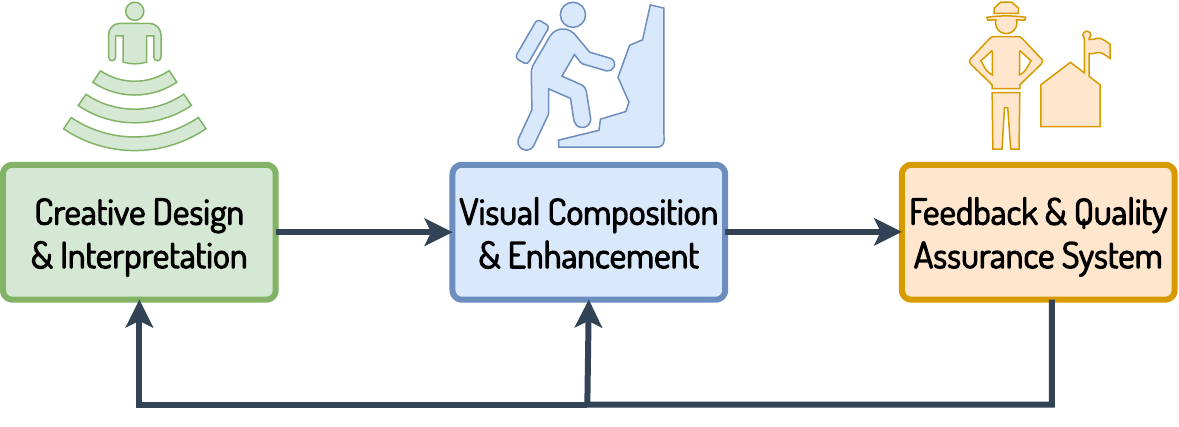}
\caption{\footnotesize{Illustrating the overall pipeline of producing high-quality graphic design.}}
\label{fig:design_pipeline}
\vspace{-5mm}
\end{figure}

The creation of high-quality graphic designs from user intentions necessitates a seamless integration of diverse roles or components, each uniquely designed to tackle specific facets of the graphic design process~\cite{wiki_graphic_design,vise2021interdisciplinary}. Initially, we present an elaborate blueprint that delineates these roles, elucidates their functions, and delineates their interactions. Figure~\ref{fig:design_pipeline} depicts this comprehensive pipeline, serving as the conceptual foundation for the design of our \ourname system.

\vspace{1mm}
\noindent \textbf{Creative Design and Interpretation}  
The core responsibility of this role is to interpret user intentions and convert them into specific design elements, including titles, subtitles, headings, subheadings, and body text. This role also establishes the tone and style of the graphic design, encompassing aspects like primary color, theme, style, and captions.

\vspace{1mm}
\noindent \textbf{Visual Composition and Enhancement}  
The key responsibility of this role is to uphold the principles of design aesthetics and layout, which involves the generation, assessment, and adjustment of visual appeal and spatial organization. The generation process mainly utilizes a comprehensive library of images, icons, and textures to enhance the design. This module further refines the initial concepts originating from the Creative Design and Interpretation engine, optimizing the layout and enriching the design with suitable assets. For instance, it is imperative to not only generate images of objects, decorations, and backgrounds, but also to predict various typographic attributes for elements such as the heading, subheading, and body text.

\vspace{1mm}
\noindent \textbf{Feedback and Quality Assurance System}  
The principal function of this role is to integrate user feedback into the design in an iterative manner, with the objective of ensuring that generated designs comply with quality benchmarks and regulatory standards. This module guarantees that the designs not only align with user expectations, but also fulfill professional and legal criteria. It maintains a continuous interaction with the first two modules to guide the design refinement process. 

\vspace{1mm}
\noindent \textbf{Collaboration Loop}   
This system ensures that the designs not only meet user expectations, but also adhere to professional and legal standards. It maintains a constant dialogue with the first two modules to navigate the design refinement process effectively.

\vspace{1mm}
\noindent \textbf{Discussion}
Unlike traditional graphic design systems that rely significantly on the expertise of professional designers, we aspire to create a DesignerAgent. This innovative system is designed to generate superior graphic designs, leveraging user intention descriptions alone. By fine-tuning state-of-the-art Large Multimodal Models (LMMs), Large Language Models (LLMs), and diffusion models, we aim to fully exploit our carefully curated, high-quality graphic design datasets to achieve this objective.

\begin{intentionprompt}
You are an excellent image analyst and caplable of guessing a image designer's intention. I will give you an image's necessary information, including image title, image format, image keywords, all text contained in the image. Your task is to output a JSON formatted string. This string contains a key value, intension, and the corresponding value is the user intention for designing this image. Please output the user intension from the perspective of the user using the image generation tool. Please note that the text information carried in the image may be helpful in the output results. Please include unique and necessary information in the entered text in the caption, such as website address, phone number, price, etc. Do not output any other irrelevent information. If needed, you can make reasonable guesses. Please refer to the example below for the desired format.
\end{intentionprompt}

\section*{B Detailed System Prompts for GPT-3.5 and \textbf{\gptv}}

We introduce the \emph{Intention Generation Prompt} for GPT-3.5, \emph{Quality Assurance Prompt} for \gptv, and \emph{Pair-wise Comparison Prompt} for \gptv as follows: The \emph{Intention Generation Prompt} is crafted to deduce the user's purpose utilizing diverse image data, such as keywords, all text visible in the image, the image's title, among others. The \emph{Quality Assurance Prompt} for \gptv is established to evaluate the quality of the created graphic design images across roughly five dimensions. These dimensions include design and layout, content pertinence, graphics and imagery, creativity, and the use of typography and color. The \emph{Pair-wise Comparison Prompt} for \gptv is proposed to objectively compare our generated images with results from other methods from six perspectives of graphic design.

\begin{qualityprompt}
You are an autonomous AI Assistant who aids designers by providing insightful, objective, and constructive critiques of graphic design projects.

Your goals are:
Deliver comprehensive and unbiased evaluations of graphic designs based on established design principles and industry standards.
Identify potential areas for improvement and suggest actionable feedback to enhance the overall aesthetic and effectiveness of the designs.
Maintain a consistent and high standard of critique.
Utilize coordinate information for data description relative to the upper left corner of the image, with the upper left corner serving as the origin, the right as the positive direction, and the downward as the positive direction.
 
Please abide by the following rules:
Strive to score as objectively as possible.
Grade seriously. A flawless design can earn 10 points, a mediocre design can only earn 7 points, a design with obvious shortcomings can only earn 4 points, and a very poor design can only earn 1-2 points.
Keep your reasoning concise when rating, and describe it as briefly as possible. If the output is too long, it will be truncated.
Only respond in JSON format, no other information.
 
Grading criteria:

Design and Layout (1-10): The graphic design should present a clean, balanced, and consistent layout. The organization of elements should enhance the message, with clear paths for the eye to follow. A score of 10 signifies a layout that maximizes readability and visual appeal, while a 1 indicates a cluttered, confusing layout with no clear hierarchy or flow.

Content Relevance and Effectiveness (1-10): The content should be not only relevant to its purpose but also engaging for the intended audience, effectively communicating the intended message. A score of 10 means the content resonates with the target audience, aligns with the design's purpose, and enhances the overall message. A score of 1 indicates the content is irrelevant or does not connect with the audience.

Typography and Color Scheme (1-10): Typography and color should work together to enhance readability and harmonize with other design elements. This includes font selection, size, line spacing, color, and placement, as well as the overall color scheme of the design. A score of 10 represents excellent use of typography and color that aligns with the design's purpose and aesthetic, while a score of 1 indicates poor use of these elements that hinders readability or clashes with the design.

Graphics and Images (1-10): Any graphics or images used should enhance the design rather than distract from it. They should be high quality, relevant, and harmonious with other elements. A score of 10 indicates graphics or images that enhance the overall design and message, while a 1 indicates low-quality, irrelevant, or distracting visuals.

Innovation and Originality (1-10): The design should display an original, creative approach. It should not just follow trends but also show a unique interpretation of the brief. A score of 10 indicates a highly creative and innovative design that stands out in its originality, while a score of 1 indicates a lack of creativity or a generic approach.
\end{qualityprompt}

\vspace{2mm}
\subsection*{C The Detailed Prompt List of \textbf{\ourbenchmark}}
A comprehensive list of designer intention prompts is summarized across Table~\ref{tab:prompt_list_tab1}, Table~\ref{tab:prompt_list_tab2}, Table~\ref{tab:prompt_list_fig11} and Table~\ref{tab:prompt_list_tab4}.   
  
Table~\ref{tab:prompt_list_tab1} details the GPT-4 augmented intention prompts corresponding to the images exhibited in Figure 1 of the main paper.

Table~\ref{tab:prompt_list_tab2} presents the intention prompts which were utilized for \ourname in Figure 2 of the main paper.  

Table~\ref{tab:prompt_list_fig11} presents the intention prompts and GPT-4 augmented intension prompts which were utilized for \ourname in Figure 11 of the main paper. 

Last, Table~\ref{tab:prompt_list_tab4} presents the intention prompts which were utilized for \ourname in Figure 12 of the main paper. 

\begin{pkprompt}
Your task is to evaluate the accuracy of images generated by different computer programs, using their accompanying captions as a reference. You will be provided with a caption and two images, which were produced by distinct software and designed to test the capabilities of image generation technology, and the texts that the images must contain to convey the intention of the image. Your role involves comparing these images based on specific criteria. Focus solely on these elements during your assessment:

Composition:
Evaluate whether the elements within each image are strategically placed and harmonious. Check if there's a pronounced center of attention in the arrangement. Ponder over the excellence of each image, focusing on the aforementioned elements, to discern which one surpasses the other. 

Image Composition:
The important thing is that the picture has a clear subject, and does not overwhelm the subject with overly complicated beautification.The key focus is to ensure the visual harmony of the image. Pay attention to the balance and arrangement of elements within the picture. Evaluate whether the image presents a cohesive and visually pleasing composition. Consider the effectiveness of the layout and the interaction of objects in the scene.

Medium: 
The most important point is that the artistic expression must fit the user's caption, otherwise the artistic expression will be meaningless.Focus on the unique artistic qualities and techniques displayed in each image. Consider the effectiveness of these artistic choices, reflecting on how well they align with the given caption. Evaluate the distinctiveness and merits of the artistic style and medium.

Overall Alignment: The most important thing is whether the picture can meet the user's creative purpose and whether it is practical. Beauty is not the only guarantee.Assess if the image faithfully represents the caption. Ensure that all elements mentioned in the caption are present and accurately portrayed in the image. Consider how well the image aligns with the caption in terms of accuracy and completeness.

Overall Asthetics:
The most important point is whether the overall picture background and text are harmonious, whether the text is clearly visible, whether the picture matches the font perfectly, the theme between the picture and text is clear, and the colors are harmonious. Consider the overall appeal and creativity of the design. Does it provide an original and innovative interpretation of the caption? Contemplate its uniqueness and whether it brings a new perspective to the table. 

Text Render:
The most important thing is to ensure that the color selection of all text and the background color are not the same. The same color will cause the visibility of the text to be very low. Please consider whether the text can be clearly displayed on the picture as an important indicator.Examine whether the text is readable and appropriately styled. Check for correct spelling. Assess the suitability of font choices, spacing, and alignment in relation to the overall design. 

Reflect on these qualities to determine which image surpasses the other. Voice your thought process and ultimately announce your preference, either as '$|$ Image 1' or '$|$ Image 2'. Adhere to these principles: 1. Avoid excessive criticism. If the key elements of the caption are mostly accurate, the image is considered satisfactory. 2. Disregard elements not directly mentioned in the caption, even if they appear in the image. 3. Acceptable if the depiction of objects is somewhat distorted, as long as they are identifiable to a person. 4. Your conclusion should be either '$|$ Image 1' or '$|$ Image 2'. 5. Aim to discern the superior image. If both appear equally lacking, a random choice is permissible. 6. Keep your explanatory remarks brief, under 50 words.
\end{pkprompt}

\subsection*{D More Details of Bin Partition Scheme within Typography-LMM}

To streamline the Typography-LMM's prediction process, we adopt a discretization strategy, i.e., bin partition scheme, for continuous floating-point values based on the approach described in~\cite{chen2021pix2seq}. This involves categorizing continuous variables into discrete bins and representing the values within each bin with a unique token. The specifics of our bin partitioning scheme are detailed below:
\begin{itemize}
\item Font Size: The range [2, 200] is divided into 100 sub-intervals.
\item Angle: The range [0, 2$\pi$] is segmented into 64 sub-intervals.
\item Color (RGB values): The range [0, 255] is split into 32 sub-intervals.
\item Text Bounding Box Coordinates: The normalized coordinates within the range [-1, 1] are partitioned into 256 subspaces.
\item Opacity: The range [0, 255] is divided into 8 subspaces.
\item Letter/Line Spacing: The normalized values within the range [0, 1] are divided into 40 sub-intervals, adjusted according to the image width.
\end{itemize}

We estimate the start and end points of all the above partitions based on the statistics derived from the training set, accommodating a broad spectrum of typography characteristics, including font size, angle, letter and line spacing, opacity, dimensions of the font bounding box (left, top, width, height), and color.

\subsection*{E More Ablation Experiments}

\noindent\textbf{Design LLM} We calculate the semantic scores (semantic textual similarity scores estimated based on Sentence-BERT) to compare the ground-truth with predictions from our Design-LLM, GPT-3.5, and GPT-4 across the test dataset. To ensure accurate predictions, we supply three in-context examples for both GPT-3.5 and GPT-4. Furthermore, we restrict GPT-3.5 and GPT-4 to generate values for only one key query at a time, simplifying the task's complexity. The results, detailed in Table~\ref{tab:designllm}, reveal that our Design-LLM significantly outperforms both GPT-3.5 and GPT-4.

\begin{table}[htbp]
\centering
\tablestyle{5pt}{1.}  
\scriptsize{
{
\begin{tabular}{l|c|c|c}
Keys in JSON & Design LLM & GPT-3.5 & GPT-4 \\
\shline
Global Caption & \textbf{68.57} & 56.31 & 59.56 \\
Category & 80.32 & \textbf{80.93} & 78.65 \\
Keywords & \textbf{89.31} & 84.67 & 84.14 \\
Background Caption & \textbf{57.69} & 48.06 & 52.37\\
Object Caption & \textbf{73.59} & 73.21 & 73.16\\
\makecell[l]{Text layer \\ (heading, sub-heading, body-text)}
& \textbf{54.75} & 46.80 & 53.56\\ 
\end{tabular}
}}
\caption{  
\footnotesize{Comparison Results for Design LLM ablation study.}}
\label{tab:designllm}
\end{table}

\noindent\textbf{Text-to-Background \& Text-to-Object}
To assess the efficacy of our text-to-background diffusion model, we conducted a comparative analysis involving DeepFolyd/IF, SDXL, and \dalle, as depicted in Figure~\ref{fig:compare_to_sota_bg}. The complete list of prompts for generating corresponding background images is listed in Table~\ref{tab:prompt_list_tab5}. Our findings reveal a notable trend among current state-of-the-art diffusion models: they often generate objects that occupy the entirety of the image, neglecting adequate space for accommodating visual text.

\begin{table}[htbp]
\vspace{-3mm}
\begin{minipage}[t]{1\linewidth}  
\centering  
\tablestyle{1pt}{1.2}  
\resizebox{1.0\linewidth}{!}  
{  
\begin{tabular}{l|>{\centering\arraybackslash}m{15cm}}  
Category & Intention Prompt \\  
\shline  
posts & \small{An eye-catching car sale ad featuring a glossy white hatchback automobile, gleaming under the glow of a soft spotlight. The car's sleek contours and sparkling rims are accentuated. Bold, crisp letters in the corner proclaim the enticing price of \$13,900, with a monthly payment estimate of \$179 highlighted beneath. A smart, clickable URL, CALIPSUN.CO.UK, is displayed prominently for further details, serving as a beacon for potential buyers.} \\ \hline
posts & \small{A grand display of our exclusive timepiece collection, each watch exuding elegance and style, set against a sophisticated black background. The watches are embedded with intricate details and shimmering dials. A bold, eye-catching '40\% off' sign is placed prominently, urging viewers to upgrade their style. On the top corner, the Instagram Story's 'Swipe Up' prompt is nestled, symbolizing the immediacy of this limited-time offer.} \\  \hline
advertising & \small{A burst of vibrant pink hues dominates the canvas: a mix of blush, fuchsia and rose clothing neatly arranged. In the corner, a gleaming starburst accentuates a glowing testimonial from Jenny Wilson, her words of praise shimmering. Beneath the testimonial, elegantly scripted contact information is highlighted: phone number 123-456-789, website WWW.FASHIONSTYLE.COM, and an address at 123 Anywhere ST., Any City.} \\  \hline
posts & \small{An energetic poster with a vibrant color palette, showing joyful children safely riding top-quality scooters. Bold text announces, "Back to School Scooter Sale!" A prominent "40\% off" starburst highlights the best scooters for kids. Parents are seen looking on contentedly, knowing they've chosen a fun, safe ride for their children to school.} \\ \hline

marketing & \small{An elegantly designed gift certificate for a reputable travel insurance company. The certificate, made of high-quality paper, is adorned with motifs of different modes of travel - airplanes, ships, and trains. A shiny, golden 30\% off stamp is prominently placed at the center. The company's logo is tastefully positioned at the top, while the certificate's edges are softly curved, enhancing its appeal.} \\ \hline

covers\& headers & \small{A vividly illuminated stage under a starry night sky, acting as the primary backdrop. Smack in the center is a caricature of John Oliver, his signature glasses glinting, holding a microphone with a wide grin. At the top, bold letters spell "Laugh Out Loud with John Oliver". Below, in elegant script, is the location "At The Comedy Club, New York". A chuckling audience silhouette completes the lively atmosphere.} \\ \hline

creative & \small{A collection of vibrant posters for a fictitious Ancient Greek Olympics. Each poster features a different sport: a fiercely competitive chariot race with horses kicking up dust, a discus thrower showcasing his strength against a setting sun, all stylized like traditional Greek pottery art, with the sportsmen clad in traditional tunics, their muscles rippling under the warm, orange glow of the Grecian sun. The borders of each poster are adorned with intricate patterns of laurel wreaths.} \\ \hline
advertising & \small{An eye-catching weekend sale advertisement for skincare products, featuring prominently a sleek bottle of cosmetic serum bathed in a soft, ethereal glow. A bold, ruby-red text proudly declares "All Items Discounted Up To 30\%". Surrounding the serum are various skincare items, all discounted, framed against a backdrop of vibrant autumn leaves, reflecting this season's promotion.}

\end{tabular}  
}  
\caption{  
\footnotesize{The augmented intention prompts with GPT-$4$ listed in Figure 1.}}
\label{tab:prompt_list_tab1}
\vspace{3mm}
\end{minipage}
\begin{minipage}[t]{1\linewidth}  
\centering  
\tablestyle{1pt}{1.1}  
\resizebox{1.0\linewidth}{!}
{  
\begin{tabular}{l|>{\centering\arraybackslash}m{15cm}}  
Category & Intention Prompt \\  
\shline  
creative & \small{Design an animated digital poster for a fantasy concert where Joe Hisaishi performs on a grand piano on a moonlit beach. The scene should be magical, with bioluminescent waves and a starry sky. The poster should invite viewers to an exclusive virtual event.} \\ \hline
creative & \small{Design an alluring poster for a blues concert in a clearing of the Amazon Rainforest. The artwork should evoke the soulful tunes of blues music echoing through the wilderness. Use a moody, midnight blue palette and incorporate silhouettes of blues instruments and nocturnal wildlife.} \\  \hline
covers\& headers & \small{Create an Instagram story promoting a special Christmas offer for a chocolate drink. The offer includes a 20\% discount and encourages customers to order now.} \\   \hline
advertising & \small{Create a customer testimonial advertisement showcasing a pink clothing collection. The advertisement includes a positive review from Jenny Wilson.} \\

\end{tabular}  
}  
\caption{  
\footnotesize{The intention prompts listed in Figure 2 are arranged in a sequence that progresses from the left to the right.}}

\label{tab:prompt_list_tab2}
\vspace{3mm}
\end{minipage}
\end{table}  

\begin{table}[htbp]
\begin{minipage}[htbp]{1\linewidth}  
\centering  
\tablestyle{5pt}{1}
\resizebox{1.0\linewidth}{!}  
{  
\begin{tabular}{l|>{\centering\arraybackslash}m{8cm}|>{\centering\arraybackslash}m{8cm}}  
Category & Intension Prompt & GPT-4 Augmented Prompt\\  
\shline
advertising & \small{Create a customer testimonial advertisement showcasing a pink clothing collection. The advertisement includes a positive review from Jenny Wilson.} & \small{A burst of vibrant pink hues dominates the canvas: a mix of blush, fuchsia and rose clothing neatly arranged. In the corner, a gleaming starburst accentuates a glowing testimonial from Jenny Wilson, her words of praise shimmering. Beneath the testimonial, elegantly scripted contact information is highlighted: phone number 123-456-789, website WWW.FASHIONSTYLE.COM, and an address at 123 Anywhere ST., Any City.}\\ \hline
events & \small{Create a birthday card with a festive and colorful design featuring a cake with fruits. The card is an invitation to a birthday party happening on June 22nd from
8pm to 11pm at 123 st. Any city 4567.} & \small{A burst of vibrant pink hues dominates the canvas: a mix of blush, fuchsia and rose clothing neatly arranged. In the corner, a gleaming starburst accentuates a glowing testimonial from Jenny Wilson, her words of praise shimmering. Beneath the testimonial, elegantly scripted contact information is highlighted: phone number 123-456-789, website WWW.FASHIONSTYLE.COM, and an address at 123
Anywhere ST., Any City.}\\ \hline
creative & \small{Design an alluring poster for a blues concert in a clearing of the Amazon Rainforest. The artwork should evoke the soulful tunes of blues music echoing through
the wilderness. Use a moody, midnight blue palette and incorporate silhouettes of blues instruments and nocturnal wildlife.} & \small{An enticing poster for a blues concert set in an Amazon Rainforest clearing. The artwork is bathed in a mysterious, midnight blue hue, resonating the soulful melody of blues music. The scene is punctuated by silhouettes of blues instruments - a haunting saxophone, a melancholic guitar, and a poignant piano. The wilderness comes alive in the shadows of nocturnal creatures - a stealthy jaguar, an elusive tree frog, and a silent owl perched on a branch, all harmoniously intertwined with the rhythm of the blues.}\\
\end{tabular}
}
\caption{  
\small{The intension prompts and GPT-4 augmented prompts listed in Figure 11 are arranged in a sequence that progresses from the top to the bottom.}}   
\label{tab:prompt_list_fig11} 
\end{minipage}
\end{table}

\begin{table}[htbp]
\vspace{-3mm}
\begin{minipage}[htbp]{1\linewidth}  
\centering  
\tablestyle{5pt}{1.1}  
\resizebox{1.0\linewidth}{!}  
{  
\begin{tabular}{l|>{\centering\arraybackslash}m{15cm}}  
Category & Intention Prompt \\  
\shline  
posts & \small{Design a funny and creative Halloween Facebook post with humorous messages and stickers. Promote a Halloween party at Under Bar with the caption 'Get your treat, Trick or Treat, dude!'} \\ \hline 
marketing& \small{Design a gift certificate for a travel insurance company that offers a 30\% discount for all kinds of travel.} \\ \hline 
covers\& headers& \small{Create a Youtube thumbnail for promoting an antique store subscription service. Showcase the variety of unique vintage goods available in the store.} \\  \hline
creative & \small{Create a recruitment post for a company seeking an illustrator to join their team.} \\ \hline
marketing & \small{Design a colorful store logo with an illustration of leaves to represent a fresh and vibrant shopping experience. The store may be named TB.} \\  \hline
posts & \small{Create a Facebook post featuring a happy family with their child enjoying a leisurely day outdoors. Capture their stylish outfits and cute smile. Frame the image with a white border and add a shadow effect.} \\ \hline
posts & \small{Create a Facebook post to promote a special menu for pancakes at a cafe or restaurant. The offer is delicious pancakes for only \$10. The website for more information is www.pancakehouse.com.} \\ \hline
posts & \small{Create a happy and joyful image of a pregnant woman enjoying a relaxing day by the seacoast. The text emphasizes the excitement and love for the baby growing inside.} \\ \hline
covers\& headers & \small{Create a Facebook cover image to announce a flower exhibition event.}  \\  \hline
posts & \small{Design a Facebook post promoting a summer collection for women. Highlight the new arrival and offer a 60\% discount for this week only. Encourage viewers to shop now.} \\ \hline
posts & \small{Design a Facebook post advertising an organic cosmetics kit with natural and organic ingredients. The kit is a must-have for skincare and wellness.} \\ \hline
posts & \small{Promote a cheesy Italian pizza with a discount offer of 30\% for the first order. Encourage customers to visit the restaurant or order online. The promotion is for a limited time only.} \\ \hline
covers\& headers & \small{Design an advertisement for a crowdfunding platform called LocalBless. Encourage users to contribute to social projects by visiting the website and donating. Create a visually appealing image with a stone pattern background and triangles to convey a sense of stability and trust.} \\ \hline
posts & \small{Design a Facebook post promoting a two-week summer camp that provides hands-on, project-based instruction in tech. The camp will take place from 12-14 August.} \\ \hline
posts & \small{Create an Instagram story frame with a futuristic and fantasy-like design that features a man in an astronaut helmet and a space-themed background. The text \#ThisIsMetaLife and SPACEDOUT are included.} \\ \hline
posts & \small{Create a minimalistic and modern home design discount offer for a website or social media post. Encourage users to visit the website www.dreamhome.com to enjoy a 10\% discount on their first design.} \\ \hline
covers\& headers & \small{Create an attractive YouTube thumbnail for a makeup review video featuring an African American woman. The video will showcase the woman's favorite makeup products and provide a weekly review.} \\ \hline
posts & \small{Promote a personal styling service for summer vacation outfits on Instagram. The contact number is +00 1357 2468 and website is www.website.com.} \\ \hline
marketing & \small{Create an advertisement poster for a college merchandise store that has been operating since 1991. The design should focus on college and university branded gear, including clothes, backpacks, and bags.} \\ \hline 
posts & \small{Create an Instagram story ad for an antique shop, promoting the sale of a timeless typewriter. Encourage customers to visit the store by swiping up on the story.} \\ \hline
covers\& headers & \small{Create a Youtube thumbnail for a video about real estate investments, providing tips from professionals.} \\ \hline
posts & \small{Create a Facebook post promoting USA Independence Day tours with a patriotic theme. Encourage people to find their freedom in life by traveling more. Mention the offer of free cancellation. Include the website patrioutus.com for more information.} \\ \hline
events & \small{Create a Valentine's Day greeting card with a cute and romantic design. The card includes the phrase 'Happy Valentine's Day' and 'Much love to you!'} \\ \hline
posts & \small{Create a Facebook post to celebrate International Pet Day with a cute little girl and a fluffy rabbit.} \\
\end{tabular}  
} 

\caption{  
\small{The intention prompts listed in Figure 12 are arranged in a sequence that progresses from the top left to the bottom right.}}
\label{tab:prompt_list_tab4}
\vspace{1mm}
\end{minipage}
\end{table}

\begin{table}[]
\begin{minipage}[htbp]{1\linewidth}  
\centering  
\tablestyle{5pt}{1}
\resizebox{1.0\linewidth}{!}  
{  
\begin{tabular}{l|>{\centering\arraybackslash}m{15cm}}  
Category & Image Generate Prompt \\  
\shline
marketing & \small{Gift Certificate. blank, empty, layout, tb, certificate, offer, illustration, pattern, floral, flowers, pink, frame, sale. The image features a white background with a pink border. The border is adorned with a decorative floral pattern, giving the image a visually appealing and elegant appearance. The combination of the white background and the pink border with the floral pattern creates a harmonious and attractive design.
} \\ \hline
posts & \small{Instagram Story. astronaut, cosmic, cosmonaut, helmet, orange, science, space, spaceman, universe, story, instagram, ig, insta, frame. The image features a man wearing a spacesuit, sitting on a planet with his legs crossed. He is positioned in the center of the scene, and the planet is displayed in the background. The man appears to be a spaceman, possibly an astronaut, as he is dressed in a suit designed for space travel. The scene is set against a dark blue sky, which adds to the overall atmosphere of the image.} \\ \hline
marketing & \small{Gift Certificate. agency, holiday, journey, service, tourism, travel, traveling, trip, vacation, tb, certificate, sale, discount, insurance, illustration, cartoon, blue. The image features a map with a variety of travel-related items scattered around it. There are several pairs of shoes, including a couple of boots, placed in different positions on the map. A hat is also present, positioned near the top left corner of the map. In addition to the shoes and hat, there are multiple books scattered across the map, with some located near the bottom left corner and others in the middle and top right areas. A pair of scissors can be seen in the top right corner of the map, and a clock is placed near the center of the map.} \\ \hline
events & \small{Business Card European. data, card, health, home, professional, service, technology, icon, house, nursing, care, modern, medical, health care, contact, information. The image features a house with a pink roof, which is situated next to a street. The house is surrounded by a white fence, and there is a person standing on the sidewalk in front of the house. The person appears to be looking at the house, possibly admiring its architecture or considering it for purchase. The scene also includes a car parked on the street, adding to the overall atmosphere of the neighborhood.} \\ \hline
events & \small{Instagram Post. wedding, dinner, rehearsal, banquet, event, holiday, celebration, marriage, food, party, restaurant, ceremony, leaf, jungle, tropical, exotic, brown, instagram, post, insta, ig, tb. The image features a brown background with a few green leaves scattered throughout the scene. The leaves are of various sizes and are positioned at different angles, creating a sense of depth and texture. The overall composition of the image is simple, with the brown background and green leaves being the main focus.} \\ \hline
events & \small{Instagram Post. empowerment, equality, feminism, holiday, march, march 8, rights, women, women's day, womensday, tb, illustration, female, girls, celebration, beautiful, greeting, sisterhood, instagram, insta, ig, post, flowers, floral. The image features two people standing next to each other, each holding a heart. The hearts are positioned above their heads, creating a visually striking scene. The people are standing in front of a backdrop of flowers, adding a touch of color and beauty to the image.} \\ \hline
covers\& headers & \small{Ticket DIN large. festival, art, abstract, geometric, graphic, illustration, pattern, shape, simple, visual, colorful, blue, ticket. The image features a colorful and abstract design, with a predominantly blue background. The design consists of various geometric shapes, including triangles, squares, and circles, arranged in a visually appealing manner. The shapes are positioned throughout the image, with some closer to the foreground and others further in the background. The overall composition creates a sense of depth and complexity, making the image visually engaging and interesting.} \\ \hline
advertising & \small{Square Graphic Post. tb, instagram, insta, ig, post, style, fashion, trendy, fashionable, stylish, beautiful, clothes, model, woman, posing, modern, female, young, casual, beauty, attractive, look, color, modeling, lady, people, outfit, order, new arrival, pink, customer, testimonial, review. The image features a pink background with a white square placed in the center. The square is surrounded by a pink hue, giving it a slightly pinkish tint. The background is predominantly pink, with no other objects or distractions visible in the scene.} \\
\end{tabular}
}
\caption{  
\small{The image generator prompts listed in Figure~\ref{fig:compare_to_sota_bg} are arranged in a sequence that progresses from the top to the bottom.}}
\label{tab:prompt_list_tab5} 
\end{minipage}
\end{table}

\begin{table*}[htb]
\centering
\tablestyle{12pt}{1.3}
\resizebox{1\linewidth}{!}
{
\begin{tabular}{l|c|c|c|c|c|c|c}
\textbf{Setting} & \textbf{Design LLM} & \multicolumn{2}{c}{\textbf{Text-to-Background DM}} & \multicolumn{2}{|c|}{\textbf{Text-to-Object DM}} & \textbf{Typography LMM} & \textbf{Reflect LMM} \\  
\cline{3-6} 
\textbf{} & \textbf{} & Stage 1 & Stage 2 & Stage 1 & Stage 2 & \textbf{} & \textbf{} \\ \shline
Training epochs & $10$ & $10$ & $10$ & $20$ & $20$ & $6$ & $4$ \\
Batch size & $64$  & $64$ & $36$ & $64$ & $24$ &$32$ & $64$ \\  
Learning rate & $5$e-$4$  & $5$e-$5$ & $5$e-$5$ & $5$e-$5$ &$5$e-$5$ & $2$e-$4$ & $2$e-$4$ \\
AdamW $\epsilon$ & $1$e-$6$ & $1$e-$8$ & $1$e-$8$&$1$e-$8$ & $1$e-$8$ & $1$e-$6$ &  $1$e-$6$ \\
AdamW $\beta$ & ($0.9$, $0.999$)  & ($0.9$, $0.999$)  & ($0.9$, $0.999$)  & ($0.9$, $0.999$) & ($0.9$, $0.999$) & ($0.9$, $0.999$) & ($0.9$, $0.999$) \\
Learning rate schedule & Constant &Cosine & Cosine& Cosine & Cosine& Cosine & Cosine \\
Warmup steps & $50$  & $50$ & $50$ & $50$ & $50$ & $100$ & $100$\\
\hline
Max token length & $1024$ & - & - & - & - & $2048$ & $2048$ \\
Training strategy & $50$\% Causal, $50$\% CM3~\cite{aghajanyan2022cm3} & - & - & - & - & Causal & Causal \\
Tuned parameters & $13$B & $76$M & $41$M & $4.3$B & $1.2$B & $0.8$B & $0.8$B \\
LoRA rank & - & $128$ & $128$ & -& -& $128$ & $128$ \\
LoRA alpha & - & - & - & - &- &  $256$ & $256$ \\
\hline
Input resolution & - & $64^2$ & $256^2$ & $64^2$ & $256^2$ & $336^2$ & $336^2$\\
\end{tabular}}
\caption{Illustrating the detailed hyperparameters used for training each generation model.}
\label{tab:hyper_designllm} 
\end{table*}

As a result, the background images produced by our specialized diffusion models prove more conducive for integration with diverse elements, such as the object image layer and the typography image layer. This suitability arises from their ability to create backgrounds that allow for seamless incorporation of these various elements, fostering enhanced compositional flexibility.

\begin{figure}[htbp]
\begin{minipage}[htbp]{1\linewidth}
\centering
\hspace{10mm}
\begin{minipage}{1\linewidth}
\centering
\includegraphics[width=\textwidth]{{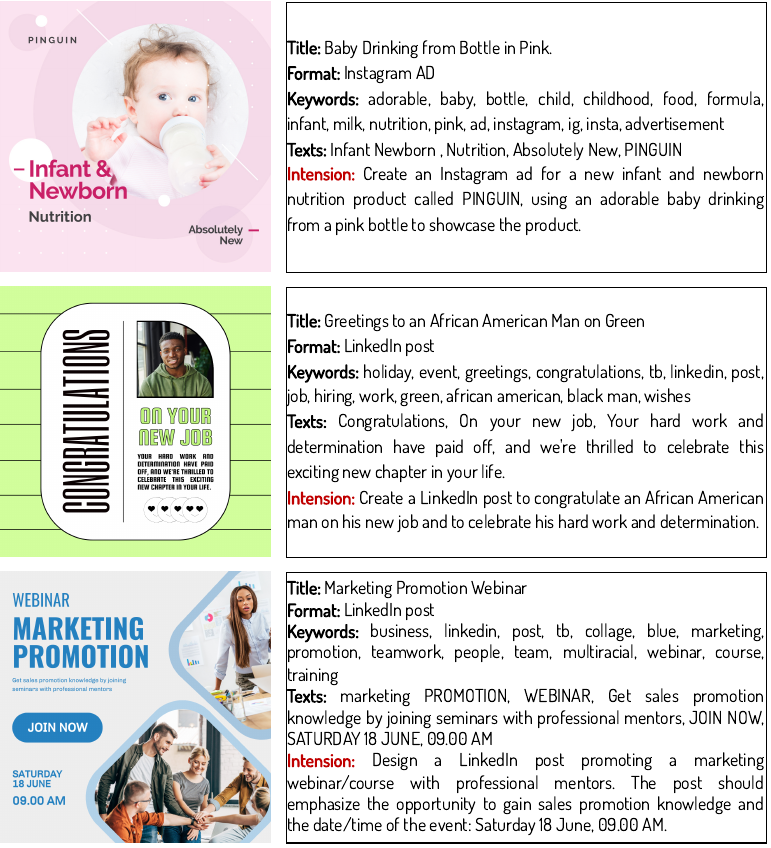}}
\end{minipage}%
\caption{\small{Illustrating some examples of generating designer intention prompts derived from the original image information present  in the raw data. We choose GPT-3.5 by default.}}
\label{fig:intention_prompt_example}
\end{minipage}
\end{figure}

\begin{figure}[htbp]
\begin{minipage}[htbp]{1\linewidth}
\vspace{5mm}
\begin{minipage}[htbp]{1\linewidth}
\centering
\begin{subfigure}[b]{0.232\textwidth}
{\includegraphics[width=\textwidth]{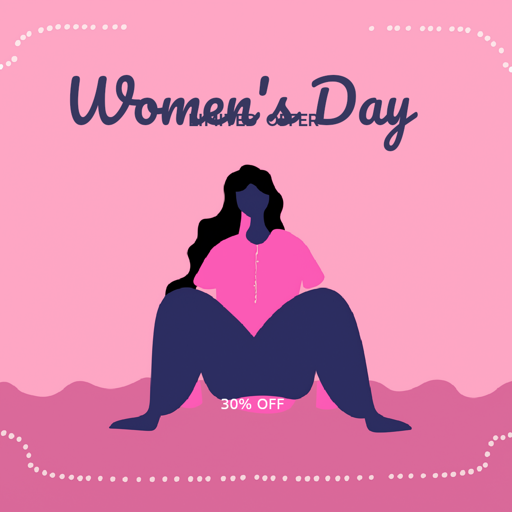}}
\vspace{-3mm}
\end{subfigure}
\begin{subfigure}[b]{0.232\textwidth}
{\includegraphics[width=\textwidth]{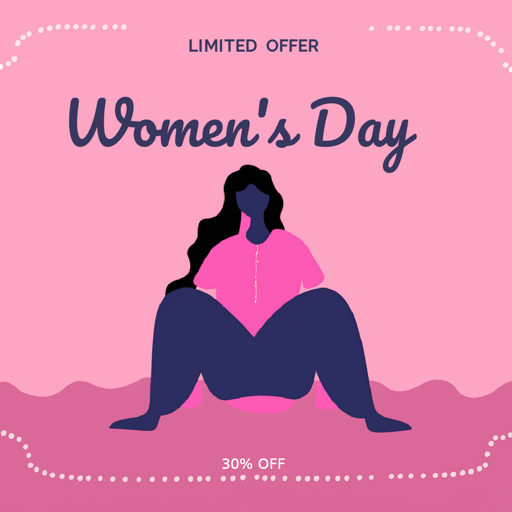}}
\vspace{-3mm}
\end{subfigure}
\begin{subfigure}[b]{0.232\textwidth}
{\includegraphics[width=\textwidth]{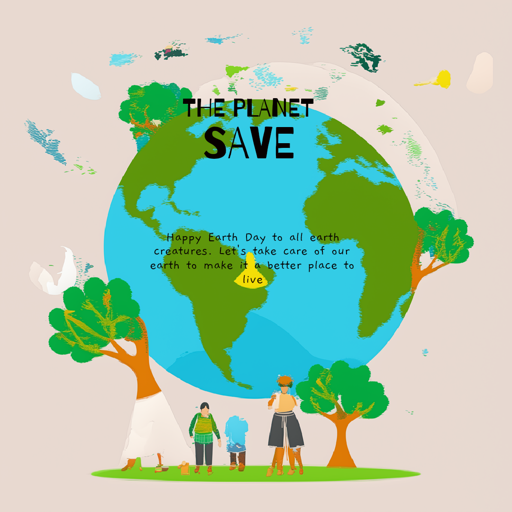}}
\vspace{-3mm}
\end{subfigure}
\begin{subfigure}[b]{0.232\textwidth}
{\includegraphics[width=\textwidth]{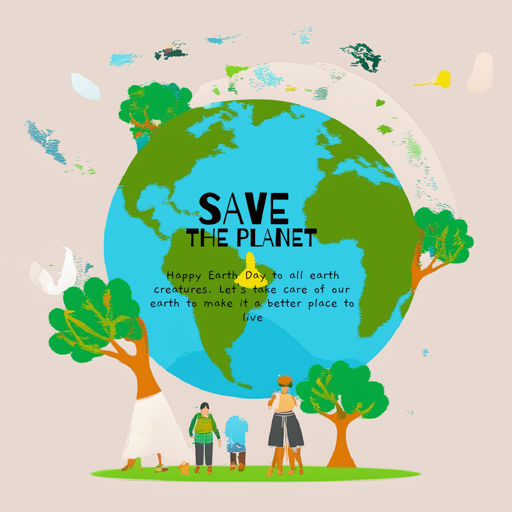}}
\vspace{-3mm}
\end{subfigure}\\
\begin{subfigure}[b]{0.232\textwidth}
{\includegraphics[width=\textwidth]{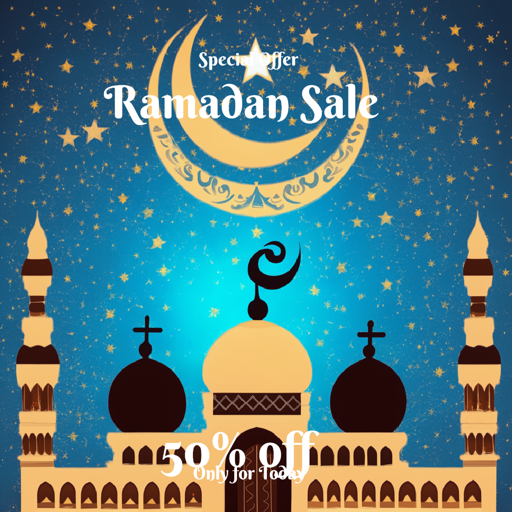}}
\caption*{\scriptsize{Typography LMM}}
\vspace{-1mm}
\end{subfigure}
\begin{subfigure}[b]{0.232\textwidth}
{\includegraphics[width=\textwidth]{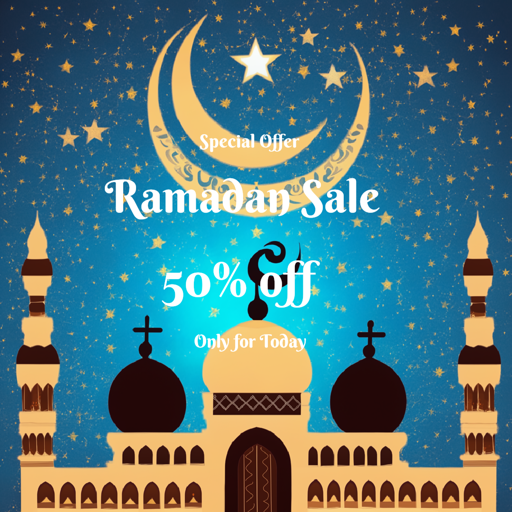}}
\caption*{\scriptsize{+ Reflect LMM}}
\vspace{-1mm}
\end{subfigure}
\begin{subfigure}[b]{0.232\textwidth}
{\includegraphics[width=\textwidth]{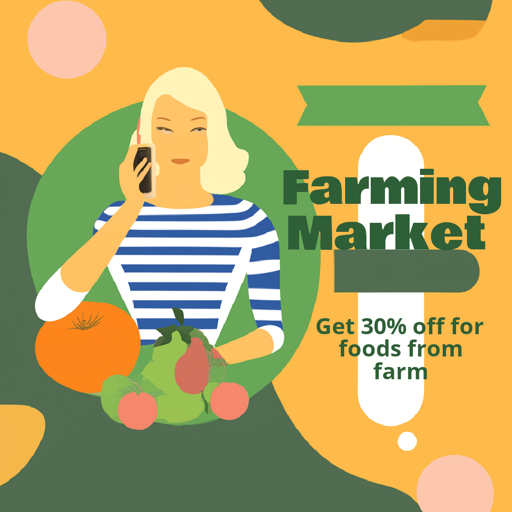}}
\caption*{\scriptsize{Typography LMM}}
\vspace{-1mm}
\end{subfigure}
\begin{subfigure}[b]{0.232\textwidth}
{\includegraphics[width=\textwidth]{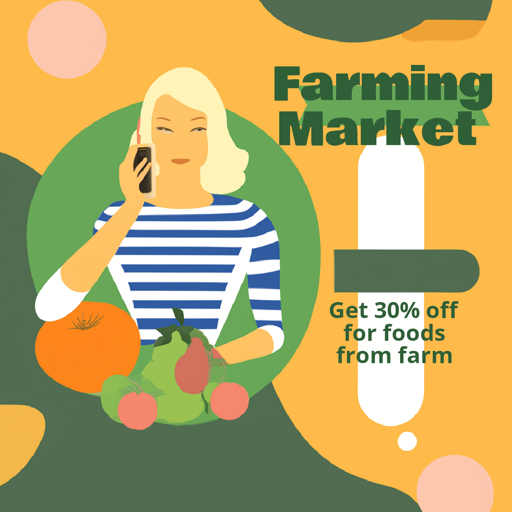}}
\caption*{\scriptsize{+ Reflect LMM}}
\vspace{-1mm}
\end{subfigure}
\end{minipage}
\caption{\footnotesize{Qualitative comparison of the typography position predictions generated by the Typography LMM and those refined by the Reflect LMM.}}
\label{fig:compare_typography_lmm_to_reflect_lmm}
\vspace{5mm}
\end{minipage}
\end{figure}

\begin{figure}
\begin{minipage}[htbp]{1\linewidth}
\centering
\hspace{5mm}
\begin{minipage}{1\linewidth}
\centering
{\includegraphics[width=1\textwidth]{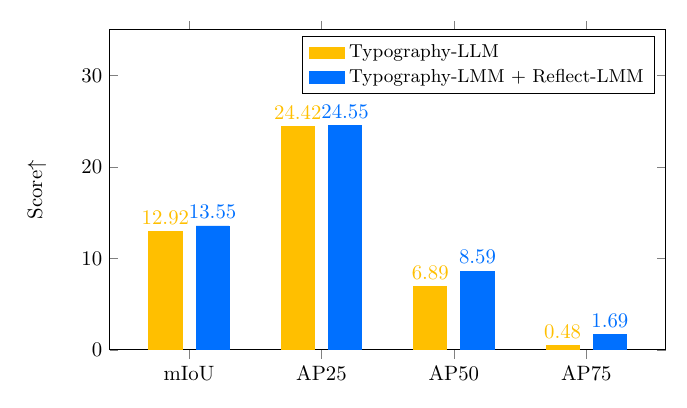}}
\end{minipage}%
\caption{\small{Illustrating the typography localization performance with Typography-LMM or Typography-LMM + Reflect-LMM.}}
\label{fig:ablate_typography_lmm}
\end{minipage}
\end{figure}

\begin{figure}[t]
\begin{minipage}[t]{1\linewidth}
\centering
\begin{subfigure}[b]{0.242\textwidth}
{\includegraphics[width=\textwidth]{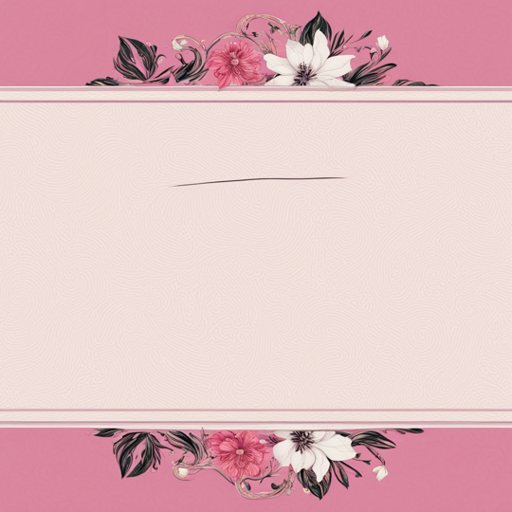}}
\vspace{-4mm}
\end{subfigure}
\begin{subfigure}[b]{0.242\textwidth}
{\includegraphics[width=\textwidth]{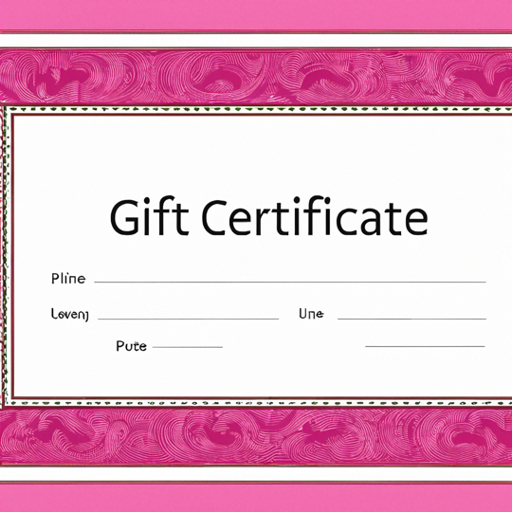}}
\vspace{-4mm}
\end{subfigure}
\begin{subfigure}[b]{0.242\textwidth}
{\includegraphics[width=\textwidth]{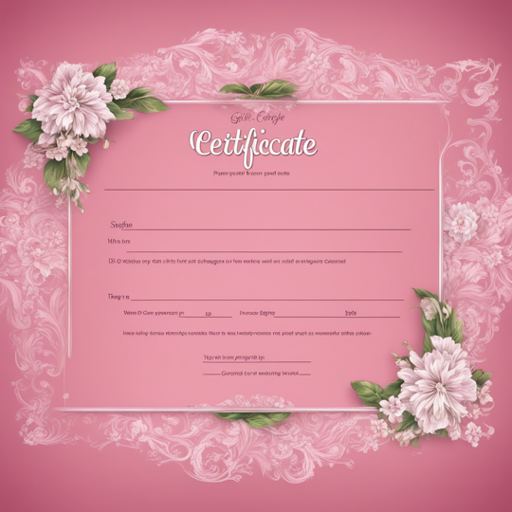}}
\vspace{-4mm}
\end{subfigure}
\begin{subfigure}[b]{0.242\textwidth}
{\includegraphics[width=\textwidth]{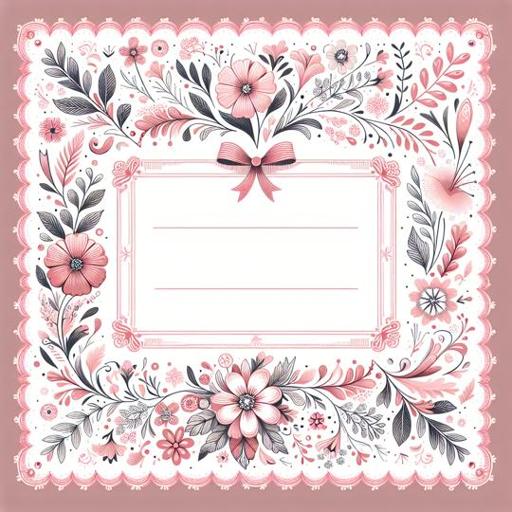}}
\vspace{-4mm}
\end{subfigure}\\
\begin{subfigure}[b]{0.242\textwidth}
{\includegraphics[width=\textwidth]{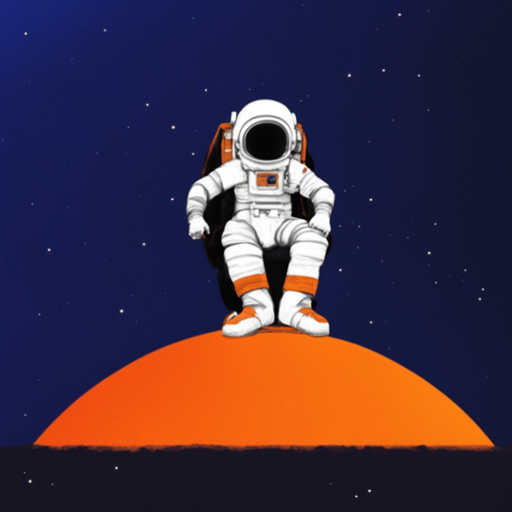}}
\vspace{-4mm}
\end{subfigure}
\begin{subfigure}[b]{0.242\textwidth}
{\includegraphics[width=\textwidth]{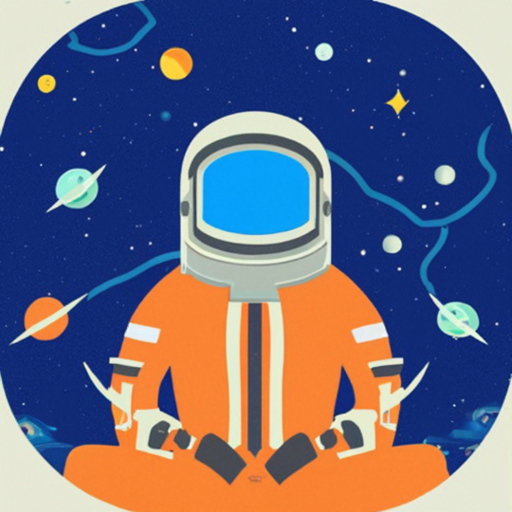}}
\vspace{-4mm}
\end{subfigure}
\begin{subfigure}[b]{0.242\textwidth}
{\includegraphics[width=\textwidth]{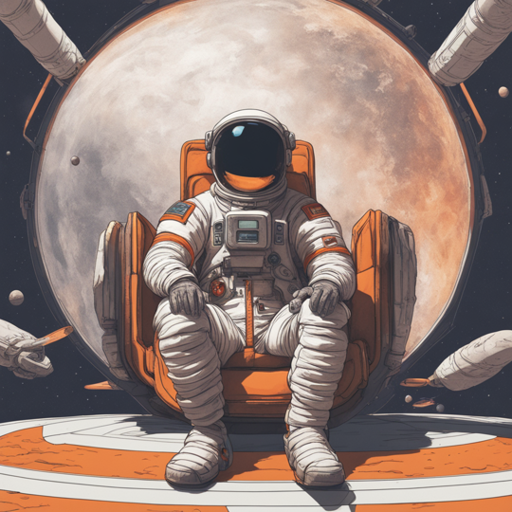}}
\vspace{-4mm}
\end{subfigure}
\begin{subfigure}[b]{0.242\textwidth}
{\includegraphics[width=\textwidth]{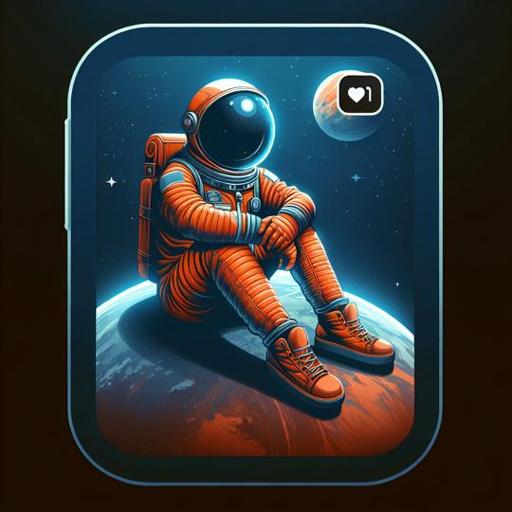}}
\vspace{-4mm}
\end{subfigure}\\
\begin{subfigure}[b]{0.242\textwidth}
{\includegraphics[width=\textwidth]{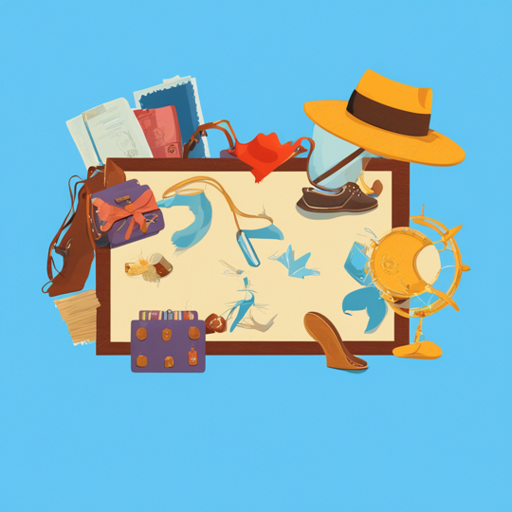}}
\vspace{-4mm}
\end{subfigure}
\begin{subfigure}[b]{0.242\textwidth}
{\includegraphics[width=\textwidth]{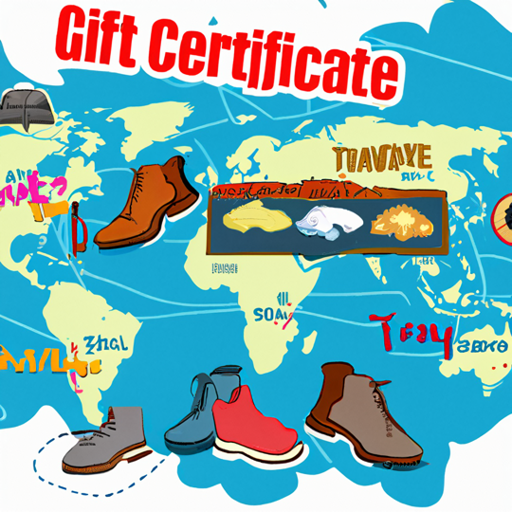}}
\vspace{-4mm}
\end{subfigure}
\begin{subfigure}[b]{0.242\textwidth}
{\includegraphics[width=\textwidth]{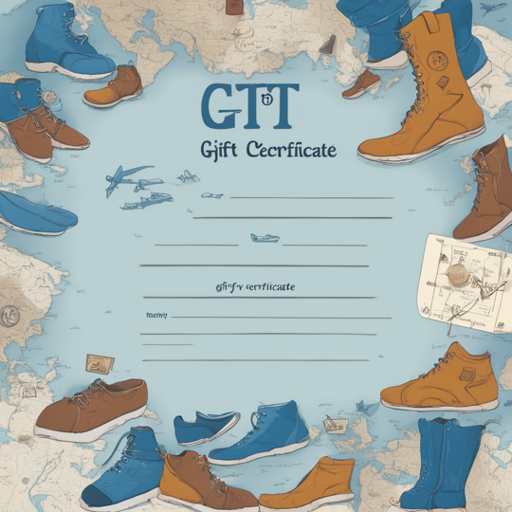}}
\vspace{-4mm}
\end{subfigure}
\begin{subfigure}[b]{0.242\textwidth}
{\includegraphics[width=\textwidth]{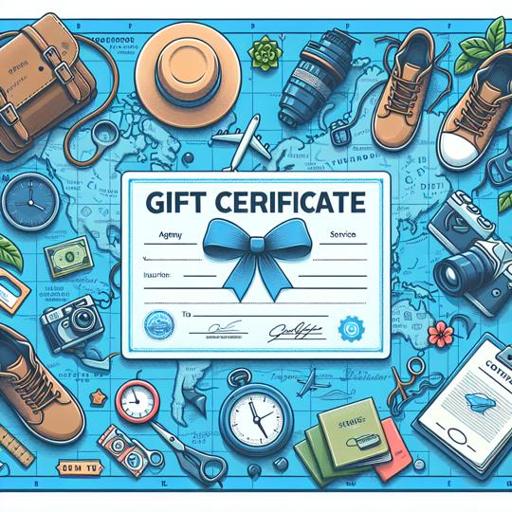}}
\vspace{-4mm}
\end{subfigure}\\
\begin{subfigure}[b]{0.242\textwidth}
{\includegraphics[width=\textwidth]{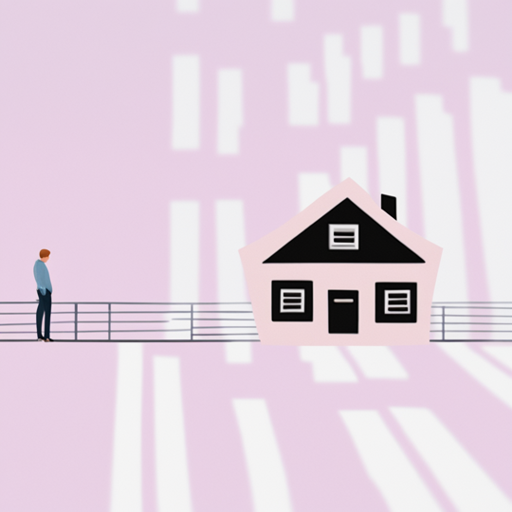}}
\vspace{-4mm}
\end{subfigure}
\begin{subfigure}[b]{0.242\textwidth}
{\includegraphics[width=\textwidth]{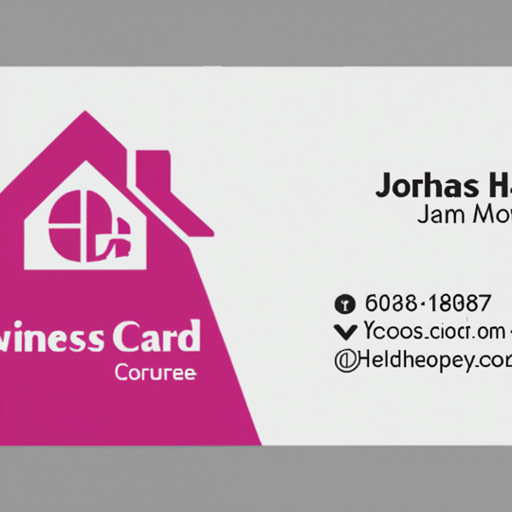}}
\vspace{-4mm}
\end{subfigure}
\begin{subfigure}[b]{0.242\textwidth}
{\includegraphics[width=\textwidth]{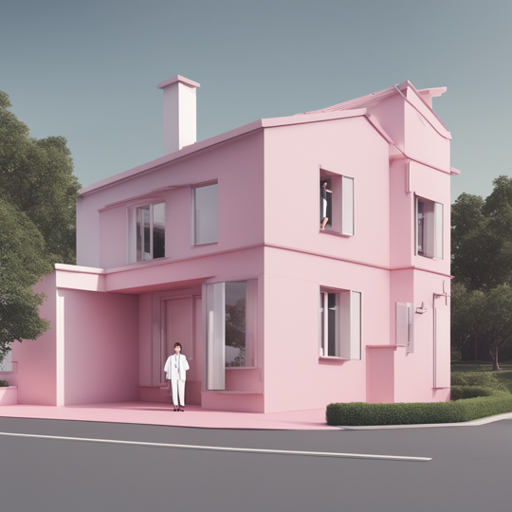}}
\vspace{-4mm}
\end{subfigure}
\begin{subfigure}[b]{0.242\textwidth}
{\includegraphics[width=\textwidth]{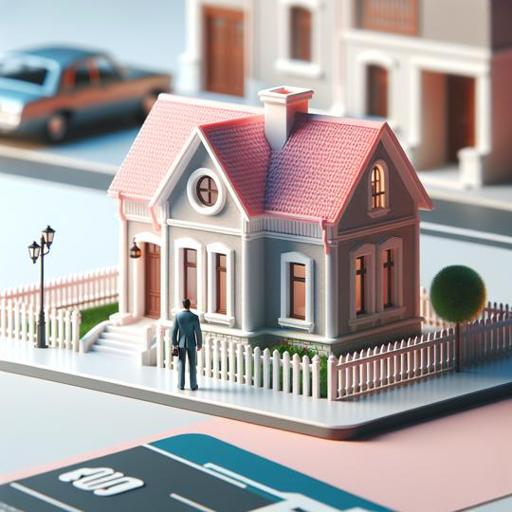}}
\vspace{-4mm}
\end{subfigure}\\
\begin{subfigure}[b]{0.242\textwidth}
{\includegraphics[width=\textwidth]{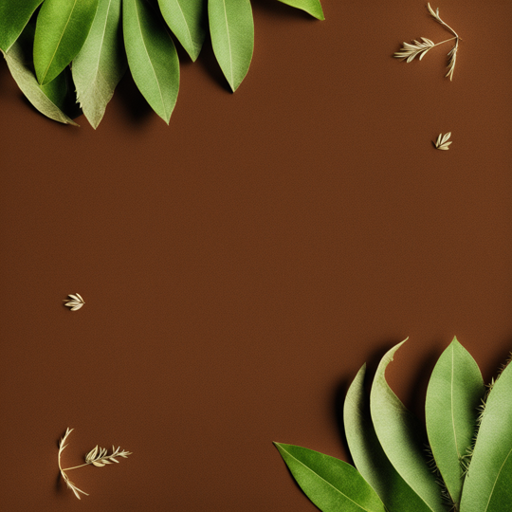}}
\vspace{-4mm}
\end{subfigure}
\begin{subfigure}[b]{0.242\textwidth}
{\includegraphics[width=\textwidth]{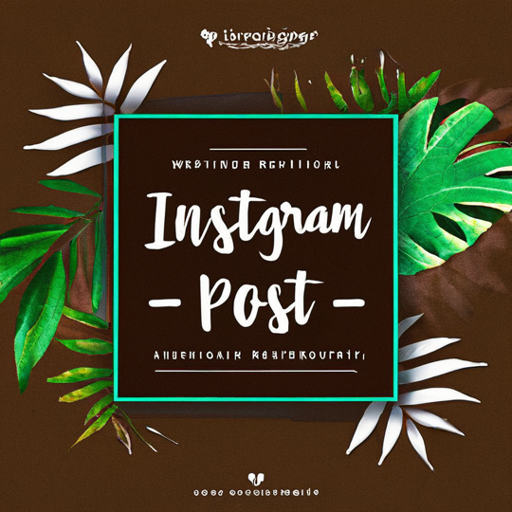}}
\vspace{-4mm}
\end{subfigure}
\begin{subfigure}[b]{0.242\textwidth}
{\includegraphics[width=\textwidth]{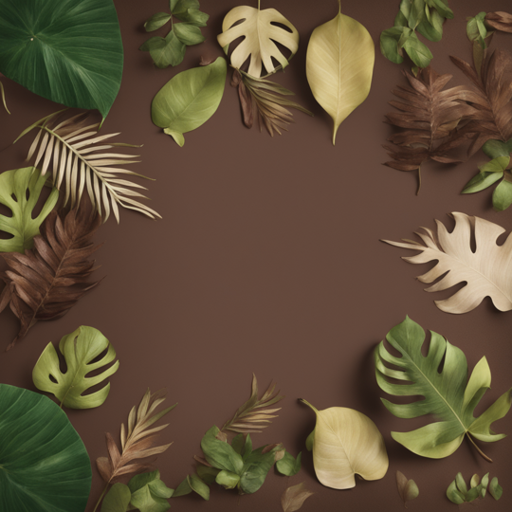}}
\vspace{-4mm}
\end{subfigure}
\begin{subfigure}[b]{0.242\textwidth}
{\includegraphics[width=\textwidth]{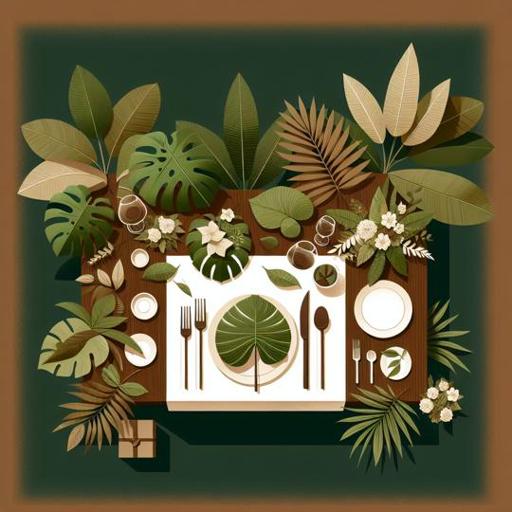}}
\vspace{-4mm}
\end{subfigure}\\
\begin{subfigure}[b]{0.242\textwidth}
{\includegraphics[width=\textwidth]{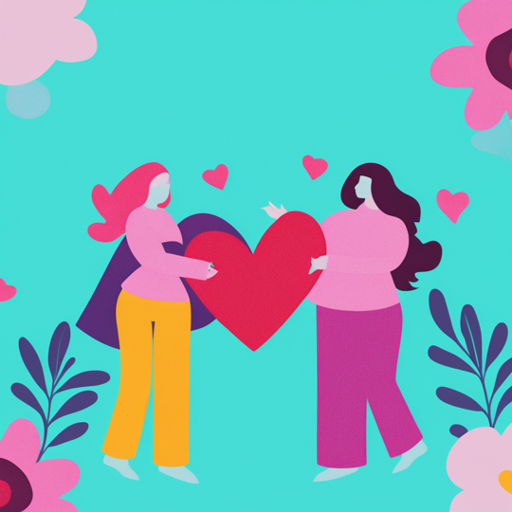}}
\vspace{-4mm}
\end{subfigure}
\begin{subfigure}[b]{0.242\textwidth}
{\includegraphics[width=\textwidth]{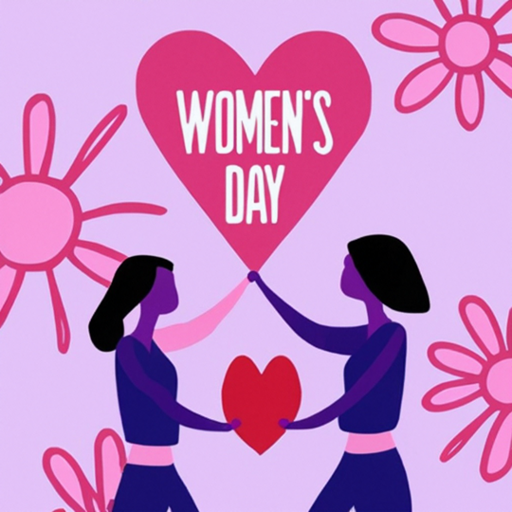}}
\vspace{-4mm}
\end{subfigure}
\begin{subfigure}[b]{0.242\textwidth}
{\includegraphics[width=\textwidth]{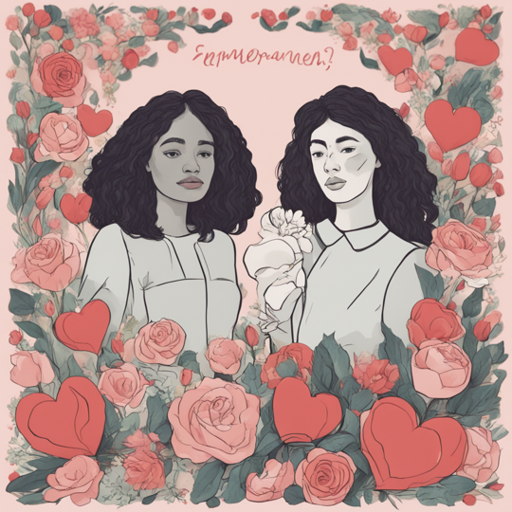}}
\vspace{-4mm}
\end{subfigure}
\begin{subfigure}[b]{0.242\textwidth}
{\includegraphics[width=\textwidth]{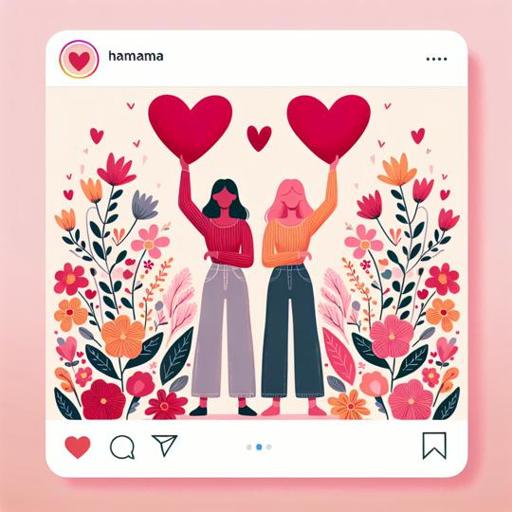}}
\vspace{-4mm}
\end{subfigure}\\
\begin{subfigure}[b]{0.242\textwidth}
{\includegraphics[width=\textwidth]{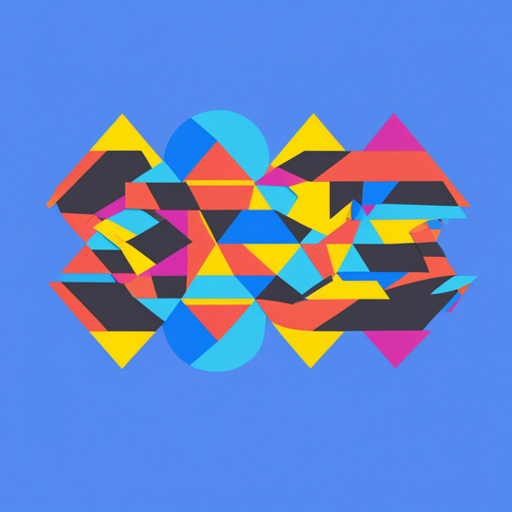}}
\vspace{-4mm}
\end{subfigure}
\begin{subfigure}[b]{0.242\textwidth}
{\includegraphics[width=\textwidth]{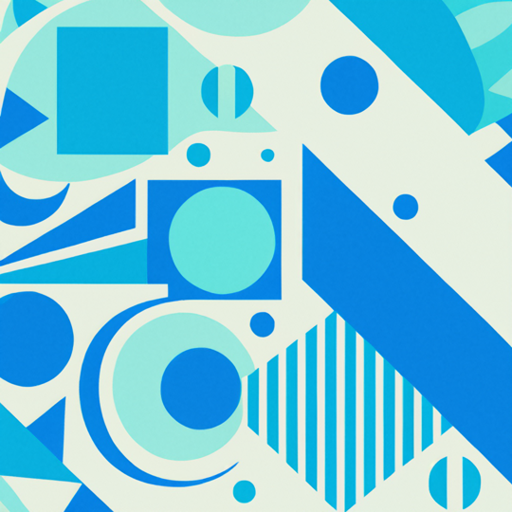}}
\vspace{-4mm}
\end{subfigure}
\begin{subfigure}[b]{0.242\textwidth}
{\includegraphics[width=\textwidth]{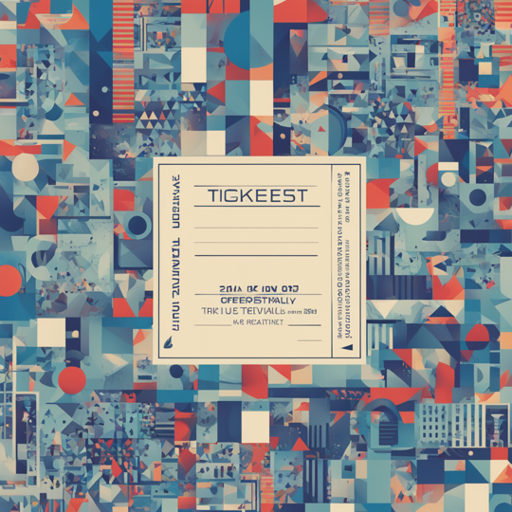}}
\vspace{-4mm}
\end{subfigure}
\begin{subfigure}[b]{0.242\textwidth}
{\includegraphics[width=\textwidth]{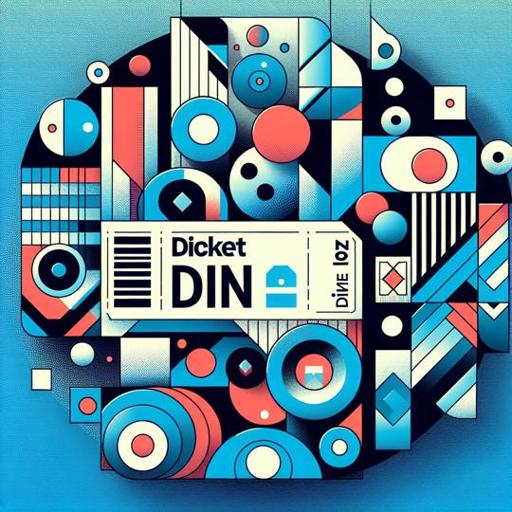}}
\vspace{-4mm}
\end{subfigure}\\
\begin{subfigure}[b]{0.242\textwidth}
{\includegraphics[width=\textwidth]{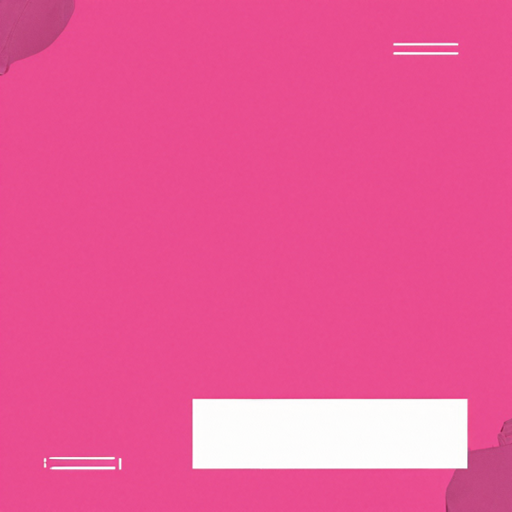}}
\caption*{\scriptsize{Ours}}
\vspace{-1mm}
\end{subfigure}
\begin{subfigure}[b]{0.242\textwidth}
{\includegraphics[width=\textwidth]{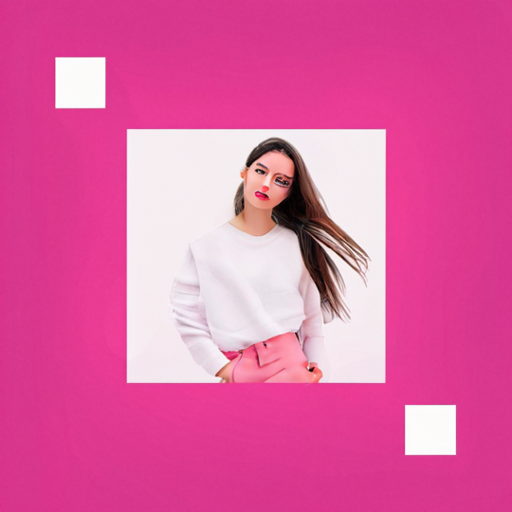}}
\caption*{\scriptsize{DeepFloyd/IF$^\dagger$}}
\vspace{-1mm}
\end{subfigure}
\begin{subfigure}[b]{0.242\textwidth}
{\includegraphics[width=\textwidth]{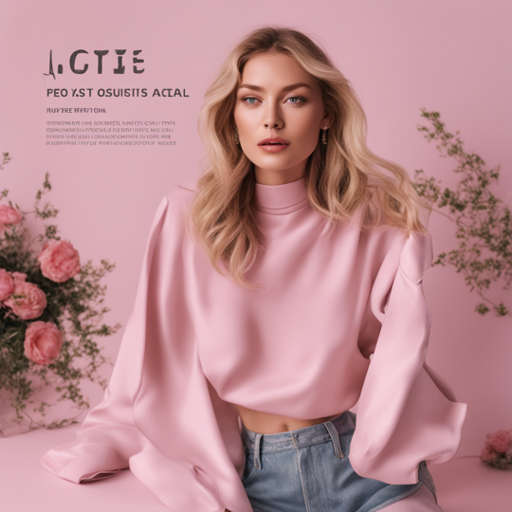}}
\caption*{\scriptsize{SDXL$^\dagger$}}
\vspace{-1mm}
\end{subfigure}
\begin{subfigure}[b]{0.242\textwidth}
{\includegraphics[width=\textwidth]{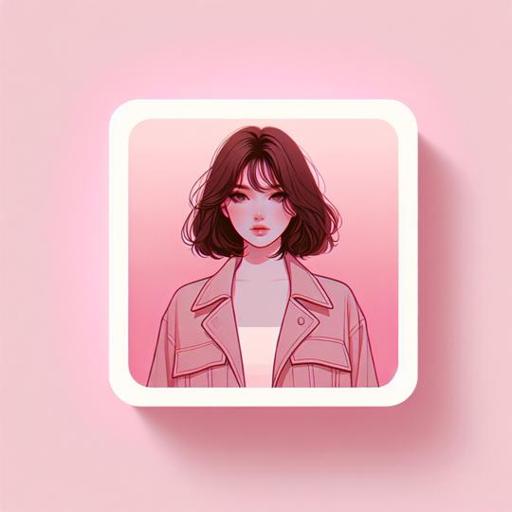}}
\caption*{\scriptsize{\dalle$^\dagger$}}
\vspace{-1mm}
\end{subfigure}\\
\caption{\footnotesize{Comparing our text-to-background image generation model to the previous state-of-the-art (SOTA) systems using the \ourbenchmark benchmark. For this comparison, we are solely focusing on the quality of the background image. We have selected the DeepFloyd/IF-XL and SDXL-1.0 + SDXL-Refiner-1.0 models for their exceptional performance. The superscript symbol $\dagger$ is used to denote the use of GPT-$4$ for enhancing the background prompt.}}
\vspace{-6mm}
\label{fig:compare_to_sota_bg}
\end{minipage}
\end{figure}

\vspace{2mm}
\noindent\textbf{Typography LMM \& Reflect LMM}
In Figure~\ref{fig:ablate_typography_lmm}, we report the localization accuracy comparison results, such as mIoU, AP$^{25}$, AP$^{50}$, and AP$^{75}$, for the predicted text boxes based on the Typography LMM and the refined text boxes derived from the Reflect LMM.
To facilitate fair comparisons and to utilize the ground-truth annotations of the text boxes, we employ the ground-truth background image and the ground-truth text value as the visual and textual conditions, respectively. Following this, the combination of these ground-truth conditions is inputted into the Typography LMM to predict the various typography attributes, including the text box positions denoted by the values of left, top, width, and height.
For the Reflect LMM, we input the rendered graphic design images (based on the Typography LMM) as the visual condition and the ground-truth text value as the textual condition, subsequently predicting the new values for left, top, width, and height.  
   
To compute the mIoU, AP$^{25}$, AP$^{50}$, and AP$^{75}$ scores, we initially compare each predicted text box with its corresponding ground-truth counterpart. Subsequently, these scores are averaged across all text boxes. As demonstrated in the comparison results in Figure~\ref{fig:ablate_typography_lmm}, we observe a significant improvement in performance, particularly for the AP$^{50}$ and AP$^{75}$ metrics.
For reference, we have illustrated the qualitative predictions derived from both the Typography LMM and Reflect LMM in Figure~\ref{fig:compare_typography_lmm_to_reflect_lmm}.

\noindent\textbf{Comparison by Professional Designer and GPT-4V}
In order to conduct a thorough and precise evaluation of the image quality produced by our \ourname and \dalle systems, we implement a strategy that incorporates both subjective and objective methodologies.  

\begin{figure*}[t]
\begin{minipage}[t]{1\linewidth}
\centering
\begin{subfigure}[b]{0.85\textwidth}
{\includegraphics[width=\textwidth]{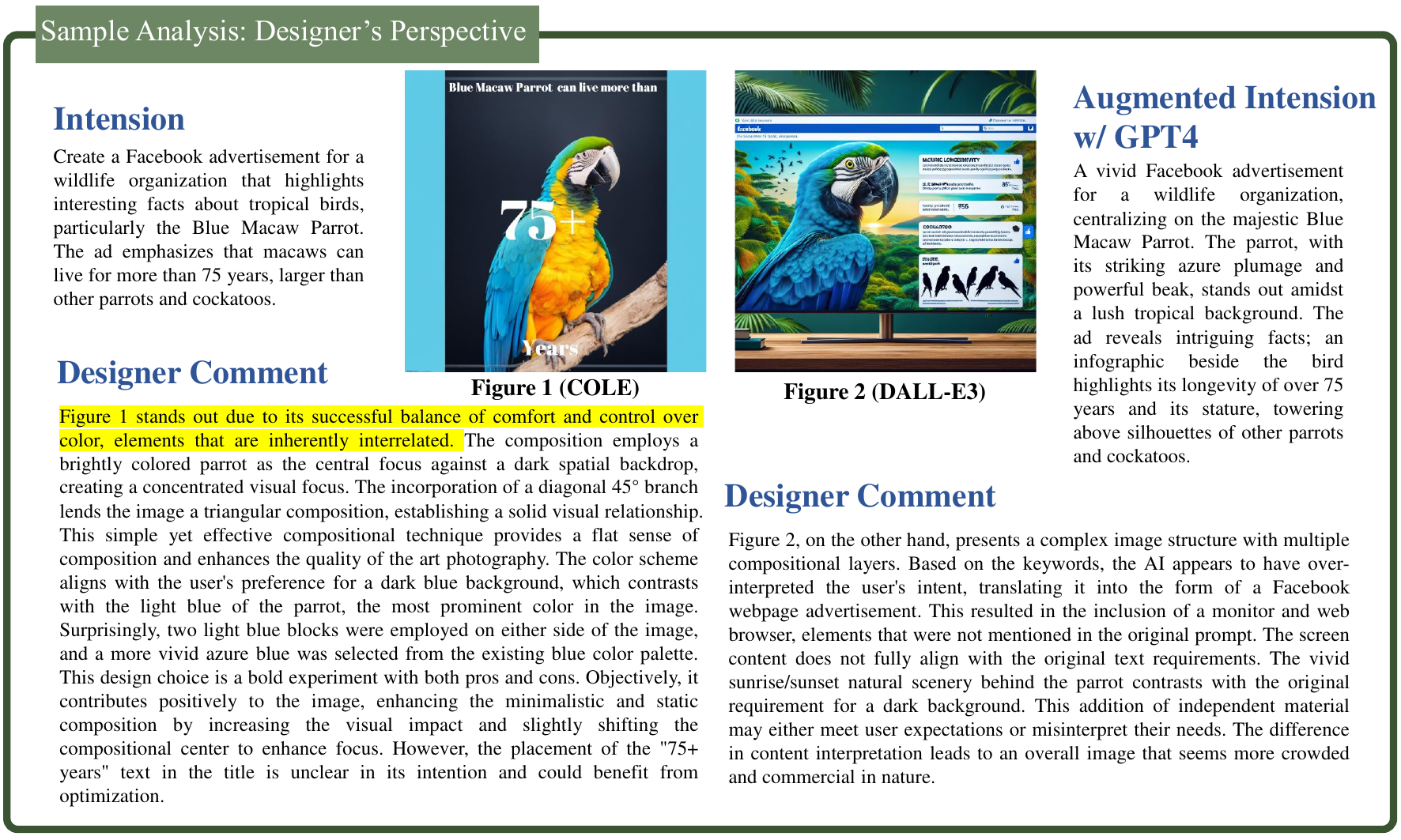}}
\end{subfigure}
\begin{subfigure}[b]{0.85\textwidth}
{\includegraphics[width=\textwidth]{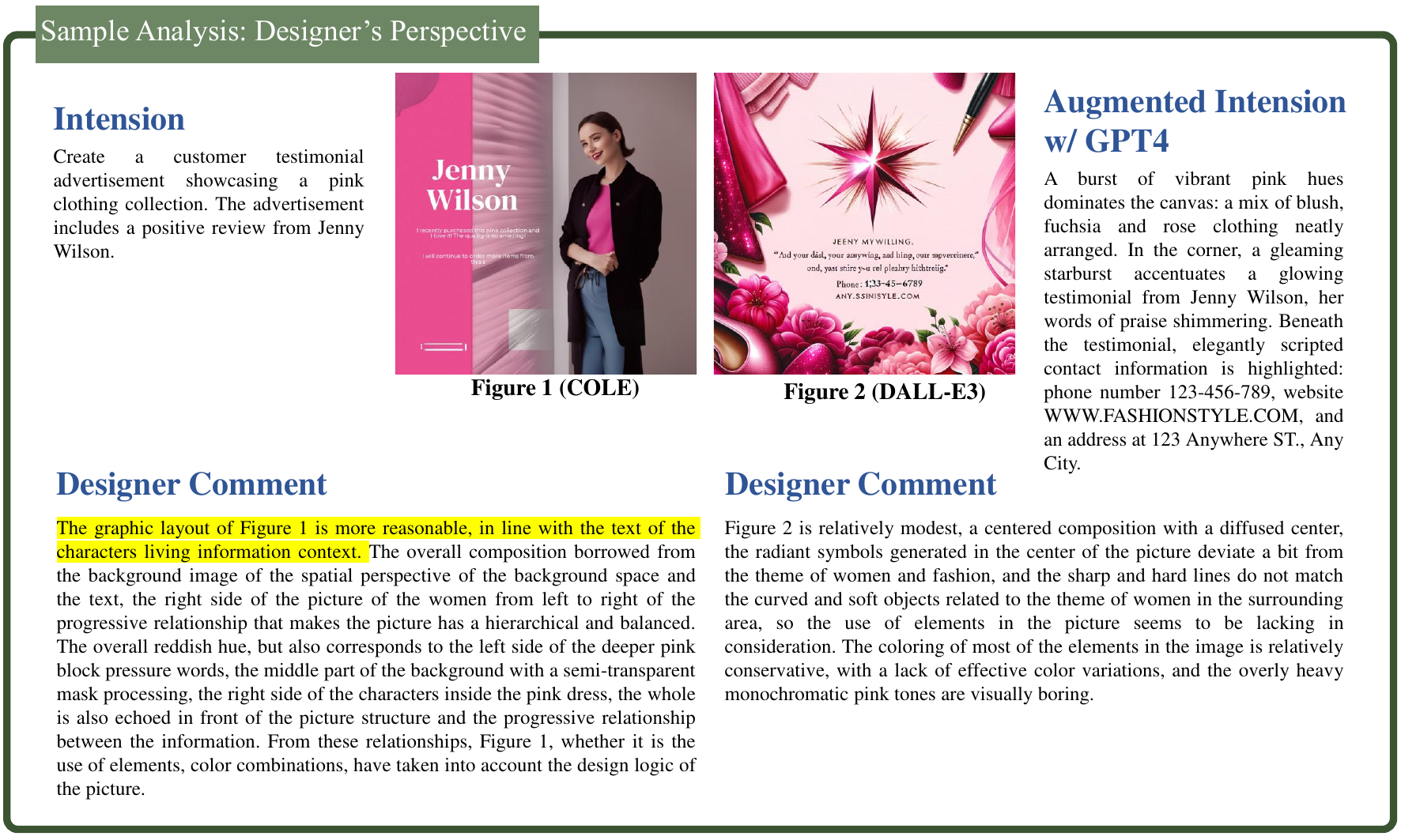}}
\end{subfigure}
\caption{\small{Illustrating the detailed comments provided by professional designer experts.}}
\label{fig:designer_comment}
\end{minipage}
\end{figure*}

On the subjective front, we leverage the expertise of a professional graphic designer to appraise the images created by \ourname and \dalle. This analysis takes into account various factors such as composition, subjective impressions, and color scheme. An elaborate presentation of this assessment can be found in Figure~\ref{fig:designer_comment}.
From an objective perspective, we create prompts akin to the \emph{Pair-wise Comparison Prompt for \gptv}, detailed in Section B, and input these prompts alongside two images into the \gptv model. We then conduct a comparative analysis of the images based on six distinct dimensions: \emph{overall composition layout}, \emph{image composition}, \emph{medium}, \emph{overall coherence}, \emph{overall aesthetics}, and \emph{text rendering}.  
   
\begin{figure*}[t]
\begin{minipage}[t]{1\linewidth}
\centering
\begin{subfigure}[b]{0.85\textwidth}
{\includegraphics[width=\textwidth]{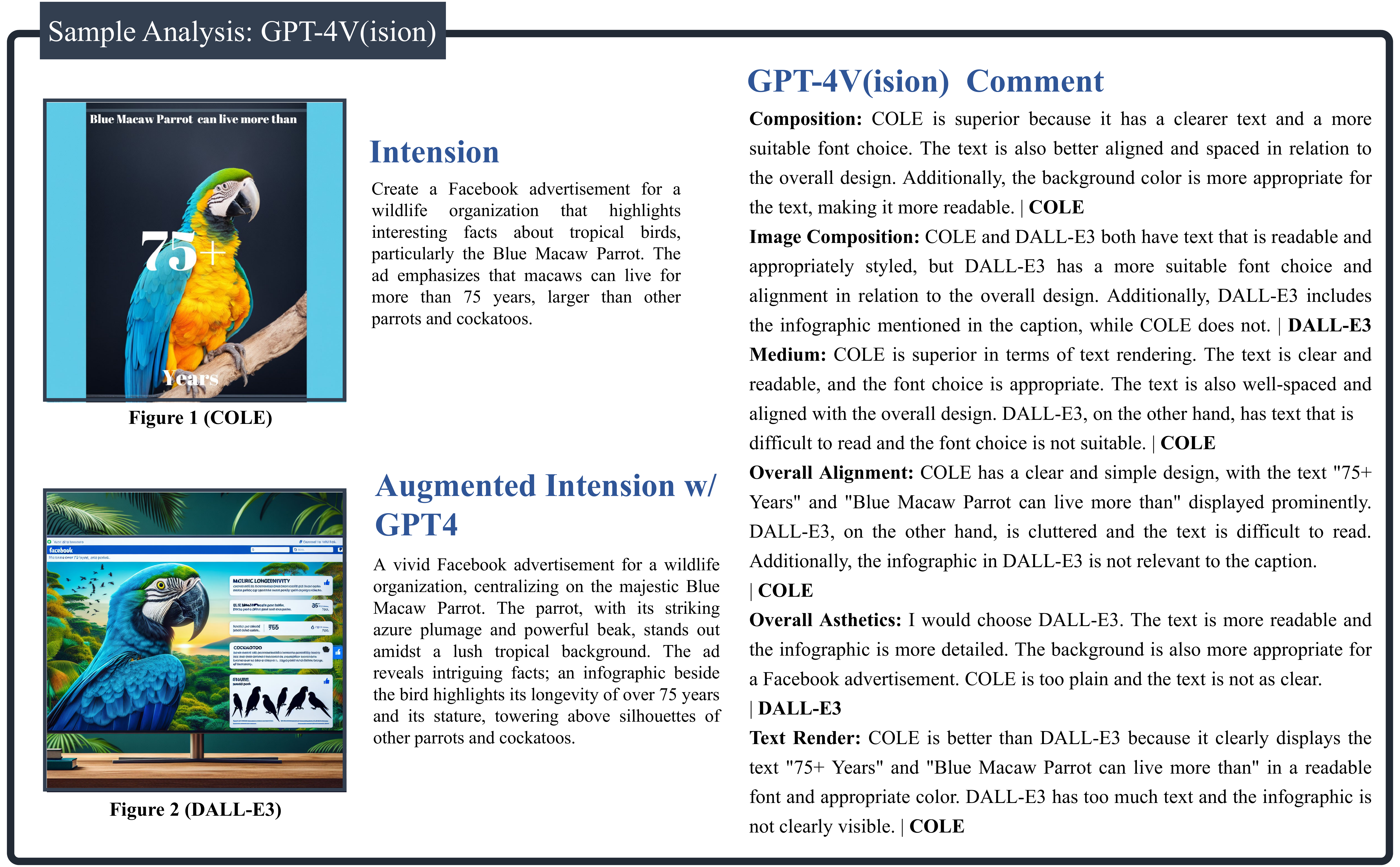}}
\end{subfigure}
\begin{subfigure}[b]{0.85\textwidth}
{\includegraphics[width=\textwidth]{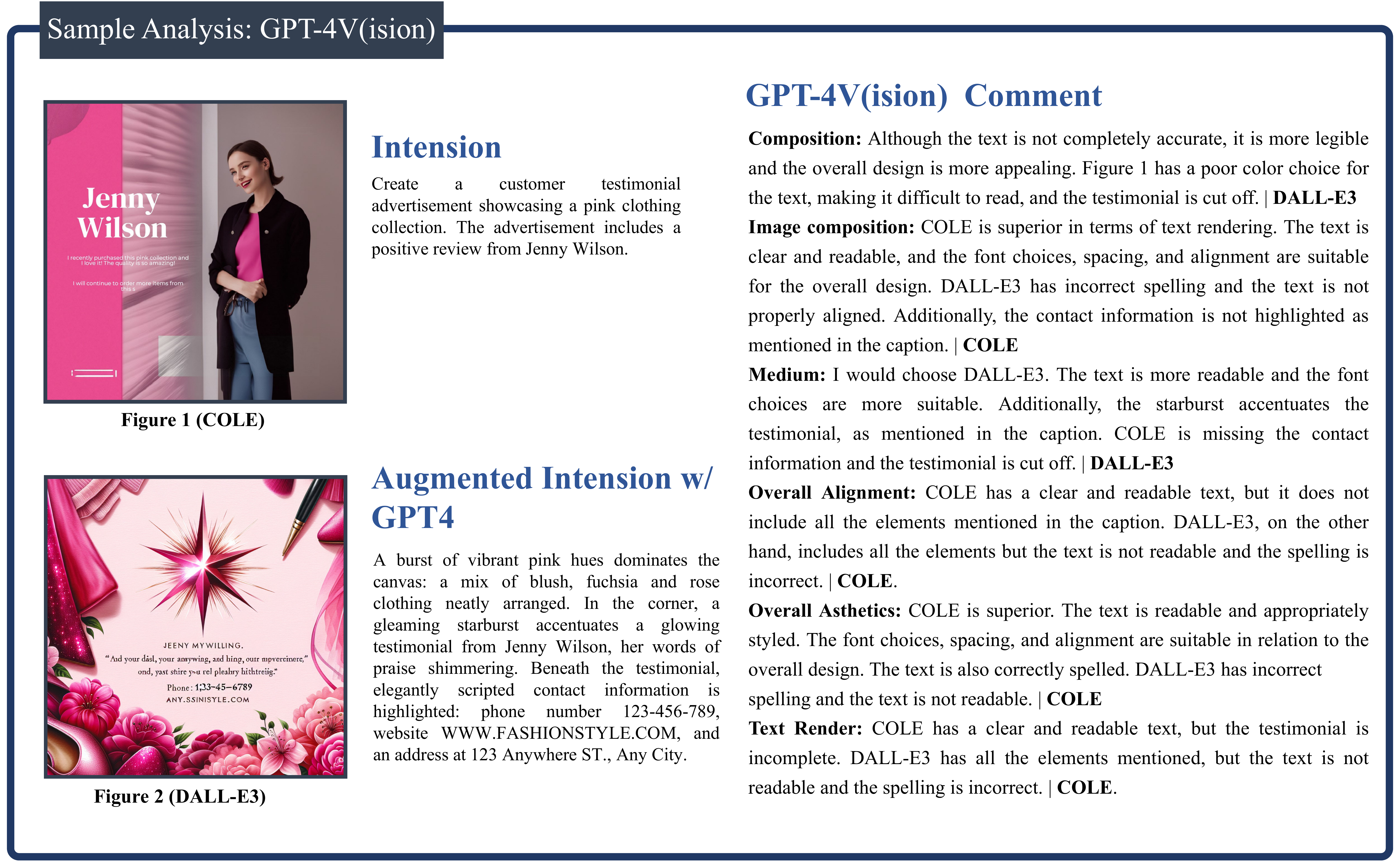}}
\end{subfigure}
\caption{\small{Illustrating the detailed comparison results provided by \gptv, covering six distinct aspects.}}
\label{fig:gpt4v_pk}
\end{minipage}
\end{figure*}

Specific outcomes of this comparison, along with the accompanying rationale, are illustrated in Figure~\ref{fig:gpt4v_pk}. As evidenced by the two provided examples, our model consistently generates images that surpass those produced by \dalle in four key areas. This highlights our model's superior performance, especially in terms of adherence to user intent and in the precision and readability of text rendering, as compared to \dalle.

\subsection*{F Hyperparameters}
Table~\ref{tab:hyper_designllm} outlines all the comprehensive training hyperparameters for our \ourname system, which is composed of a hierarchical series of five generative models. The table is segmented into three parts: training settings, model settings, and data settings, listed in that order from top to bottom.

\end{document}